\documentclass[10pt,twocolumn,letterpaper]{article}

\usepackage{cvpr}              

\usepackage{multirow,colortbl,xcolor,booktabs}
\newcommand{\whitemask}{\multicolumn{1}{>{\columncolor{white}}l}{}}
\newcommand{\ShadedOurs}[5]{%
  \whitemask &
  \multicolumn{1}{>{\columncolor{gray!20}}l}{\textbf{Ours (JS)}} &
  \multicolumn{1}{>{\columncolor{gray!20}}c}{\textbf{#1}} &
  \multicolumn{1}{>{\columncolor{gray!20}}c}{\textbf{#2}} &
  \multicolumn{1}{>{\columncolor{gray!20}}c}{\textbf{#3}} &
  \multicolumn{1}{>{\columncolor{gray!20}}c}{#4} &
  \multicolumn{1}{>{\columncolor{gray!20}}c}{#5}\\
}

\newcommand{\tabincell}[2]{\begin{tabular}{@{}#1@{}}#2\end{tabular}}

\usepackage{subcaption}
\usepackage{graphicx}
\usepackage{capt-of} 
\newcommand{\colhead}[1]{%
  \parbox[t]{0.315\linewidth}{\centering\footnotesize\bfseries \textit{#1}}%
}

\newcommand{\eimg}[1]{\includegraphics[width=\linewidth]{#1}}
\newcommand{\oimg}[1]{\includegraphics[width=\linewidth]{#1}}

\newcommand{\panel}[6]{%
  \begin{minipage}[t]{#1}\raggedright
  \textbf{(#2) #3}\par\vspace{2pt}
  \begin{minipage}[t]{0.495\linewidth}\eimg{#4}\end{minipage}\hfill
  \begin{minipage}[t]{0.495\linewidth}\oimg{#5}\end{minipage}\par\vspace{2pt}
  {\footnotesize #6}
  \end{minipage}
}

\usepackage{tikz}
\usetikzlibrary{arrows.meta,positioning,fit,calc}

\usepackage{algorithm}
\usepackage{algorithmic} 

\usepackage{tcolorbox}
\tcbuselibrary{theorems}

\usepackage{amsmath,amssymb,mathtools}

\newcommand{\E}{\mathbb{E}}
\newcommand{\MSE}{\operatorname{MSE}}
\newcommand{\Var}{\operatorname{Var}}
\newcommand{\norm}[1]{\left\lVert #1 \right\rVert}

\usepackage{amsthm}
\usepackage{aliascnt}
\usepackage{xcolor}
\usepackage[framemethod=TikZ]{mdframed}

\makeatletter
\@ifpackageloaded{hyperref}{%
  \hypersetup{
    pagebackref=true,
    breaklinks=true,
    colorlinks=true,
    allcolors=cvprblue
  }%
}{%
  \usepackage[pagebackref,breaklinks,colorlinks,allcolors=cvprblue]{hyperref}%
}
\@ifpackageloaded{cleveref}{}{%
  \usepackage[nameinlink,capitalize]{cleveref}%
}
\makeatother

\definecolor{lightgraybox}{rgb}{0.96,0.96,0.96}

\mdfdefinestyle{lighttheobox}{
  backgroundcolor=lightgraybox,
  linecolor=black,
  linewidth=0.4pt,
  roundcorner=3pt,
  skipabove=8pt,
  skipbelow=8pt,
  innertopmargin=-1pt,
  innerbottommargin=6pt,
  innerleftmargin=8pt,
  innerrightmargin=8pt
}

\newtheorem{theorem}{Theorem}[section]

\newaliascnt{proposition}{theorem}
\newtheorem{proposition}[proposition]{Proposition}
\aliascntresetthe{proposition}

\newaliascnt{lemma}{theorem}

\aliascntresetthe{lemma}

\newaliascnt{assumption}{theorem}

\aliascntresetthe{assumption}

\crefname{theorem}{Theorem}{Theorems}
\Crefname{theorem}{Theorem}{Theorems}
\crefname{proposition}{Proposition}{Propositions}
\Crefname{proposition}{Proposition}{Propositions}
\crefname{lemma}{Lemma}{Lemmas}
\Crefname{lemma}{Lemma}{Lemmas}
\crefname{assumption}{Assumption}{Assumptions}
\Crefname{assumption}{Assumption}{Assumptions}

\surroundwithmdframed[style=lighttheobox]{theorem}
\surroundwithmdframed[style=lighttheobox]{proposition}
\surroundwithmdframed[style=lighttheobox]{lemma}
\surroundwithmdframed[style=lighttheobox]{assumption}
\usepackage[accsupp]{axessibility}

\definecolor{cvprblue}{rgb}{0.21,0.49,0.74}
\usepackage[pagebackref,breaklinks,colorlinks,allcolors=cvprblue]{hyperref}

\def\paperID{10774} 
\def\confName{CVPR}
\def\confYear{2025}

\title{Designing Instance-Level Sampling Schedules via REINFORCE with James-Stein Shrinkage}

\author{Peiyu Yu$^{1,3}$\footnote{Work done during an internship at Google}, Suraj Kothawade$^{1}$, Sirui Xie$^{2}$, Ying Nian Wu$^{3}$, Hongliang Fei$^{1}$
\\
$^1$Google, $^2$Google DeepMind, $^3$UCLA\\
{\tt\small yupeiyu98@g.ucla.edu; ywu@stat.ucla.edu; \{skothawade,siruix,hongliangfei\}@google.com}
}

\begin{document}
\twocolumn[{%
\maketitle
\centering
\vspace{-12pt}

\begin{tabular*}{\textwidth}{@{\extracolsep{\fill}}p{.49\textwidth}p{.49\textwidth}@{}}
\panel{.49\textwidth}{\textit{a}}{\textit{Few-Step Sampling}}
{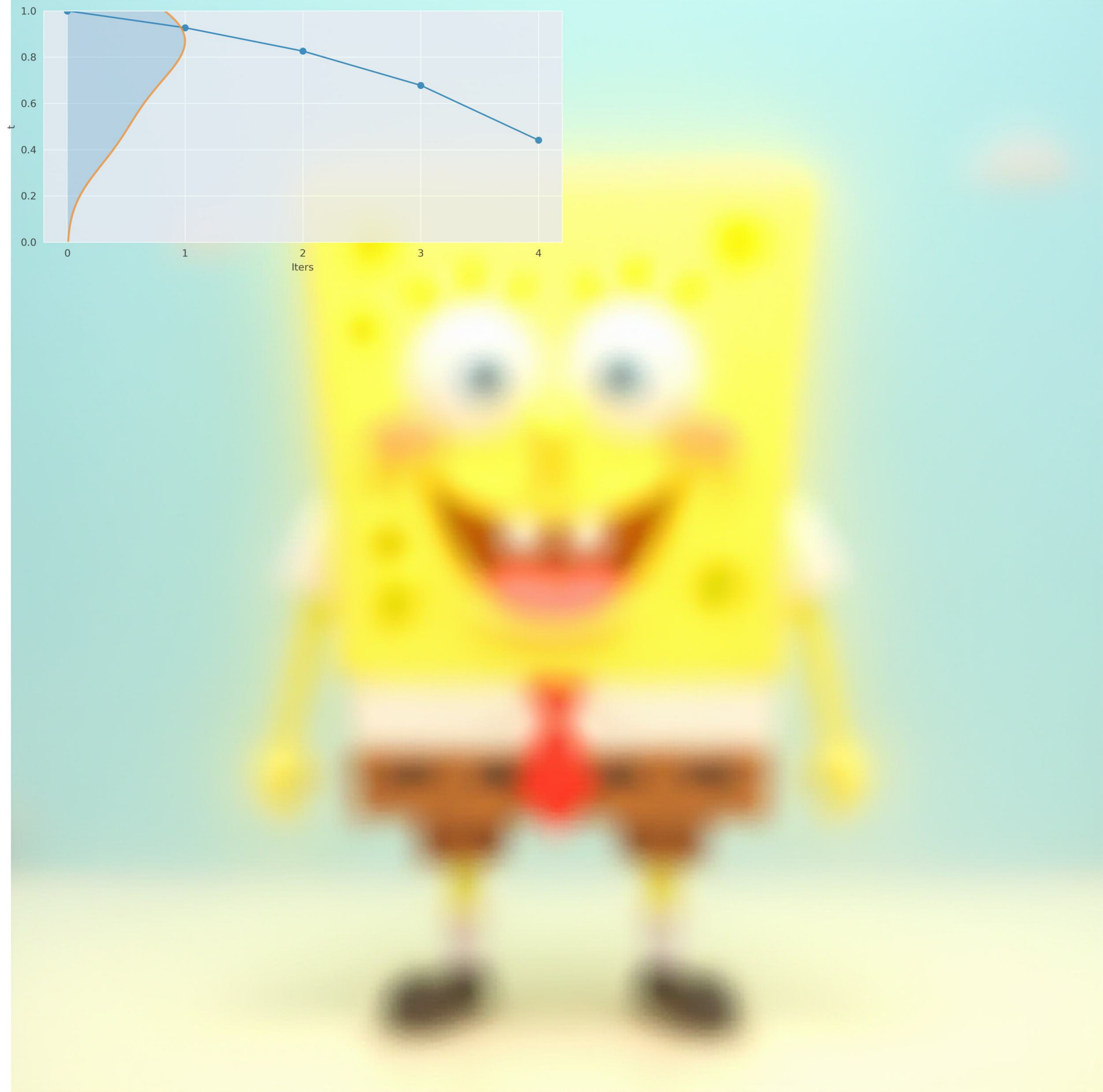}
{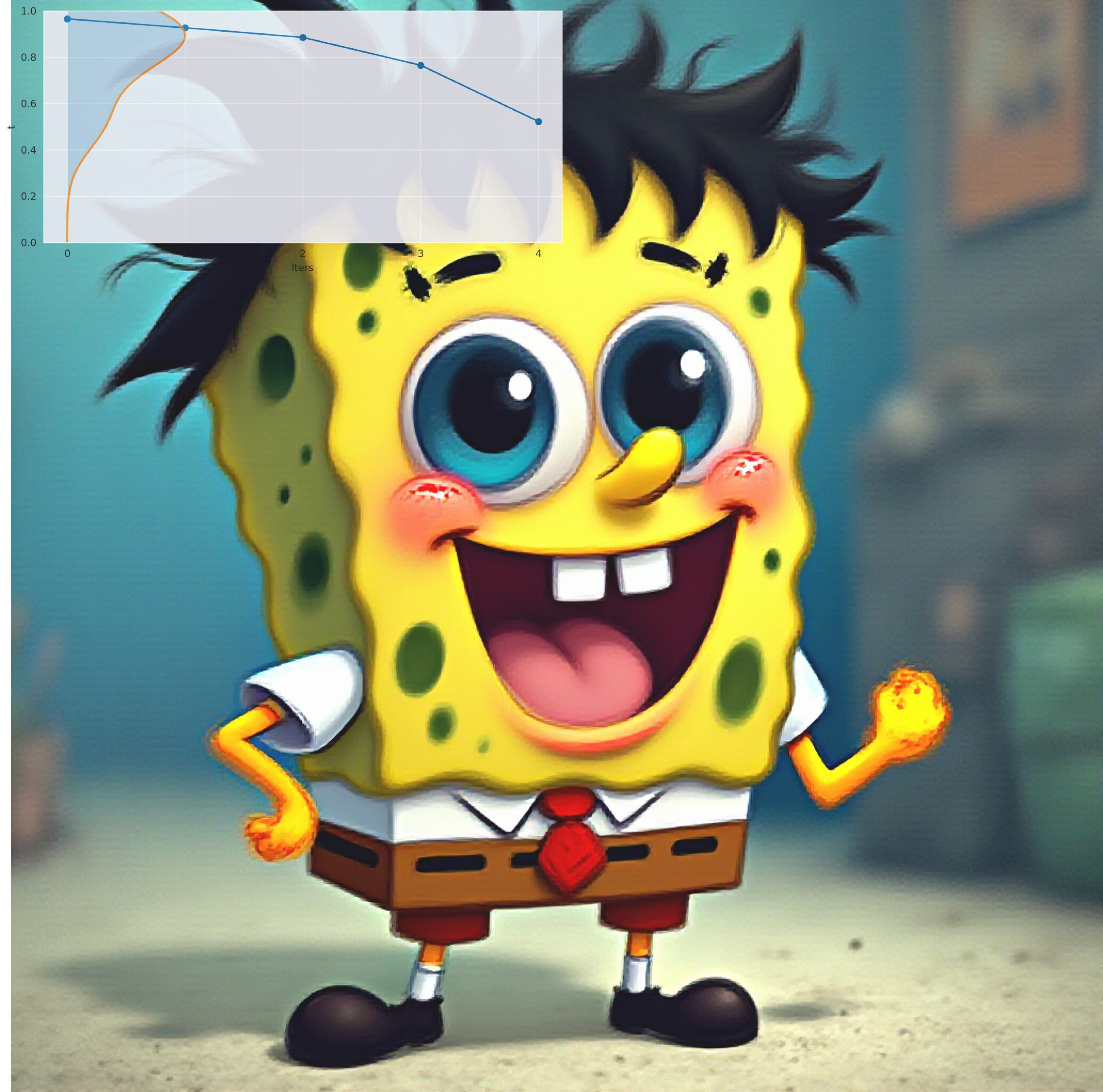}
{\textit{Spongebob depicted in the {\color{red}style of Dragon Ball Z}.}}
&
\panel{.49\textwidth}{\textit{b}}{\textit{Text-Image Alignment}}
{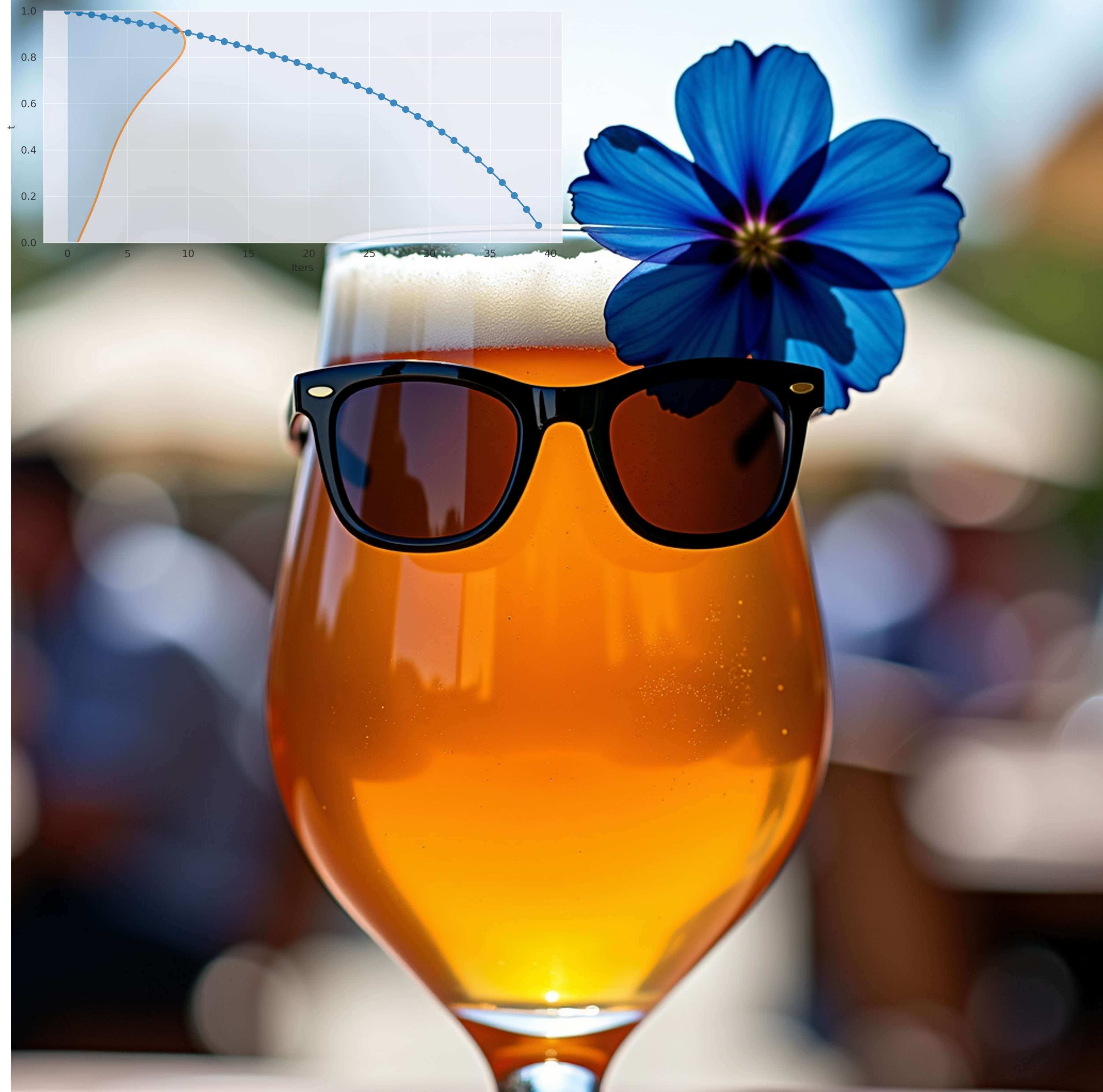}
{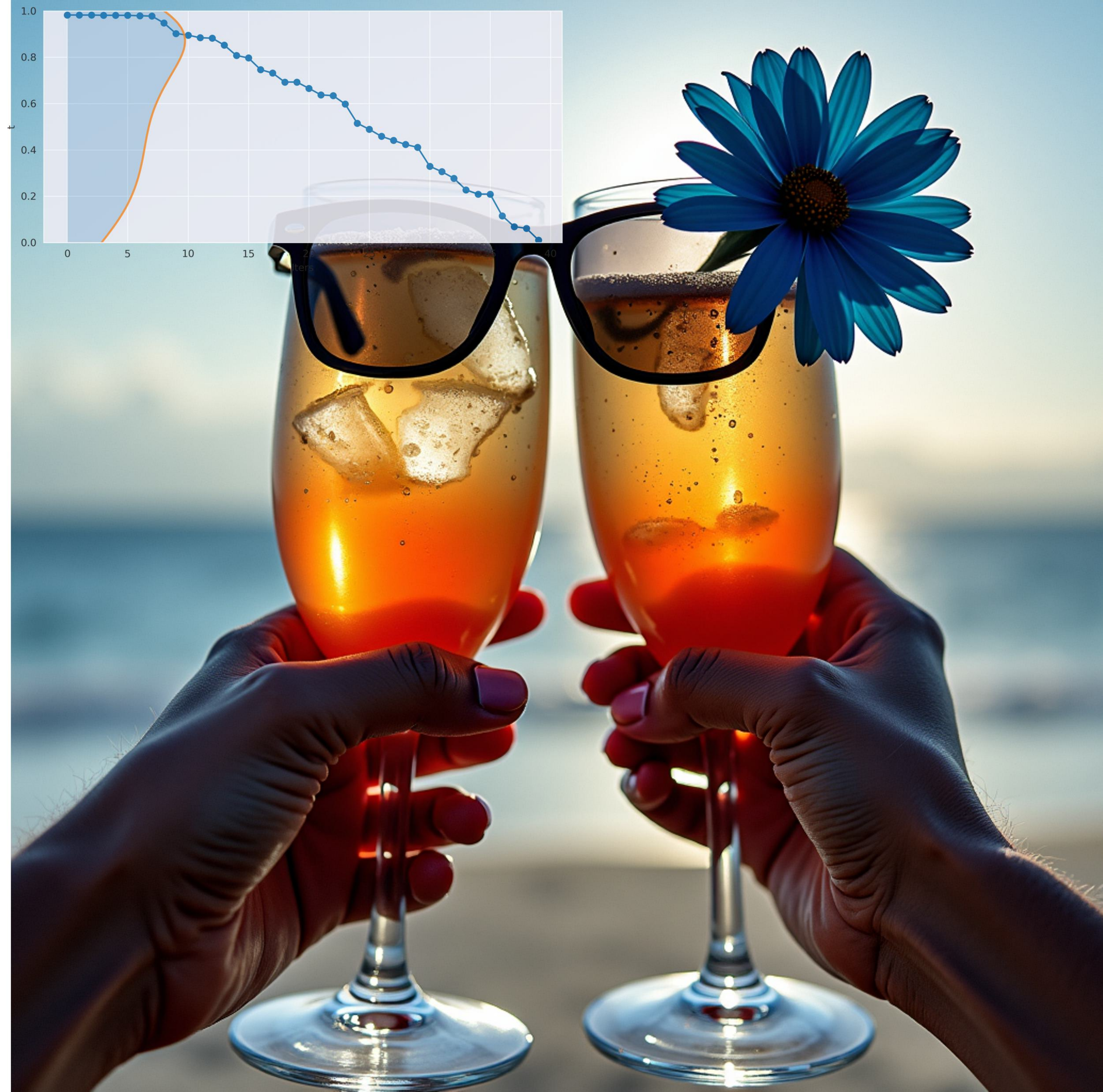}
{\textit{A {\color{red}toast} with black sunglasses and a blue flower on the top-right corner.}}
\\ \addlinespace[4pt]
\panel{.49\textwidth}{\textit{c}}{\textit{Text Rendering}}
{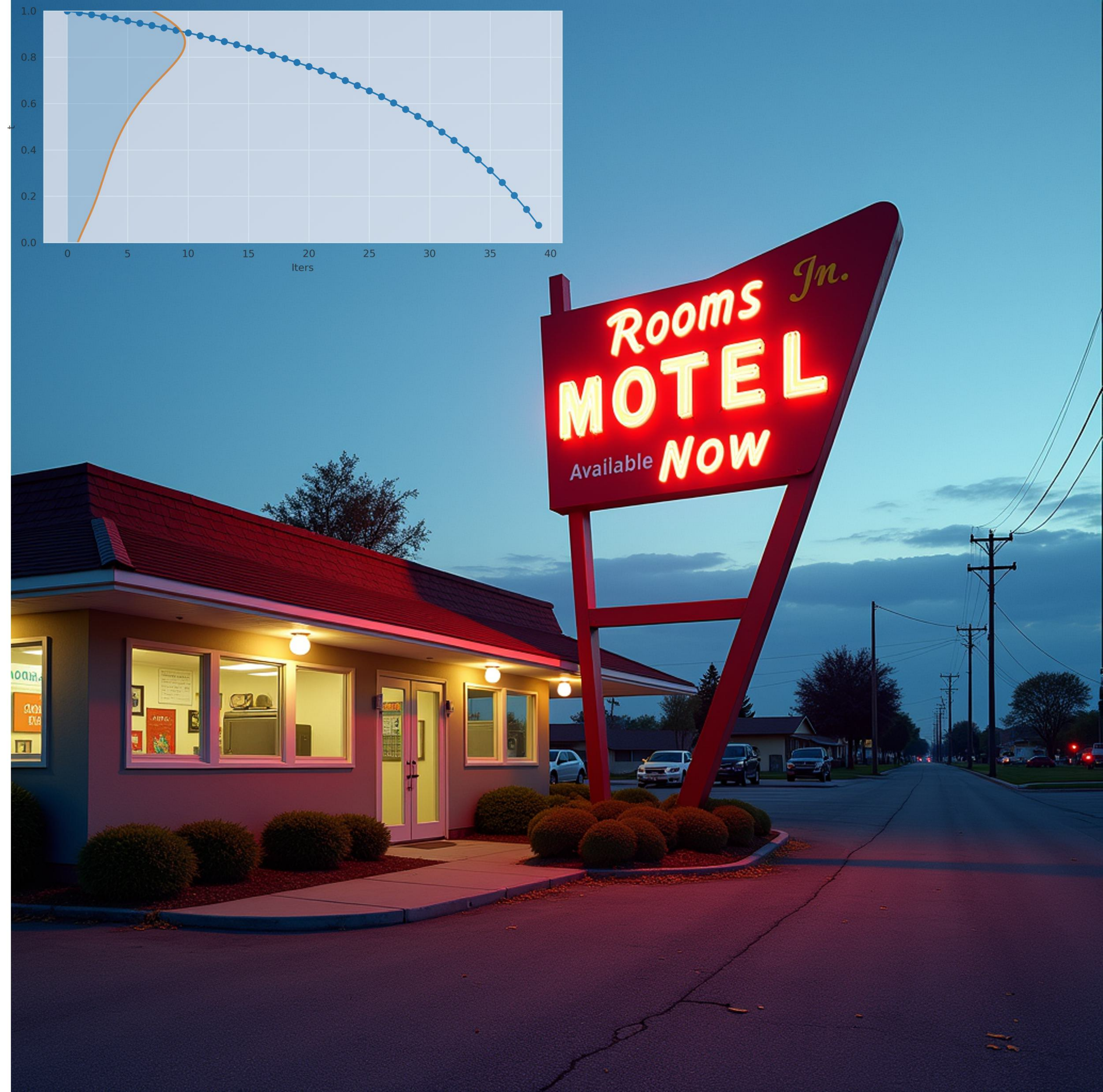}
{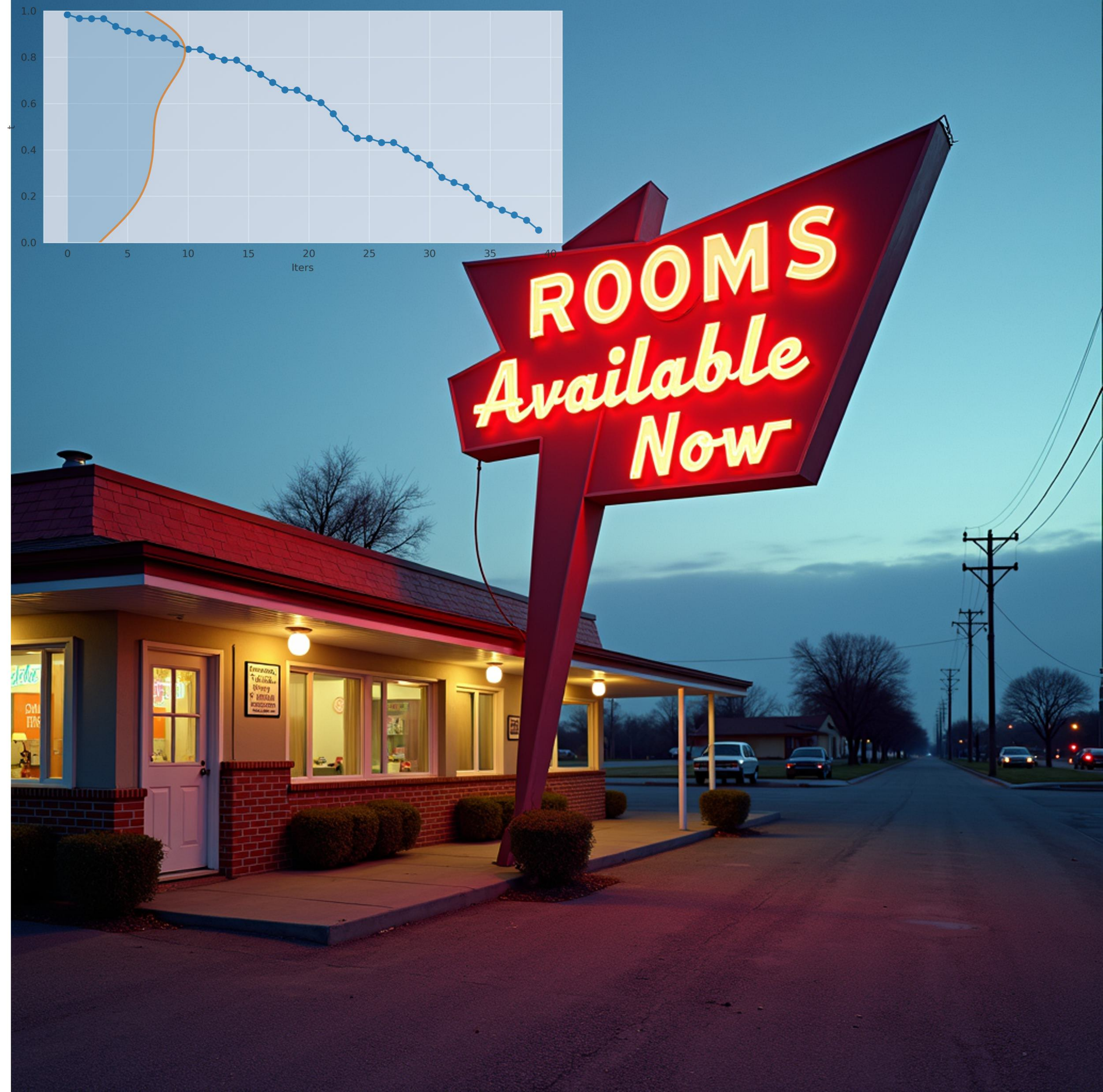}
{\textit{A vintage 1950s motel...{\color{red} neon sign blinking ``Rooms Aailable Now''}...}}
&
\panel{.49\textwidth}{\textit{d}}{\textit{Counting}}
{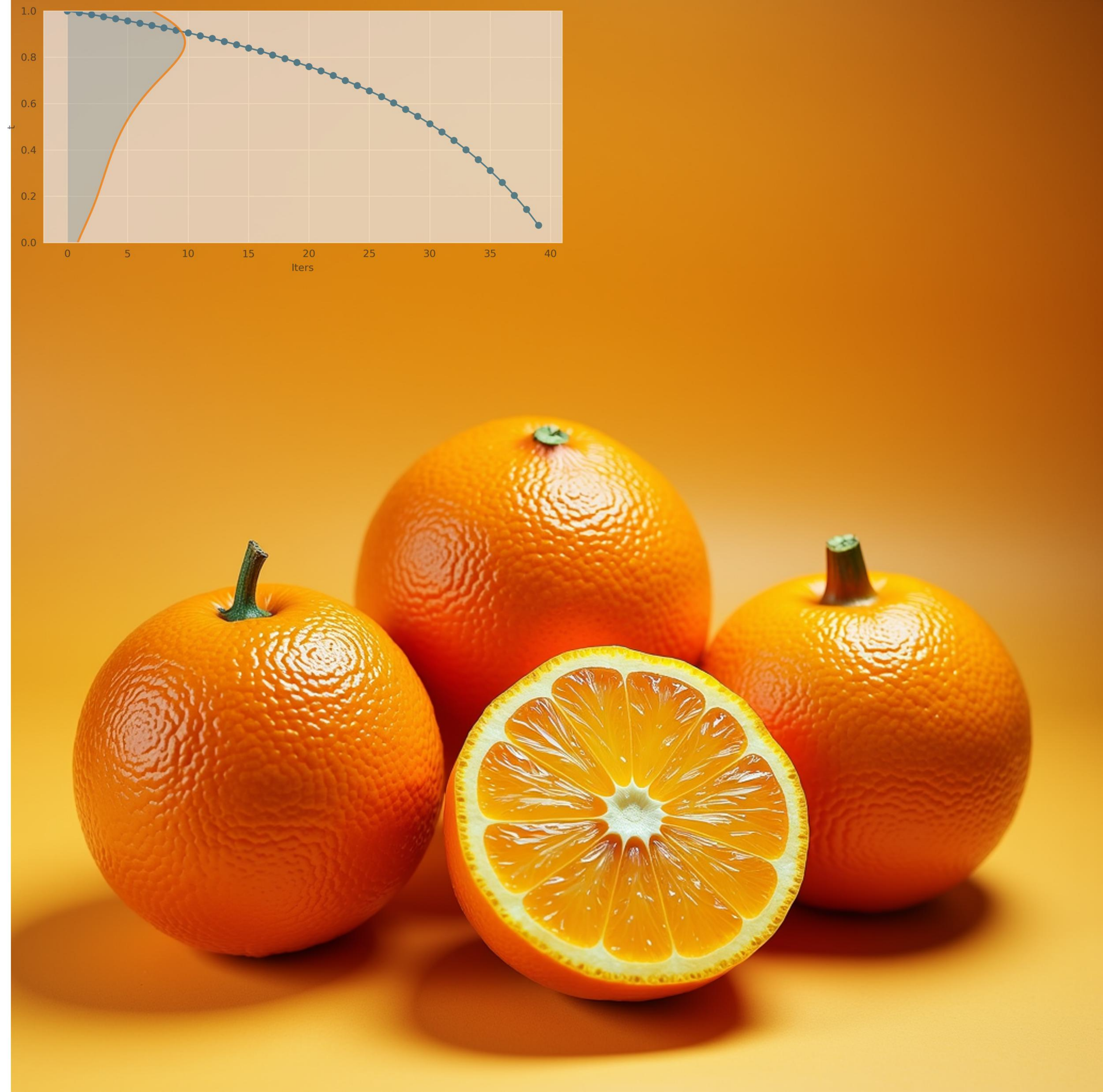}
{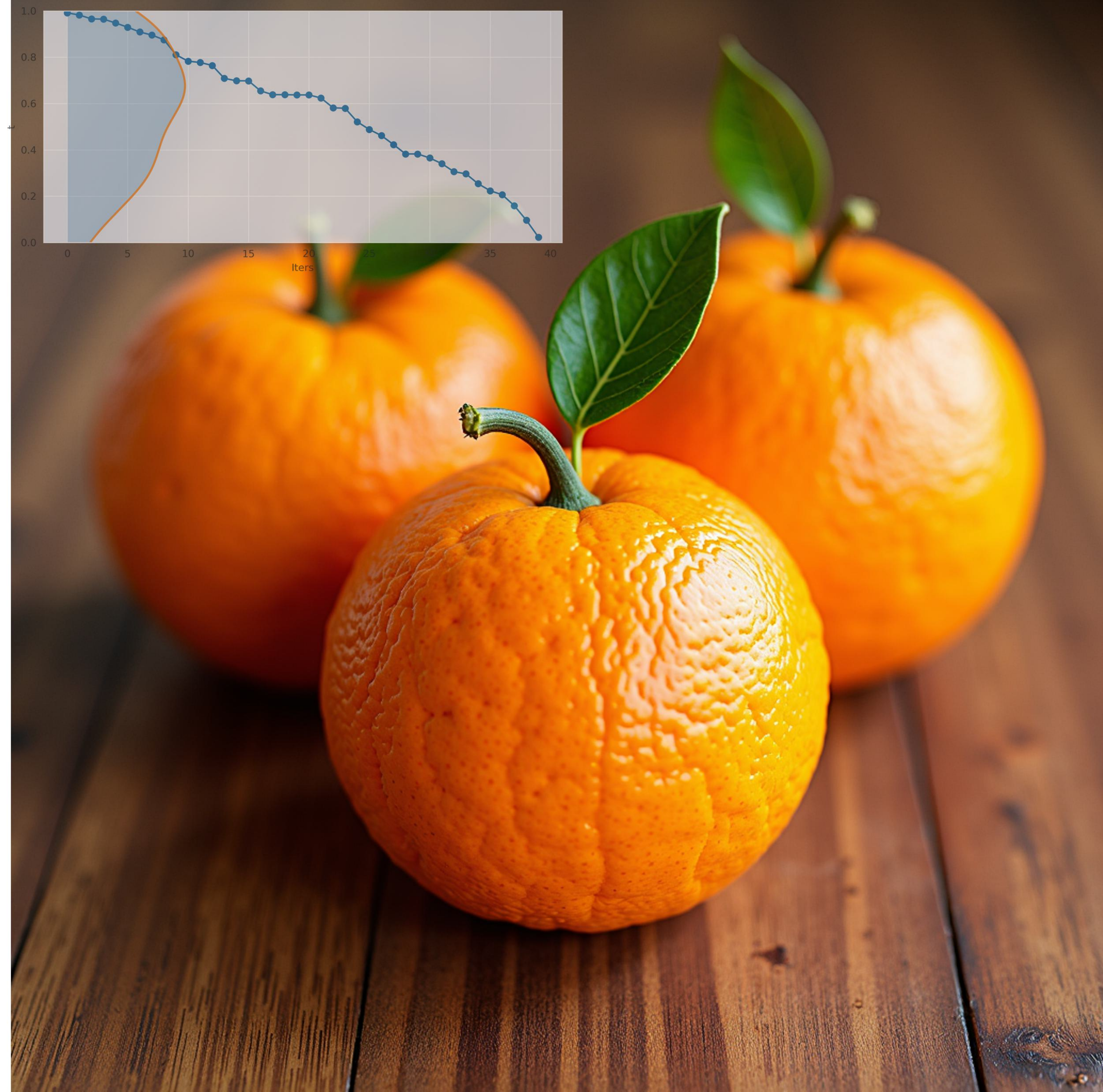}
{\textit{A photo of {\color{red}three} oranges}.}
\end{tabular*}
\vspace{-7pt}
\captionof{figure}{\textbf{Instance-level schedules improve text-to-image generation.} \textbf{(a)-(d)} illustrate four aspects where pretrained models like Flux-Dev benefit from our schedules. Samplers using our schedules (right) show consistent improvements over those with the default schedule (left), which are even more pronounced at only 5 inference steps (a). We visualize the schedules at the top left corner of each image; X-axis denotes the number of inference steps, and Y-axis denotes the actual timestep values. We plot the KDE of t values along the Y-axis.}
\label{fig:teaser}
\vspace{7pt} 
}]
\renewcommand{\thefootnote}{}
\footnotetext{$\ast$ Work done during an internship at Google.}
\begin{abstract}
Most post-training methods for text-to-image samplers focus on model weights: either fine-tuning the backbone for alignment or distilling it for few-step efficiency.
We take a different route: \emph{rescheduling} the sampling timeline of a frozen sampler.
Instead of a fixed, global schedule, we learn instance-level (prompt- and noise-conditioned) schedules through a single-pass Dirichlet policy. 
To ensure accurate gradient estimates in high-dimensional policy learning, we introduce a novel reward baseline based on a principled James–Stein estimator; it provably achieves lower estimation errors than commonly used variants and leads to superior performance. Our rescheduled samplers consistently improve text–image alignment including text rendering and compositional control across modern Stable Diffusion and Flux model families. Additionally, a 5-step 
Flux-Dev sampler with our schedules can attain generation quality comparable to deliberately distilled samplers like Flux-Schnell. We thus position our scheduling framework as an emerging model-agnostic post-training lever that unlocks additional generative potential in pretrained samplers.
\end{abstract}
    
\vspace{-10pt}
\section{Introduction}
\label{sec:intro}
Diffusion and flow-matching models have achieved remarkable progress in prompt-based generation. By iteratively transforming an initial noise sample into a final image through multi-step denoising or probability flow trajectories, modern samplers demonstrate not only outstanding image fidelity but also impressive adherence to textual prompts~\cite{rombach2022high,esser2024scaling,flux2024}. At inference time, however, generation quality depends critically on the sampling schedule: how a limited step budget is allocated along the continuous sampling trajectory. Most production backbones adopt a single, globally fixed schedule for all inputs; while convenient, such a universal schedule is unlikely to be optimal across the diverse range of prompts encountered at test time.

Intuitively, different prompts demand different spatial or semantic details, and distinct noise seeds may induce varying initial conditions that affect image quality. Recent analyses of ``golden noise''~\cite{zhou2025golden} further highlight such seed sensitivity, reinforcing the need for \textbf{\textit{instance-level scheduling}}. We therefore study schedule learning: adapting the timestep allocation per prompt and seed while keeping the total step budget fixed, to further exploit the generative potential of modern samplers. Prior work has begun to explore adjacent directions. One contemporaneous approach, TPDM~\cite{ye2025schedule}, proposes the next timestep \textit{autoregressively} from internal latents and current time. While TPDM primarily aims at early stopping and variable-length sampling to improve efficiency, our focus differs: we care about the Pareto frontier of the quality–vs.–steps tradeoff; we study reallocation of a fixed sampling budget to reach for the best possible generative quality per instance. A side-by-side comparison against a single-pass TPDM-style baseline under our settings further illustrates the distinctions (see \cref{tab:hpdt2i_steps}). 

\begin{figure}[!ht]
\centering

\begin{subfigure}{0.9\linewidth}
  \centering
  \includegraphics[width=\linewidth]{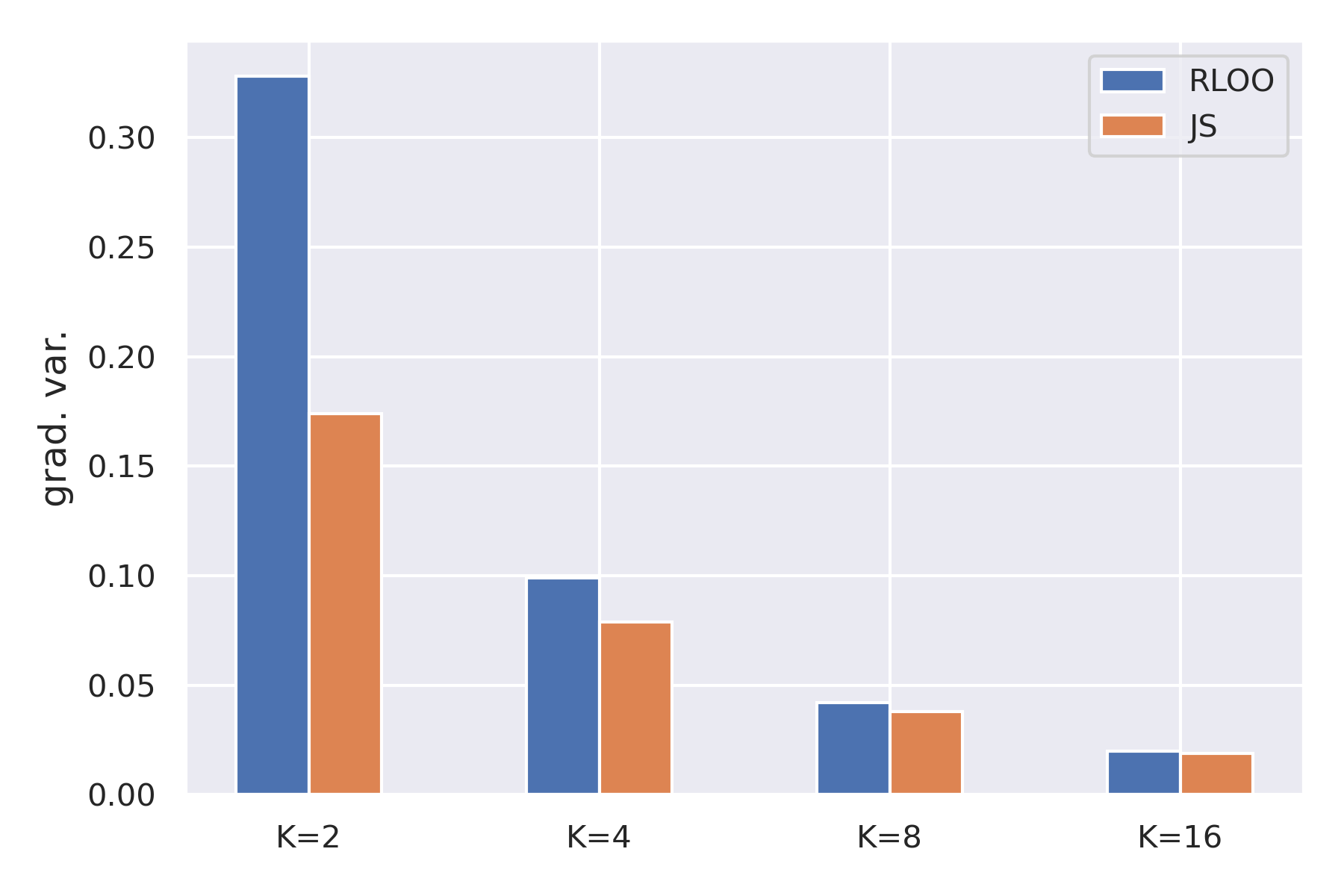}
  \caption{Simulated gradient estimation errors.}
  \label{fig:js_variance}
\end{subfigure}

\vspace{8mm} 

\begin{subfigure}{0.9\linewidth}
  \centering
  \begin{tikzpicture}[
    >=Latex, font=\small,
    node distance=9mm and 12mm,
    eq/.style={font=\scriptsize, inner sep=0pt, outer sep=0pt},
    tight/.style={inner sep=0pt, outer sep=0pt},
    sumop/.style={draw,circle,inner sep=0pt,minimum size=3.8mm, line width=0.4pt, fill=white}
  ]

  \node[eq] at (1.5, 1.00) {within-context};
  \node[eq] at (1.5,-1.05) {cross-context};

  \node[eq] (batch) at (0,0) {Batch $\{c_i\}_{i=1}^B$};

  \node[eq, above right=4mm and 5mm of batch] (c_j)  {$c_j$};
  \node[eq, below right=4mm and 5mm of batch] (c_nj) {$\{c_{-j}\}$};

  \draw[->] (batch.east) -- ++(2mm,0) |- (c_j.west);
  \draw[->] (batch.east) -- ++(2mm,0) |- (c_nj.west);

  \node[eq, right=27.5mm of c_j] (rloo) {$b_{\mathrm{RLOO}}$};
  \node[eq, above=1.2mm of rloo] {Eq.~\eqref{eq:rloo}};
  \draw[->] (c_j) -- node[eq, above]{$\sim \{r^{(j,k)}\}_{k=1}^{K}$} (rloo);

  \node[eq, right=23mm of c_nj] (xctx) {$b_{\mathrm{xctx}}$};
  \node[eq, below=1.2mm of xctx] {Eq.~\eqref{eq:xctx_rloo}};
  \draw[->] (c_nj) -- node[eq, above]{$\sim \{r^{(-j,k)}\}_{k=1}^{K}$} (xctx);

  \node[sumop, right=5mm of rloo] (sum) {$+$};
  \draw[->] (rloo) -- (sum);

  \draw[->, shorten >=1pt] (xctx.east) -- ($(sum.south |- xctx.east)$) -- (sum.south);

  \node[eq, right=5mm of sum] (js) {$b_{\mathrm{JS}}$};
  \node[eq, above=0mm of js] {Eq.~\eqref{eq:js-baseline}};
  \draw[->] (sum) -- (js);

  \end{tikzpicture}
  \vspace{1mm}
  \caption{James-Stein baseline workflow.}
  \label{fig:js_diagram_panel}
\end{subfigure}

\caption{\textbf{James--Stein (JS) reward baseline.}
\subref{fig:js_variance} Simulation results from anlaytical policies showcasing JS is consistently better than RLOO for different number of rollouts $K$. More details in supplemental materials.
\subref{fig:js_diagram_panel} Diagram of JS baseline: combining $b_{\mathrm{RLOO}}$ and $b_{\mathrm{xctx}}$ into
$b_{\mathrm{JS}} = \alpha_c b_{\mathrm{RLOO}} + (1-\alpha_c) b_{\mathrm{xctx}}$ (Eq.~\eqref{eq:js-baseline}).}
\label{fig:js_diagram}
\end{figure}
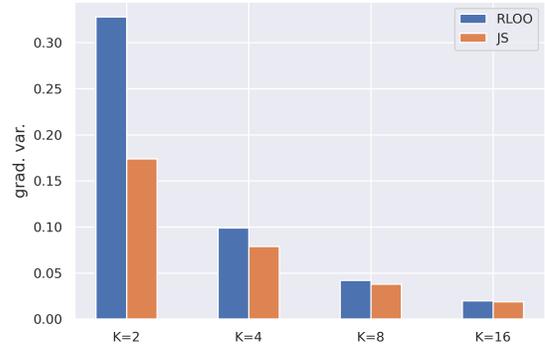
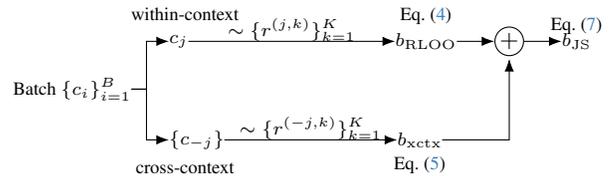

Specifically, in this paper we consider scheduler design as learning a non-autoregressive \textit{single-pass policy} that directly outputs the entire sampling schedule, with policy gradient. Unlike autoregressive next-step-prediction policies whose test-time overhead scales linearly with the number of timesteps, this design choice amortizes the cost entirely, producing instance-specific schedules with constant overhead. However, this advantage comes with a challenge. General policy gradient estimators in high dimensional spaces like our settings may induce substantial variances. To overcome this issue, we introduce a principled \textit{\textbf{James--Stein (JS) reward baseline}}: a control variate that adaptively shrinks each reward toward a batch-wise global mean; it guarantees strictly lower gradient estimation errors than popular alternatives, including REINFORCE Leave-One-Out (RLOO) \cite{kool2019buy,ahmadian2024back} (see \cref{fig:js_diagram}). The JS baseline is simple to implement, incurs negligible extra computation, and can serve as a drop-in replacement for standard baselines in generic policy-gradient training. Empirically, when the sampling budget is generous, our rescheduling approach further improves general text-to-image alignment, text rendering, and fine-grained capabilities such as counting. Under tight budgets, down to as few as 5 steps, remarkably, our framework achieves generation quality comparable to deliberately distilled samplers. It therefore also offers a complementary and compelling path toward efficient few-step generation orthogonal to distillation-based methods. We summarize our \textbf{contributions} as follows:
\begin{itemize}
    \item \textbf{Instance-level scheduler framework.} We formulate a single-shot policy that outputs the entire denoising schedule per input condition. Our design choice removes completely the linear cost of stepwise proposals and successfully performs credit assignment under terminal rewards.
    \item \textbf{JS baseline as agnostic var-reduction primitive.} We propose a principled, lightweight and unbiased variance-reduction baseline for policy-gradient training that adaptively shrinks rewards across contexts, substantially stabilizing REINFORCE in long-horizon one-shot settings.
    \item \textbf{Rescheduling as new post-training lever.} Our text-conditioned scheduler improves general and fine-grained text-image preference alignment under same budgets, even achieving competitive visual quality with 5 steps comparable to deliberately distilled samplers, with negligible inference overhead.
\end{itemize}

\section{Related Works}

\paragraph{T2I samplers and inference-time schedules.}
Modern text-to-image diffusion and flow-matching models achieve high-fidelity synthesis by iteratively denoising with globally defined timestep schedules \cite{rombach2022high,liu2022flow,lipman2022flow,esser2024scaling,flux2024}. Considerable progress has been made on solver design and handcrafted timestep allocations (e.g., ODE solvers and higher-order variants) to improve efficiency and stability under limited step budgets \cite{song2020denoising,lu2022dpm,zhao2023unipc,karras2022elucidating,sabour2024align}. However, these advances retain a \emph{one-size-fits-all} schedule agnostic to instance-specific trajectory geometry. We instead learn a prompt-conditional schedule as a policy. Recent work such as TPDM \cite{ye2025schedule} also explores RL-based optimization of diffusion sampler, but it autoregressively predicts next timesteps during sampling, aiming at early stopping and requiring $O(L)$ policy calls for inference; $L$ denotes the inference steps. In contrast, our method issues a \emph{single-pass} schedule proposal for the Pareto frontier of the quality–vs.–steps tradeoff, by leveraging a principled James–Stein control variate to facilitate training. Under the same compute budget, our design yields significantly stronger results than the previous counterpart.

\paragraph{Few-step sampling and distillation.}
Another active direction compresses long-run samplers into few-step generators through progressive or consistency-style distillation \cite{salimans2022progressive,song2023consistency,yin2024one,xie2024distillation,yunoisier}. While highly effective, these methods retrain or distill the backbone often at additional training cost. We position our approach as an orthogonal yet compelling path: we do not retrain the backbone but reallocate an existing step budget via learned scheduling. Empirically, we find that simple rescheduling already unlocks strong few-step performance; pretrained samplers possess innate few-step capabilities even without distillation.

\paragraph{RL for generative alignment and variance reduction.}
Reinforcement learning has recently emerged as a versatile paradigm for aligning generative models with human or task-specific objectives. In text-to-image diffusion, this idea has been realized through policy-gradient and preference-based optimization frameworks such as DDPO, DPOK, and Diffusion-DPO \cite{fan2023dpok,black2023training,wallace2024diffusion}. These methods fine-tune pretrained generators toward reward signals that capture aesthetic quality, faithfulness, or compositional alignment, often by treating the denoising latents throughout the generative process as a sequence of actions. 

We adopt a complementary view: the \emph{entire schedule} is treated as a high dimensional, open-loop and forward-looking plan proposed in one pass. This formulation exposes a key challenge common to sequence-level policy gradients—the high variance of gradient estimators. Classical variance-reduction techniques rely on hand-crafted heuristics or learned baselines \cite{williams1992simple,kool2019buy,ahmadian2024back}. For instance, RLOO \cite{kool2019buy} draws multiple samples per context and uses the average reward as a plug-in baseline. Although empirically effective, it remains brittle under heterogeneous reward scales, as we illustrate in \cref{sec:exps}. To address this, we first show that the RLOO baseline can be interpreted as a Monte Carlo estimate of the variance-optimal reward baseline. We then introduce a James--Stein shrinkage baseline that \emph{provably always} reduces estimator variance beyond RLOO and effectively facilitates policy learning in high-dimensional spaces. Empirically, this yields stronger T2I samplers under same compute budgets; it demonstrates the potential of principled variance reduction for general policy optimization.
\section{Method}
\label{sec:method}

We begin by formulating our objective: to learn an instance-adaptive scheduler. We cast this as a policy learning problem, where the policy, i.e., our scheduler, proposes the entire schedule in one pass. To reduce gradient estimation errors and make policy learning practical for high dimensional space of this kind, we (i) revisit the crucial role of the baseline in variance reduction and derive the theoretical variance–optimal baseline, (ii) reinterpret the commonly used per-context RLOO baseline as its empirical Monte Carlo approximation, and (iii) further improve upon it through the lens of the James–Stein theorem \cite{james1961estimation}, introducing a James–Stein (JS) shrinkage baseline that \emph{provably} achieves lower variance and estimation error than RLOO under mild assumptions. Finally, we instantiate the policy as a Dirichlet scheduler that outputs a timestep allocation, enabling flexible and instance-specific sampling without modifying the pretrained diffusion backbone.

\subsection{Problem Formulation}
\label{subsec:problem-setup}
Let $\mathcal{S}_{\phi}$ denote a pretrained sampler that maps an initial noise tensor $\mathbf{x}_T$ to an image $\mathbf{x}_0$ given a text condition $\mathbf{c}$. A \emph{schedule} is defined as a strictly decreasing sequence of $L$ normalized timesteps
\[
\tau = (t_1, t_2, \ldots, t_L), \quad 1 >= T = t_1 > \cdots > t_L >= 0,
\]
which specifies the noise levels the sampler traverses. Different schedule lengths correspond to different computational budgets and yield varying image quality.

For a given context $(\mathbf{x}_T, \mathbf{c})$, we parameterize a stochastic \emph{schedule policy}
$\pi_{\theta}(\tau \mid \mathbf{x}_T, \mathbf{c})$
that samples schedules conditioned on both the text and noise. Executing $\tau$ through $\mathcal{S}_{\phi}$ produces a final image $\mathbf{x}_0$ and an associated scalar reward $r(\mathbf{x}_0(\tau); \mathbf{c})$, such as a CLIP-based alignment score or a learned preference-model output. The objective is to maximize the expected reward:
\begin{equation}
J(\theta)
= \mathbb{E}_{\tau \sim \pi_\theta(\cdot \mid \mathbf{x}_T, \mathbf{c})}
  \big[\, r(\mathbf{x}_0(\tau); \mathbf{c}) \,\big].
\label{equ:max_rew}
\end{equation}

Our goal is to optimize the policy parameters $\theta$ to maximize $J(\theta)$, enabling the model to learn instance-specific schedules that adapt to diverse text and noise conditions without modifying the frozen sampler backbone.


\subsection{Policy Gradient with Baselines}
\label{subsec:policy-gradient}

\paragraph{Score-function estimator.}
To optimize the expected reward in Eq.~\eqref{equ:max_rew}, we apply the standard REINFORCE or score-function estimator:
\begin{equation}
\nabla_\theta J(\theta)
= \E_{\tau \sim \pi_\theta}
  \big[ (r(\tau) - b)\,
  \nabla_\theta \log \pi_\theta(\tau \mid \mathbf{x}_T, \mathbf{c}) \big],
\label{eq:reinforce}
\end{equation}
where $b$ is any baseline independent of $\theta$.  
The baseline leaves the estimator unbiased while allowing variance reduction—a critical consideration in our single-pass setting.

\paragraph{Variance–optimal baseline.}
Among all unbiased estimators of the form in \eqref{eq:reinforce}, there exists a theoretically optimal choice of $b$ that minimizes the estimator’s variance.

\begin{proposition}[Variance–optimal baseline]
\label{prop:opt-baseline}
For the gradient estimator in \eqref{eq:reinforce}, the baseline that minimizes
$\Var[(r - b)\nabla_\theta \log \pi_\theta(\tau)]$
is
\[
b^{*}
= \frac{\E\!\left[r(\tau)\,\|\nabla_\theta \log \pi_\theta(\tau)\|^2\right]}
       {\E\!\left[\|\nabla_\theta \log \pi_\theta(\tau)\|^2\right]}.
\]
\textit{Sketch.} Differentiate the variance w.r.t. $b$ and set the derivative to zero. See supplemental materials for a detailed proof.
\end{proposition}

In practice, computing $b^{*}$ exactly is infeasible for high-dimensional policies.  
However, when the policy is nearly deterministic (i.e., $\nabla_\theta \log \pi_\theta$ varies little across samples within a context), the numerator and denominator decouple, yielding the approximation
\begin{equation}
\label{eq:approx_opt_b}
b^{*} \approx \E_{\tau \sim \pi_\theta}[r(\tau) \mid \mathbf{x}_T, \mathbf{c}],
\end{equation}
which corresponds to the \emph{contextual mean reward}.  
This motivates empirical estimators that approximate this conditional expectation from sampled rewards.

\paragraph{Per-context RLOO baseline.}
A common design choice for instantiating the baseline $b$ in \eqref{eq:reinforce} is the
\emph{per-context} RLOO estimator \cite{kool2019buy,ahmadian2024back}, which can be viewed as a simple Monte Carlo approximation to the variance–optimal baseline $b^{*}$.
In our problem setup, we collect a mini-batch of $B$ independent contexts
$\{(\mathbf{x}_T^{(c)}, \mathbf{c}^{(c)})\}_{c=1}^B$.
For each context $c$, we draw $K_c$ schedule samples
$\{\tau^{(c,i)}\}_{i=1}^{K_c}$ with corresponding rewards
$\{r^{(c,i)}\}_{i=1}^{K_c}$.
The RLOO baseline estimates the contextual mean reward as
\begin{equation}
b_{\mathrm{RLOO}}^{(c,i)} = \frac{1}{K_c - 1} 
\sum_{\substack{j=1,j \neq i}}^{K_c} r^{(c,j)}.
\label{eq:rloo}
\end{equation}
This estimator remains unbiased because $b_{\mathrm{RLOO}}^{(c,i)}$ is independent of the current rollout $i$, while removing shared contextual effects from the prompt/noise pair $(\mathbf{x}_T^{(c)}, \mathbf{c}^{(c)})$. $b_{\mathrm{RLOO}}$ RLOO therefore calculates the contextual mean reward, as a straightforward, within-context Monte Carlo approximation to $b^{*}$.  
However, when the number of rollouts per context $K_c$ is small (e.g., 2–4) or when reward scales vary substantially across prompts, $b_{\mathrm{RLOO}}$ becomes a noisy estimator of the contextual mean (see \cref{sec:exps} and \cref{fig:js_diagram}).

\paragraph{Cross-context baseline.}
A natural extension is to exploit reward information \emph{across} contexts by forming a shared baseline for the entire mini-batch:
\begin{equation}
b_{\mathrm{xctx}}^{(c,i)}
= \frac{1}{(\sum_{c'} K_{c'}) - 1}
  \Bigg(
  \sum_{c'=1}^{B}\sum_{j=1}^{K_{c'}} r^{(c',j)} - r^{(c,i)}
  \Bigg),
\label{eq:xctx_rloo}
\end{equation}
which averages rewards over all rollouts except the current one.  
Like $b_{\mathrm{RLOO}}$, this leave-one-out construction preserves the unbiasedness of the REINFORCE estimator, but now draws upon the entire batch to estimate the mean reward.  
Such pooling can substantially reduce estimator variance when reward magnitudes are comparable across prompts, effectively anchoring each contextual baseline to a global reference.  
However, this naive global averaging ignores systematic differences in reward scales across contexts and thus may over- or under-compensate for certain prompts. We next introduce a principled \emph{shrinkage} formulation that interpolates between the two extremes—context-wise RLOO and fully pooled cross-context baselines—providing a balanced use of shared information that provably improves approximation to the variance-optimal baseline $b^{*}$.


\subsection{James--Stein Shrinkage Reward Baseline}
\label{subsec:js}
We now introduce a principled shrinkage formulation that unifies the
per-context baseline $b_{\mathrm{RLOO}}$ and the cross-context baseline $b_{\mathrm{xctx}}$.
The key idea is to treat contextual rewards as random effects drawn from a shared distribution, and to shrink each context’s baseline toward the batch-level anchor in a data-dependent manner.

\paragraph{Random effects view.}
We model the observed rewards with a two-level random effects hierarchy:
\begin{equation}
\begin{aligned}
r^{(c,i)} &= \mu_c + \varepsilon^{(c,i)}, &
\varepsilon^{(c,i)} &\sim \mathcal{N}(0, \sigma^2), \\
\mu_c &= \mu_0 + \xi^{(c)}, &
\xi^{(c)} &\sim \mathcal{N}(0, \delta^2),
\end{aligned}
\label{eq:random-effects}
\end{equation}
where $\sigma^2$ represents within-context reward variance and $\delta^2$ quantifies heterogeneity across contexts (e.g., varying prompt difficulty).  
For each context $c$, we denote by $\bar r_c = \tfrac{1}{K_c}\sum_i r^{(c,i)}$ its empirical mean reward.  
Under the model in \eqref{eq:random-effects}, the posterior mean of the latent mean reward $\mu_c$ is
\[
\E[\mu_c \mid \{r^{(c,i)}\}]
= (1 - \alpha_c^*)\,\bar r_c + \alpha_c^*\,\mu_0,
\quad
\alpha_c^* = \frac{\sigma^2 / K_c}{\sigma^2 / K_c + \delta^2},
\]
which adaptively shrinks $\bar r_c$ toward the global mean $\mu_0$.  
The shrinkage strength $\alpha_c^*$ increases when the number of rollouts per context $K_c$ is small or when contexts are relatively homogeneous (small~$\delta^2$).

\paragraph{Empirical James--Stein baseline.}
In practice, the variance components $(\sigma^2, \delta^2)$ are unknown.
Replacing them with empirical estimates gives rise to the
\emph{James--Stein (JS) reward baseline}, expressed directly as a convex interpolation between the previously defined
$b_{\mathrm{RLOO}}$ and $b_{\mathrm{xctx}}$:
\begin{equation}
b_{\mathrm{JS}}^{(c,i)}
= (1 - \widehat{\alpha}_c)\, b_{\mathrm{RLOO}}^{(c,i)}
  + \widehat{\alpha}_c\, b_{\mathrm{xctx}}^{(c,i)},
\label{eq:js-baseline}
\end{equation}
where both baselines are computed in a leave-one-out manner, excluding the same sample $(c,i)$.  
The shrinkage coefficient is estimated as
\begin{equation}
\widehat{\alpha}_c
= \frac{\widehat{\sigma}^2 / (K_c - 1)}
       {\widehat{\sigma}^2 / (K_c - 1) + \widehat{\delta}^2}.
\label{eq:alpha-hat}
\end{equation}
This formulation preserves the unbiasedness of the REINFORCE gradient while adaptively anchoring each context’s baseline to the global batch mean.
When $\widehat{\alpha}_c \!\to\! 0$, the estimator reduces to $b_{\mathrm{RLOO}}$;
when $\widehat{\alpha}_c \!\to\! 1$, it approaches $b_{\mathrm{xctx}}$.
Thus, $b_{\mathrm{JS}}$ provides a data-driven compromise that balances contextual specificity and cross-context regularization.

\paragraph{Estimating variance components.}
We estimate the within- and across-context variances directly from the two baseline statistics.
The within-context variance $\widehat{\sigma}^2$ is computed from the dispersion of individual rewards around $b_{\mathrm{RLOO}}$,
while the across-context variance $\widehat{\delta}^2$ is estimated via a \emph{method-of-moments (MoM)} correction that de-biases the between-context variance:
\begin{align}
\widehat{\sigma}^2
&= \frac{1}{\sum_c (K_c - 1)} \sum_{c=1}^{B}\sum_{i=1}^{K_c}
  (r^{(c,i)} - b_{\mathrm{RLOO}}^{(c,i)})^2, \\
\widehat{\delta}^2
&= \max\!\Big(0,\;
  \frac{1}{B-1}\sum_{c=1}^{B} 
  \big(\bar r_c - b_{\mathrm{xctx}}^{(c,\cdot)}\big)^2
  - \frac{\widehat{\sigma}^2}{\bar K}\Big),
\label{eq:var-est}
\end{align}
where $b_{\mathrm{xctx}}^{(c,\cdot)}$ is the cross-context leave-one-out mean for context $c$
and $\bar K = \tfrac{1}{B}\sum_c K_c$.
The subtraction of $\widehat{\sigma}^2 / \bar K$ corresponds to the MoM correction term
that removes within-context noise from the between-context estimate,
and the $\max(0,\cdot)$ term ensures non-negativity.
These quantities are re-estimated each iteration and add only negligible computational overhead. We next provide a theoretical justification of our JS baseline.

\paragraph{Empirical James-Stein is Bayesian optimal.}
With the configuration described above, we can now claim that the empirical James--Stein baseline has the following appealing properties under the listed assumptions:

\begin{theorem}[MSE improvement and empirical Bayes optimality]
\label{thm:js-mse}
Under the random effects model in Eq.~\eqref{eq:random-effects} and for $B \ge 3$ contexts:
\begin{enumerate}[label=(\roman*),leftmargin=*]
\item The James--Stein baseline strictly improves the mean-squared error over the unbiased RLOO baseline:
\[
\E\big[(b_{\mathrm{JS}}^{(c,i)} - \mu_c)^2\big]
< \E\big[(b_{\mathrm{RLOO}}^{(c,i)} - \mu_c)^2\big].
\]
\item $b_{\mathrm{JS}}^{(c,i)}$ coincides with the empirical Bayes posterior mean of $\mu_c$, and is therefore the MSE-optimal convex combination of $b_{\mathrm{RLOO}}^{(c,i)}$ and $b_{\mathrm{xctx}}^{(c,i)}$.
\end{enumerate}
\end{theorem}

\textbf{Remarks.}
Theorem~\ref{thm:js-mse} guarantees that the proposed baseline achieves strictly lower expected MSE than RLOO for \emph{any} number of rollouts $K_c < \infty$, offering a provably and consistently better variance-reduction strategy without biasing the gradient.
Empirically, as shown in \cref{fig:js_diagram} and \cref{sec:exps},
this theoretical advantage translates into consistent and measurable improvements across all rollout sizes, 
leading to lower gradient variance and improved downstream performance across diverse text prompt distributions.

\subsection{Single-Pass Scheduler as Dirichlet Policy}
\label{subsec:policy-param}

\paragraph{Dirichlet parameterization.}
We treat the entire denoising schedule as a \emph{single, joint action} rather than a sequence of stepwise decisions.  
Concretely, we define
$\tau \sim \mathrm{Dirichlet}(\boldsymbol{\alpha}_\theta(\mathbf{x}_T, \mathbf{c})),$
where the policy network outputs nonnegative parameters
$\boldsymbol{\alpha}_\theta \in \mathbb{R}_{+}^{L+1}$ conditioned on the initial noise $\mathbf{x}_T$ and text prompt $\mathbf{c}$.
Each sampled component $\tau_t$ represents a nonnegative interval, and the simplex constraint
$\sum_{t=1}^{L+1} \tau_t = 1$ ensures that these intervals partition the unit interval $[0,1]$.
We convert them into an ordered sampling schedule by cumulative summation:
\[
\tilde{t}_\ell = \sum_{j=1}^{\ell} \tau_j,
\quad \text{and} \quad
t_\ell = 1 - \tilde{t}_\ell,
\]
yielding a descending sequence $1 = t_0 > t_1 > \cdots > t_L > t_{L+1}=0$.
The final interval $\tau_{L+1}$ acts as a learnable \emph{stopping margin},
allowing the policy to dynamically adjust the effective sampling horizon.

This formulation naturally enforces simplex constraints, defines a smooth distribution over valid schedules,
and avoids the need for per-step discrete decisions.
Unlike conventional stepwise RL approaches whose complexity scales as $O(L)$,
our \emph{single-pass} Dirichlet policy maintains constant policy-gradient cost while producing flexible, instance-dependent denoising schedules.

\paragraph{Policy network.}
We implement the policy $\boldsymbol{\alpha}_\theta(\mathbf{x}_T, \mathbf{c})$ using a lightweight (20M parameters, $<\%1$ than that of the sampler network) encoder–decoder architecture.
The initial noise $\mathbf{x}_T$ is processed by multiscale convolutional blocks,
whose spatial features are fused with the pretrained text embedding $\mathbf{c}$ via cross-attention layers.
The fused representation is then projected through MLP layers to $(L\!+\!1)$ channels,
corresponding to the Dirichlet parameters that define interval allocations along the sampling timeline.
A final \texttt{softplus} activation followed by a small constant offset ($10^{-3}$)
ensures strictly positive outputs, guaranteeing valid Dirichlet parameters and numerically stable gradients.
The extra $(L\!+\!1)$-th channel naturally learns the residual “stopping” interval in the Dirichlet parameterization. Detailed network architectures and compute cost for the scheduler network can be found in the supplemental materials.

\paragraph{Training algorithm.}
At each iteration, we sample a mini-batch of $B$ contexts and, for each context $c$, draw $K_c$ schedules from the scheduler policy.
Each sampled schedule is executed by the sampler $\mathcal{S}$ to obtain a reward $r^{(c,i)}$,
and the corresponding baseline $b_{\mathrm{JS}}^{(c,i)}$ (Eq.~\eqref{eq:js-baseline}) is detached to maintain unbiasedness.
The policy is then updated by averaging the REINFORCE gradients with JS baselines across all samples. Of note, we (i) \emph{exclude} KL constraints or other regularizers, and (ii) initialize the scheduler policies \emph{from scratch}, in order to highlight the effectiveness of our framework and keep the experimental setting clean and interpretable.
Overall procedure is summarized in \cref{alg:rjs-train}.

\begin{algorithm}[!t]
  \caption{Policy gradient with James–Stein Baseline}
  \label{alg:rjs-train}
  \small
  \begin{algorithmic}[1]
    \STATE \textbf{Inputs:} sampler $\mathcal{S}$, policy $\pi_\theta$, reward $R$, optimizer $\mathsf{Opt}$
    \REPEAT
      \STATE Sample contexts $\{c\}_{1}^{B}$; set $g \leftarrow 0$
      \FOR{$c = 1~\TO~B$}
        \FOR{$i = 1~\TO~K_c$}
          \STATE $\tau^{(c,i)} \!\sim\! \pi_\theta(\cdot\!\mid\!c)$;\,
          $r^{(c,i)} \!\leftarrow\! R(\mathcal{S}(c,\tau^{(c,i)}), c)$
          \STATE Compute $b_{\mathrm{JS}}^{(c,i)}$ (Eq.~\eqref{eq:js-baseline})
          \STATE $g = g+ \frac{1}{BK_c}\! (r^{(c,i)} \!-\! b_{\mathrm{JS}}^{(c,i)}) \nabla_\theta \log \pi_\theta(\tau^{(c,i)}\!\mid\!c)$
        \ENDFOR
      \ENDFOR
      \STATE $\theta \leftarrow \mathsf{Opt.step}\!\left(\theta, g\right)$
    \UNTIL{convergence}
  \end{algorithmic}
\end{algorithm}

\section{Experiments}
\label{sec:exps}

\subsection{General Text-to-Image Experiments}
\label{sec:t2i}

\paragraph{Dataset and protocol.}
We first conduct extensive experiments on the Human Preference Dataset (HPD) v2 text-to-image benchmark \cite{wu2023human} to assess our method. We use the official train split $\sim 100K$ prompts to optimize the schedule policy and the held-out 3200 test split prompts for evaluation. We use the Human Preference Score (HPS) v2 model \cite{wu2023human} as our reward for these experiments.
For each backbone and sampling budget $L\in\{5,10,20,40,80\}$, we generate images for the same set of prompts and fixed noise seeds to ensure strict comparability across methods.
We study four representative pretrained samplers: SD-XL \cite{podell2023sdxl}, SD-3.5M/L \cite{esser2024scaling}, and Flux-Dev \cite{flux2024}, covering both UNet- and MMDiT-based backbones. For all models we consistently use a rollout number of 2. Due to memory limit, we use a batch size of 32 for smaller models like SD-XL or SD-3.5M, and a batch size of 16 for larger ones like SD-3.5L and Flux-Dev. See supplemental materials for more details.

We compare scheduling strategies under a unified training setup:
(1) the \emph{Default} fixed schedule provided by each backbone, schedules optimized with the (2) \emph{Cross-Context RLOO} baseline, (3) \emph{RLOO} baseline, and
(4) \emph{Ours (JS)} shrinkage baseline.
For Flux we additionally include a \emph{TPDM-style PPO} variant using the same architecture for a direct comparison. All other configurations of TPDM follow its default setups. We provide detailed description and discussions in the supplemental materials.
All methods share identical optimization settings and data pipelines, ensuring that performance differences arise solely from the choice of reward baseline and policy learning formulation.

\begin{figure}[t]
\centering

\colhead{... wearing {\color{red}... Rick Owens clothing}}\hfill
\colhead{... depicted {\color{red}in World of Warcraft}}\hfill
\colhead{... man camouflaged {\color{red}as a waffle on a plate}}

\begin{subfigure}[t]{0.315\linewidth}
  \centering
  \includegraphics[width=\linewidth]{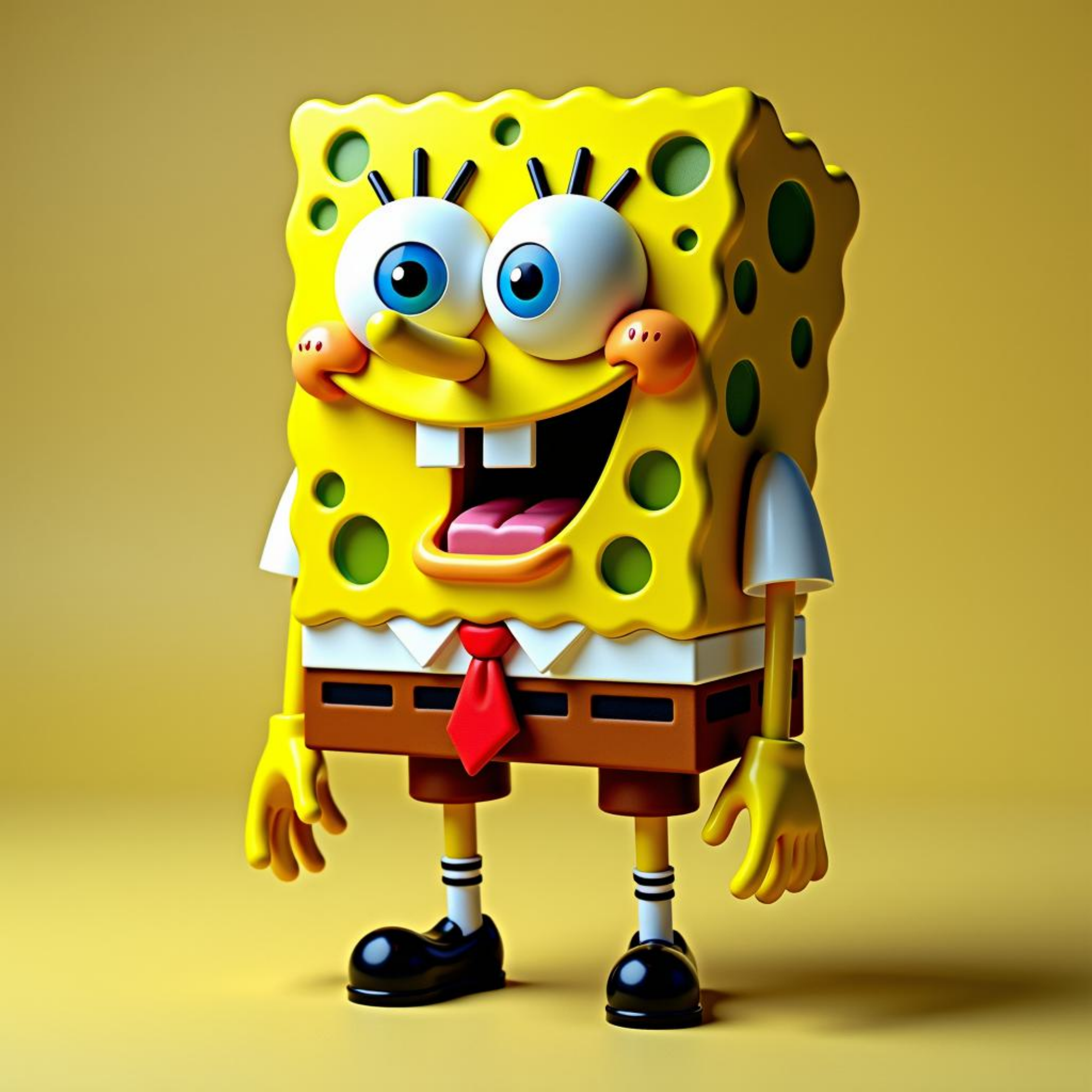}
  \label{fig:gt_r1c1}
\end{subfigure}\hfill
\begin{subfigure}[t]{0.315\linewidth}
  \centering
  \includegraphics[width=\linewidth]{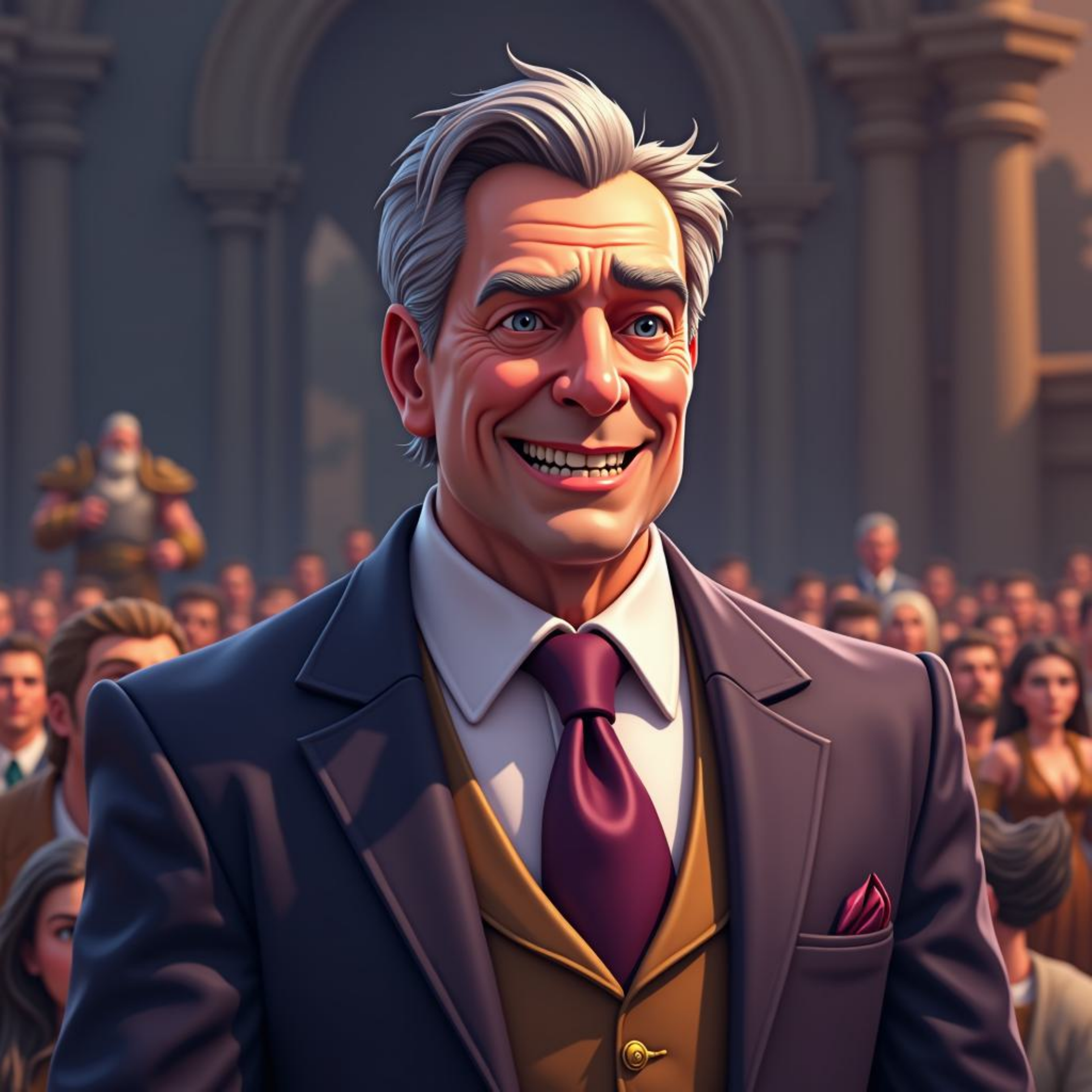}
  \label{fig:gt_r1c2}
\end{subfigure}\hfill
\begin{subfigure}[t]{0.315\linewidth}
  \centering
  \includegraphics[width=\linewidth]{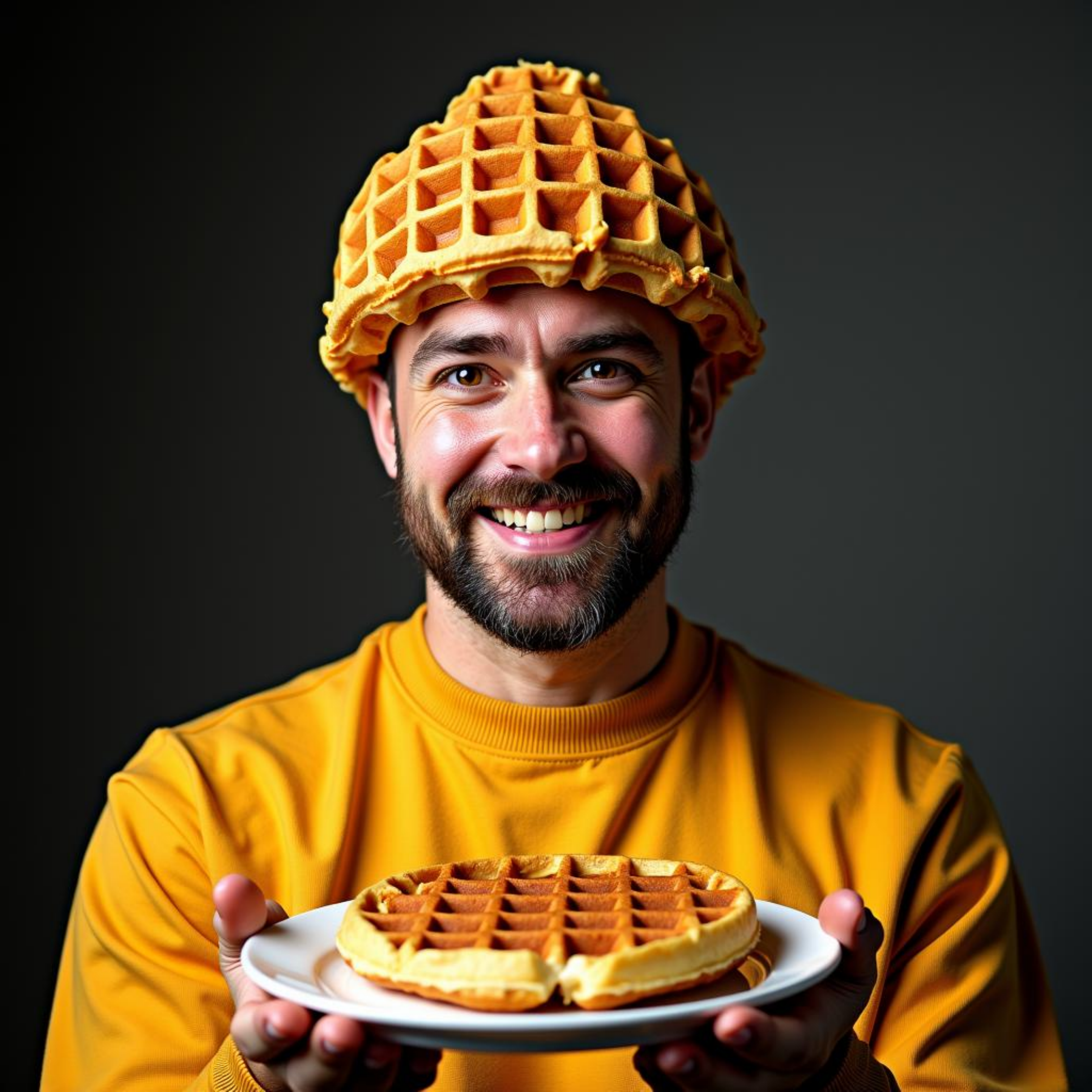}
  \label{fig:gt_r1c3}
\end{subfigure}

\vspace{-10pt} 

\begin{subfigure}[t]{0.315\linewidth}
  \centering
  \includegraphics[width=\linewidth]{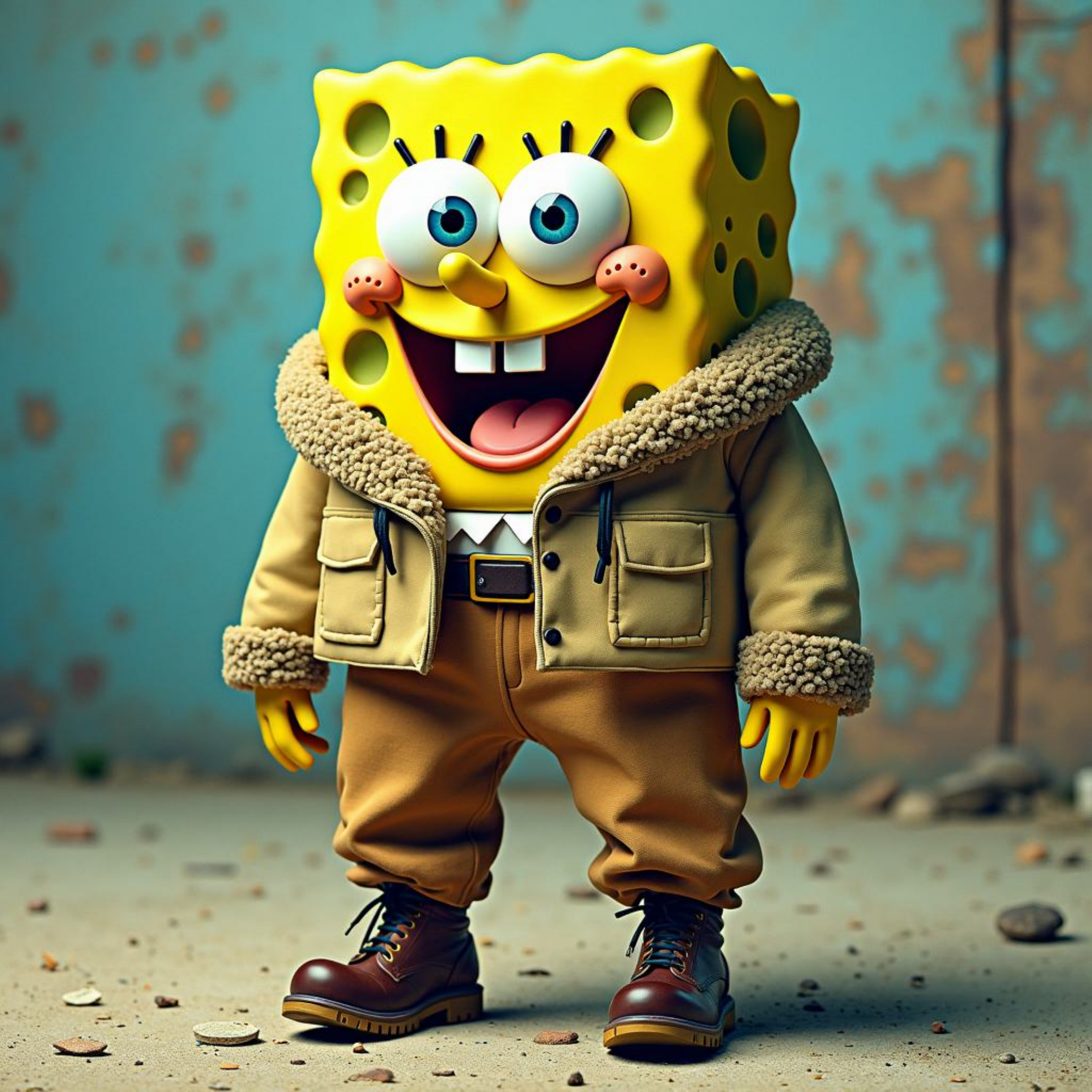}
  \label{fig:gt_r2c1}
\end{subfigure}\hfill
\begin{subfigure}[t]{0.315\linewidth}
  \centering
  \includegraphics[width=\linewidth]{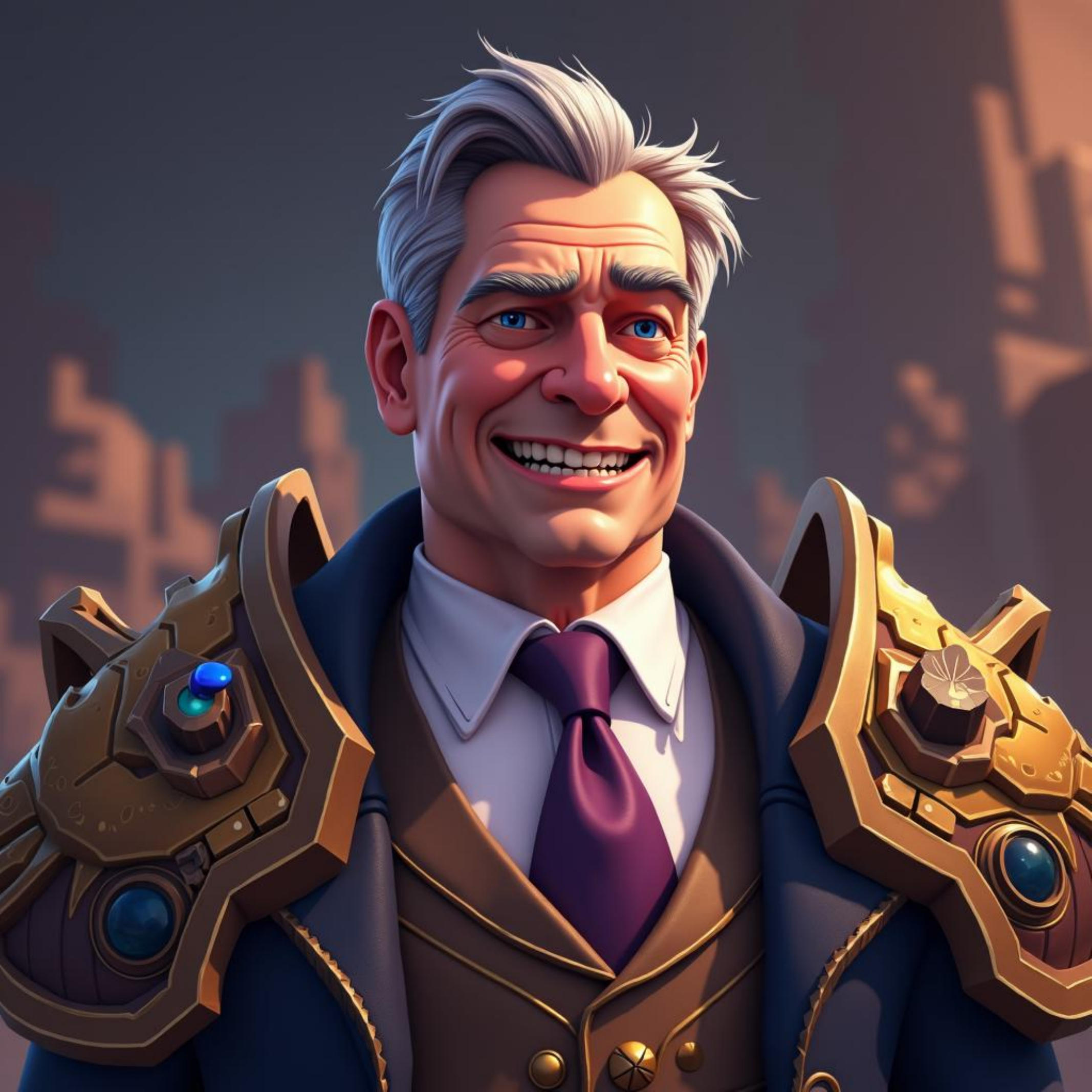}
  \label{fig:gt_r2c2}
\end{subfigure}\hfill
\begin{subfigure}[t]{0.315\linewidth}
  \centering
  \includegraphics[width=\linewidth]{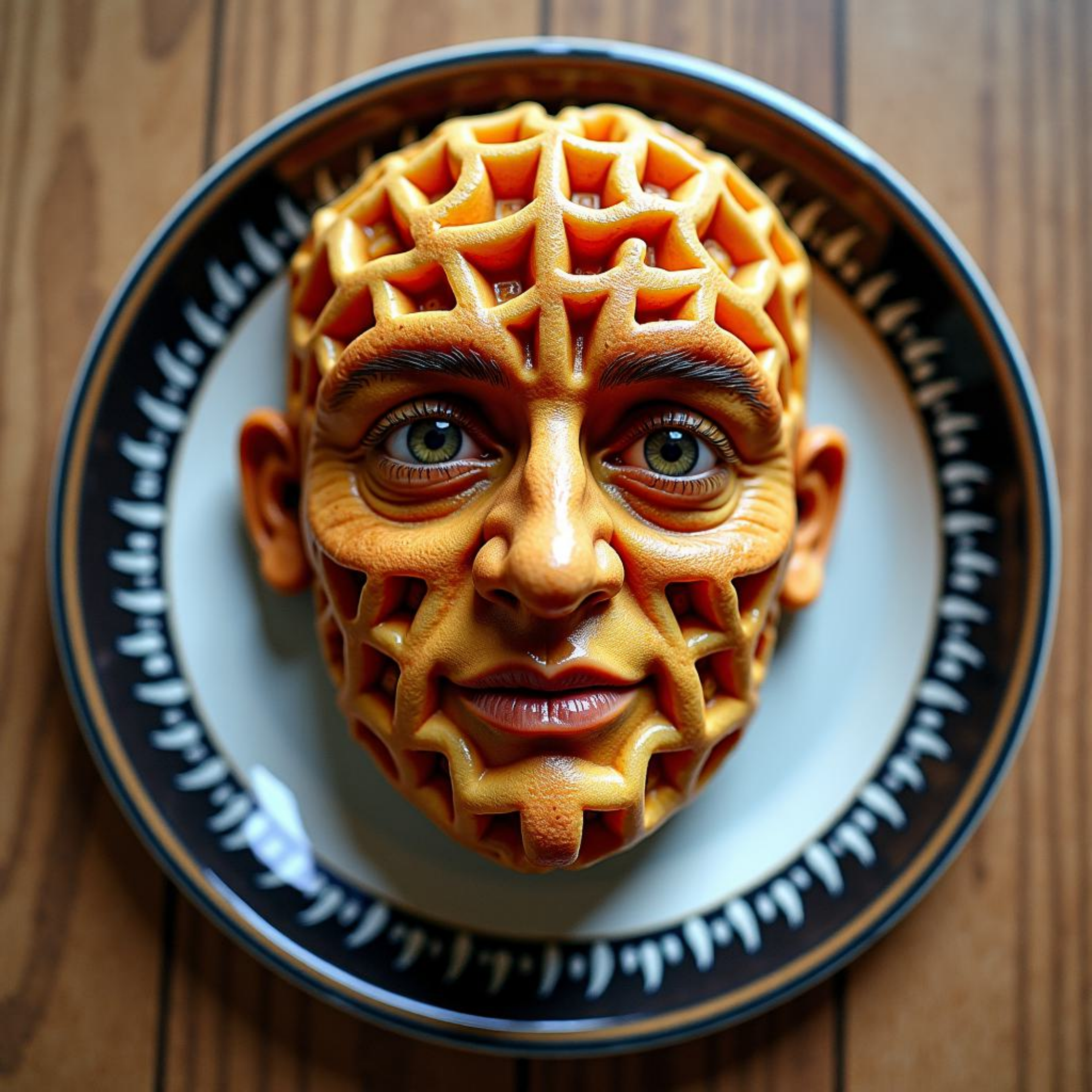}
  \label{fig:gt_r2c3}
\end{subfigure}
\vspace{-10pt}
\caption{\textbf{Rescheduling improves general T2I alignment.} Head-to-head comparisons between images generated with default schedules (upper) and our learned schedules (lower) from Flux-Dev with 40 steps. Figures henceforth follow the same format.}
\label{fig:gen_t2i}
\vspace{-10pt}
\end{figure}

\begin{figure*}[t]
  \centering
  \begin{minipage}[t]{0.54\textwidth}\vspace{0pt}
    \captionsetup{type=table} 
    \caption{\textbf{HPSv2 scores on HPD2 held-out prompts.} Scores vs.\ number of inference steps ($L$).
      Rows per backbone: Default, Cross-Context RLOO (XCTX), RLOO, TPDM-style PPO (Flux only), and Ours (JS).
      Best per column in \textbf{bold}; second-best \underline{underlined}.}
    \label{tab:hpdt2i_steps}

    \setlength{\tabcolsep}{8pt}
    \renewcommand{\arraystretch}{1.12}
    \scriptsize

    \resizebox{\linewidth}{!}{%
      \begin{tabular}{l lccccc}
        \toprule
        \textbf{Backbone} & \textbf{Methods / Steps ($L$)} & \textbf{5} & \textbf{10} & \textbf{20} & \textbf{40} & \textbf{80} \\
        \midrule

        \multirow{4.5}{*}{\cellcolor{white} SD\mbox{-}XL}
          & Default            & 18.25 & 25.47 & 27.69 & \underline{28.52} & \underline{28.55} \\
          \cmidrule(lr){2-2}
          & XCTX               & 24.03 & 26.44 & 27.30 & 27.59 & 28.19 \\
          & RLOO               & \underline{24.10} & \underline{26.59} & \underline{27.85} & 28.42 & 28.50 \\
          \ShadedOurs{24.22}{26.89}{27.98}{\textbf{28.53}}{\textbf{28.66}}
        \cmidrule(lr){1-7}

        \multirow{4.5}{*}{\cellcolor{white} SD3.5\mbox{-}M}
          & Default            & 21.59 & 26.31 & \underline{28.66} & \textbf{29.61} & \textbf{29.91} \\
          \cmidrule(lr){2-2}
          & XCTX               & 23.96 & 25.32 & 27.53 & 26.70 & 28.14 \\
          & RLOO               & \underline{24.05} & \underline{27.20} & 28.52 & 29.03 & 29.36 \\
          \ShadedOurs{24.31}{27.31}{28.87}{\underline{29.47}}{\underline{29.62}}
        \cmidrule(lr){1-7}

        \multirow{4.5}{*}{\cellcolor{white} SD3.5\mbox{-}L}
          & Default            & 24.24 & 28.04 & \underline{29.85} & \textbf{30.43} & \underline{30.61} \\
          \cmidrule(lr){2-2}
          & XCTX               & 24.92 & 27.39 & 28.44 & 28.03 & 29.30 \\
          & RLOO               & \underline{25.55} & \underline{28.61} & 29.37 & 29.75 & 30.27 \\
          \ShadedOurs{26.28}{28.88}{29.98}{\underline{30.41}}{\textbf{30.64}}
        \cmidrule(lr){1-7}

        \multirow{5.5}{*}{\cellcolor{white} Flux\mbox{-}Dev}
          & Default            & 23.73 & 28.06 & 29.88 & 30.84 & 31.04 \\
          \cmidrule(lr){2-2}
          & TPDM PPO           & 15.73 & 16.24 & 17.05 & 17.98 & 18.65 \\
          & XCTX               & \underline{26.92} & 29.39 & 30.09 & 29.94 & 30.88 \\
          & RLOO               & 26.48 & \underline{30.41} & \underline{30.77} & \underline{30.92} & \underline{31.10} \\
          \ShadedOurs{29.21}{30.86}{31.12}{\textbf{31.23}}{\textbf{31.36}}
        \bottomrule
      \end{tabular}%
    }
  \end{minipage}
  \hfill
  \begin{minipage}[t]{0.43\textwidth}\vspace{0pt}
    \centering
    \begin{subfigure}[t]{0.49\linewidth}
      \centering
      \includegraphics[width= \linewidth]{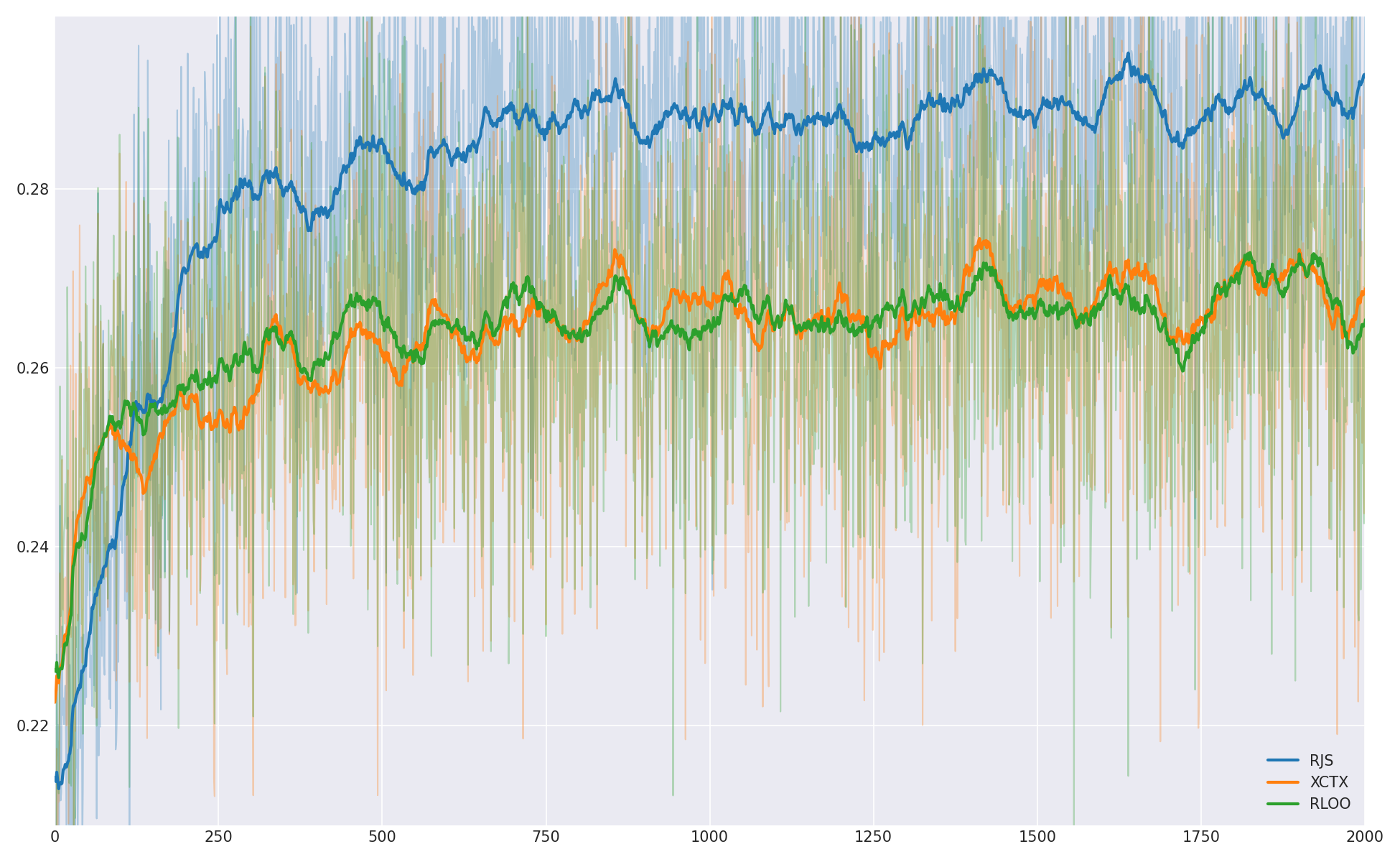}
      \caption{$L=5$}
    \end{subfigure}
    \begin{subfigure}[t]{0.49\linewidth}
      \centering
      \includegraphics[width= \linewidth]{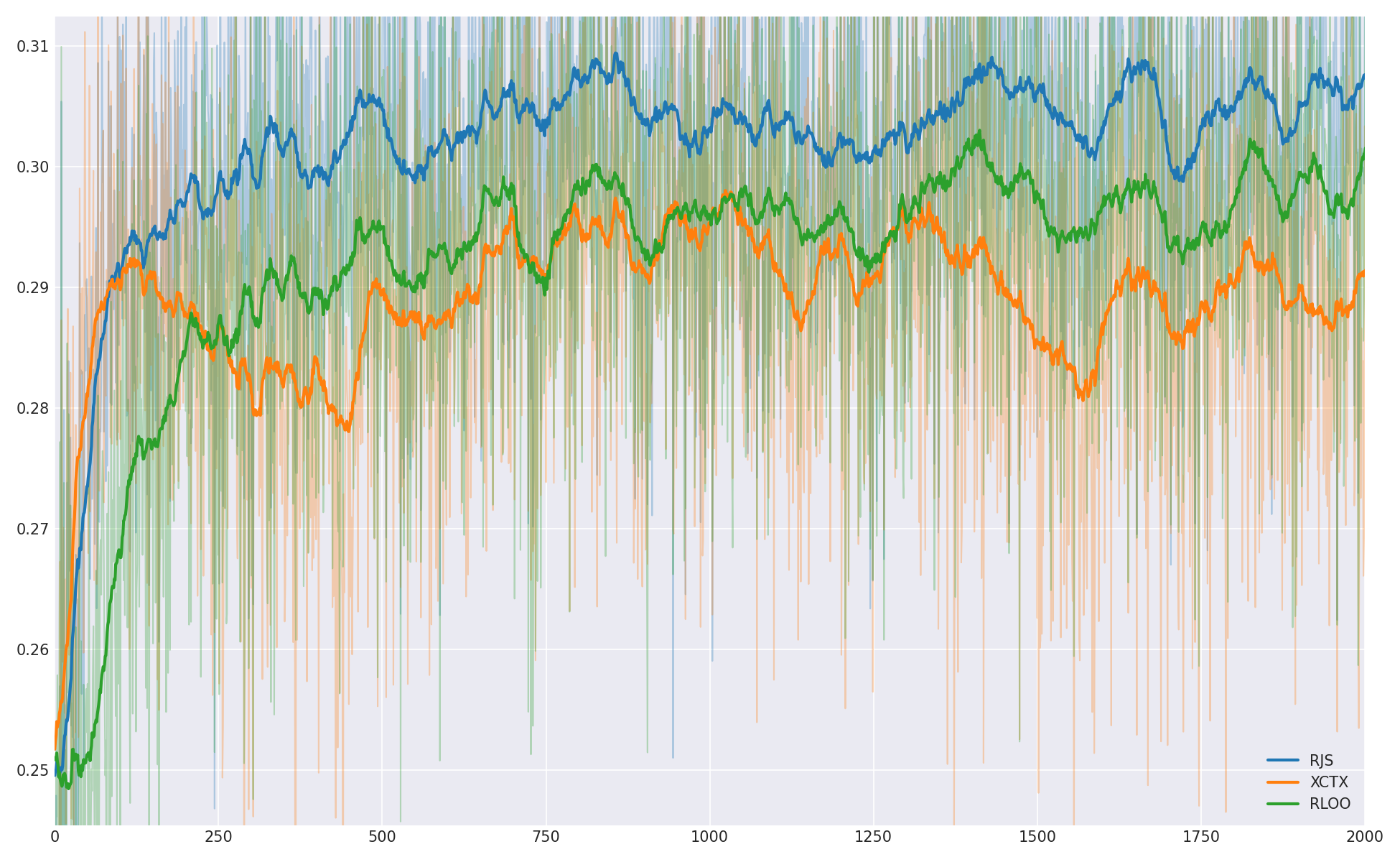}
      \caption{$L=10$}
    \end{subfigure}\vspace{4pt}

    \begin{subfigure}[t]{0.49\linewidth}
      \centering
      \includegraphics[width= \linewidth]{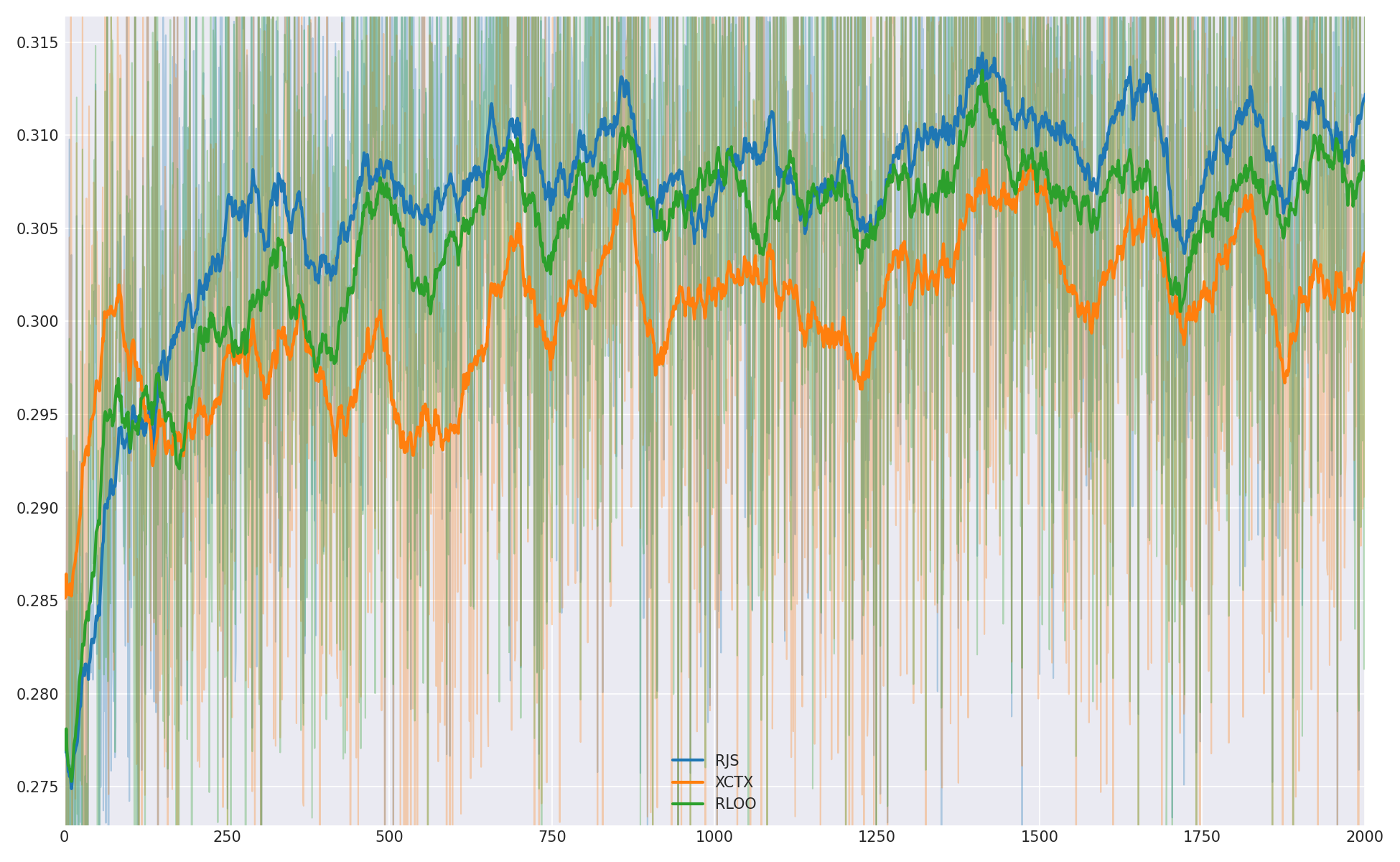}
      \caption{$L=20$}
    \end{subfigure}
    \begin{subfigure}[t]{0.49\linewidth}
      \centering
      \includegraphics[width= \linewidth]{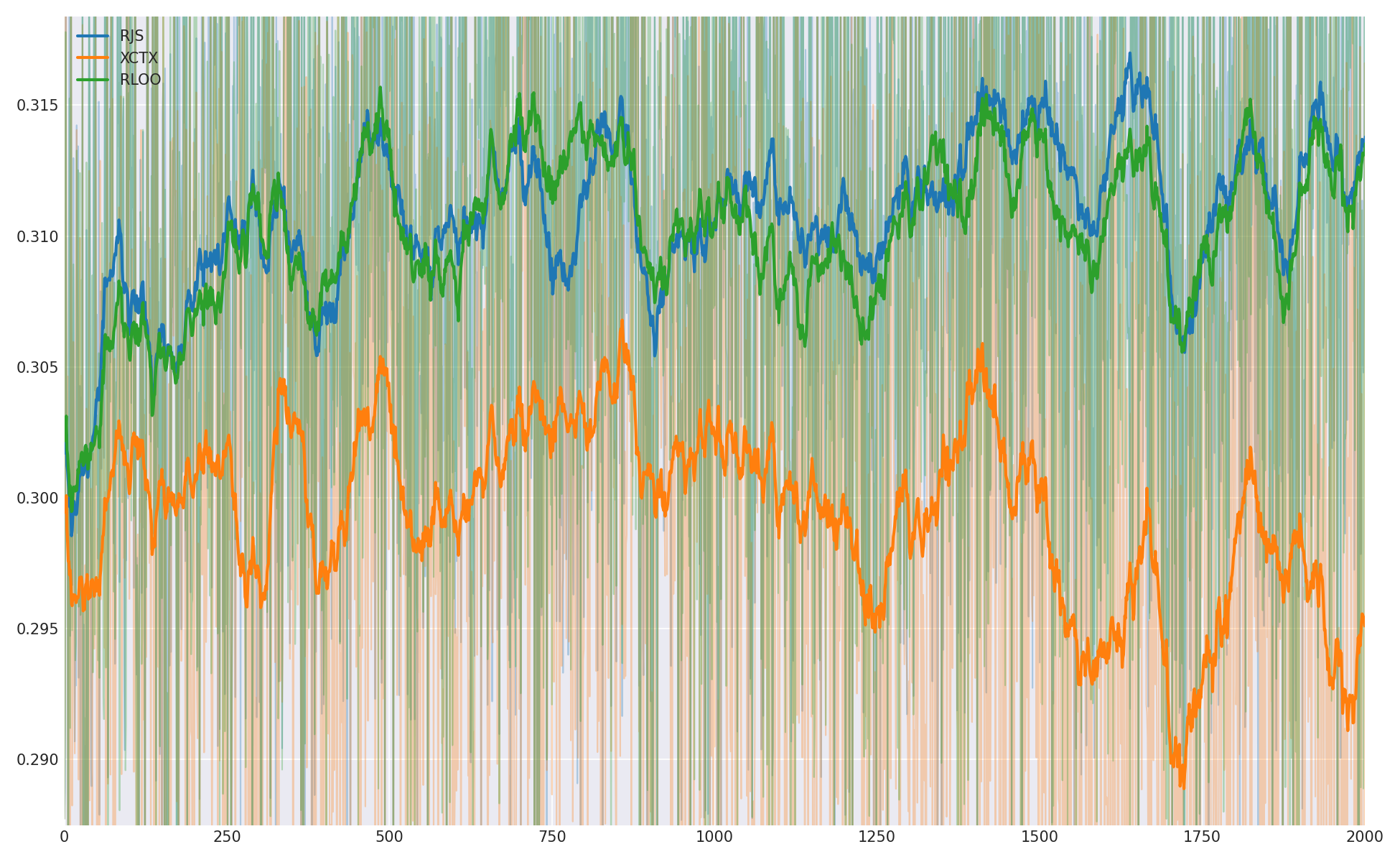}
      \caption{$L=40$}
    \end{subfigure}
    \captionof{figure}{\textbf{Representative training curves (Flux-Dev) for different baselines and sampling steps $L$.} Y-axis denotes the aggregated rewards; X-axis denotes the num. of iterations. We observe that {\color{blue}JS} baseline consistently outperforms {\color{orange}XCTX} and {\color{green!50!blue}RLOO} baselines; the gap is clear at low budget and persists even when $L$ is large when the discretization error of sampling trajectory diminishes.}
    \label{fig:training_curves}
  \end{minipage}

  \vspace{-2mm}
\end{figure*}

\paragraph{Findings.}
Across all backbones and sampling budgets, our JS policy achieves consistently higher alignment scores than all other baselines (\cref{tab:hpdt2i_steps}).
The gains are most pronounced in the low-budget regime ($L\!\le\!20$), where sampling variance dominates and variance reduction is most impactful.
At sufficiently large budgets, discretization error along the sampling trajectory diminishes, naturally reducing the relative variance among different baselines.
Even in that regime, JS remains uniformly better across all backbones, with the most substantial margins observed on the Flux-Dev model.
These results confirm that JS effectively reduces gradient estimation errors under few-sample conditions; it retains consistent advantages as the discretization error diminishes with more steps.

\paragraph{Few-step sampling.}
The benefits of rescheduling are particularly striking under severe step constraints (\cref{tab:hpdt2i_steps,tab:flux5}). With only $L=5$ steps, \emph{Ours (JS)} on Flux-Dev nearly matches the performance of the deliberately distilled Flux-schnell sampler, despite operating on the non-distilled backbone. This highlights that a learned, prompt-adaptive schedule can significantly improve the \emph{effective} sampling efficiency, revealing that pretrained samplers possess meaningful few-step capabilities even without distillation. Across all backbones, JS consistently outperforms its counterparts in the few-step regime, confirming that dynamically reallocating the denoising budget provides tangible gains precisely where discretization error and reward heterogeneity are most pronounced.

\begin{table}[t]
\centering
\caption{Flux-Dev 5-step sampling on HPDv2.}
\label{tab:flux5}
\setlength{\tabcolsep}{6pt}
\renewcommand{\arraystretch}{1.12}
\scriptsize
\begin{tabular}{ccccc|c}
\toprule
\textbf{Default} &
\textbf{TPDM PPO} &
\textbf{Cr. RLOO} &
\textbf{RLOO} &
\textbf{RJS} &
\textbf{Schnell} \\
\midrule
23.73 &
15.73 &
26.92 &
26.48 &
\underline{\cellcolor{gray!20}29.21} &
\textbf{29.42} \\
\bottomrule
\end{tabular}
\end{table}

\begin{figure}[t]
\centering

\colhead{Lionel Messi portrayed as a sitcom character}\hfill
\colhead{A helmet-wearing monkey skating}\hfill
\colhead{A kangaroo in an orange hoodie... holding ``Welcome Friends''}

\begin{subfigure}[t]{0.315\linewidth}
  \centering
  \includegraphics[width=\linewidth]{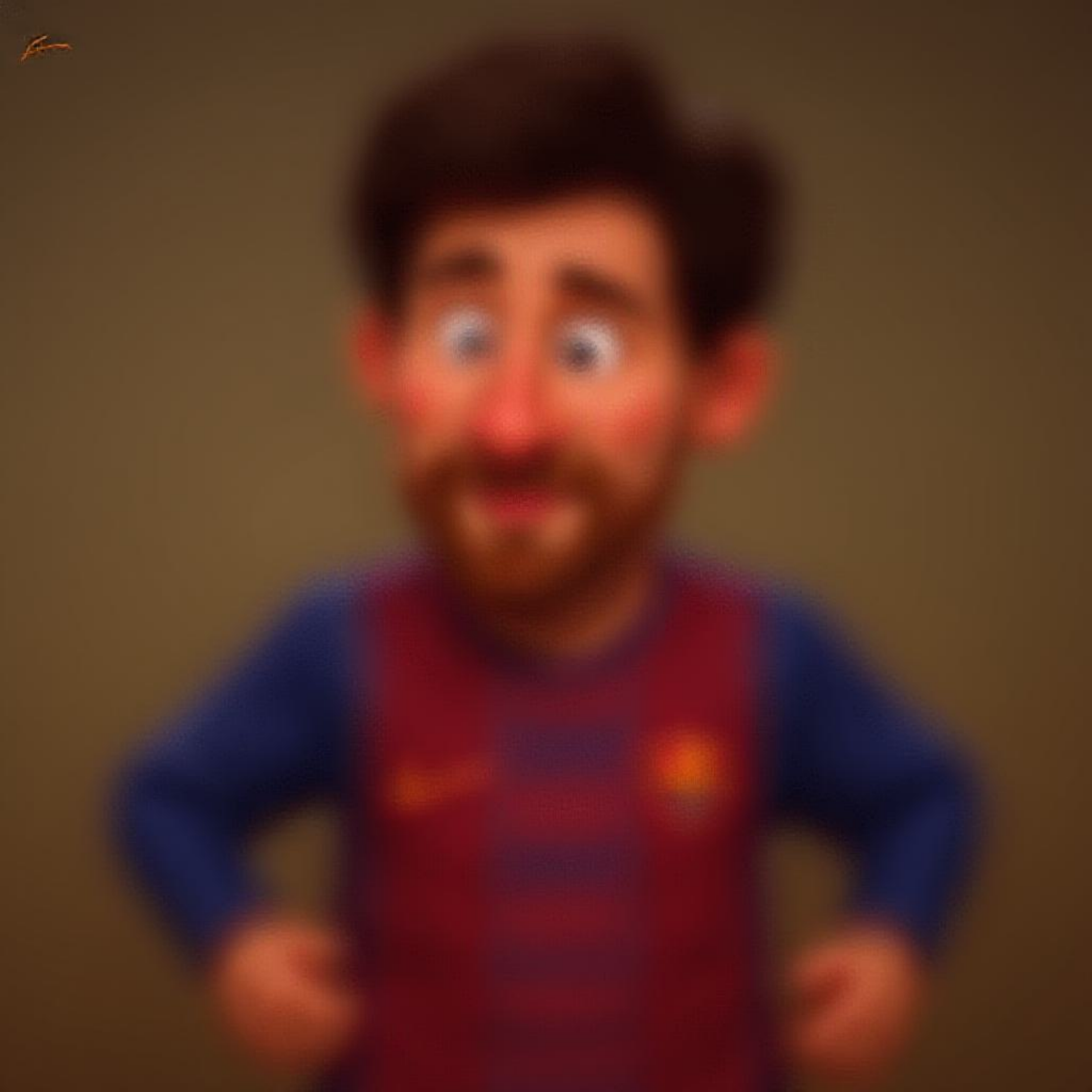}
  \label{fig:fs_r1c1}
\end{subfigure}\hfill
\begin{subfigure}[t]{0.315\linewidth}
  \centering
  \includegraphics[width=\linewidth]{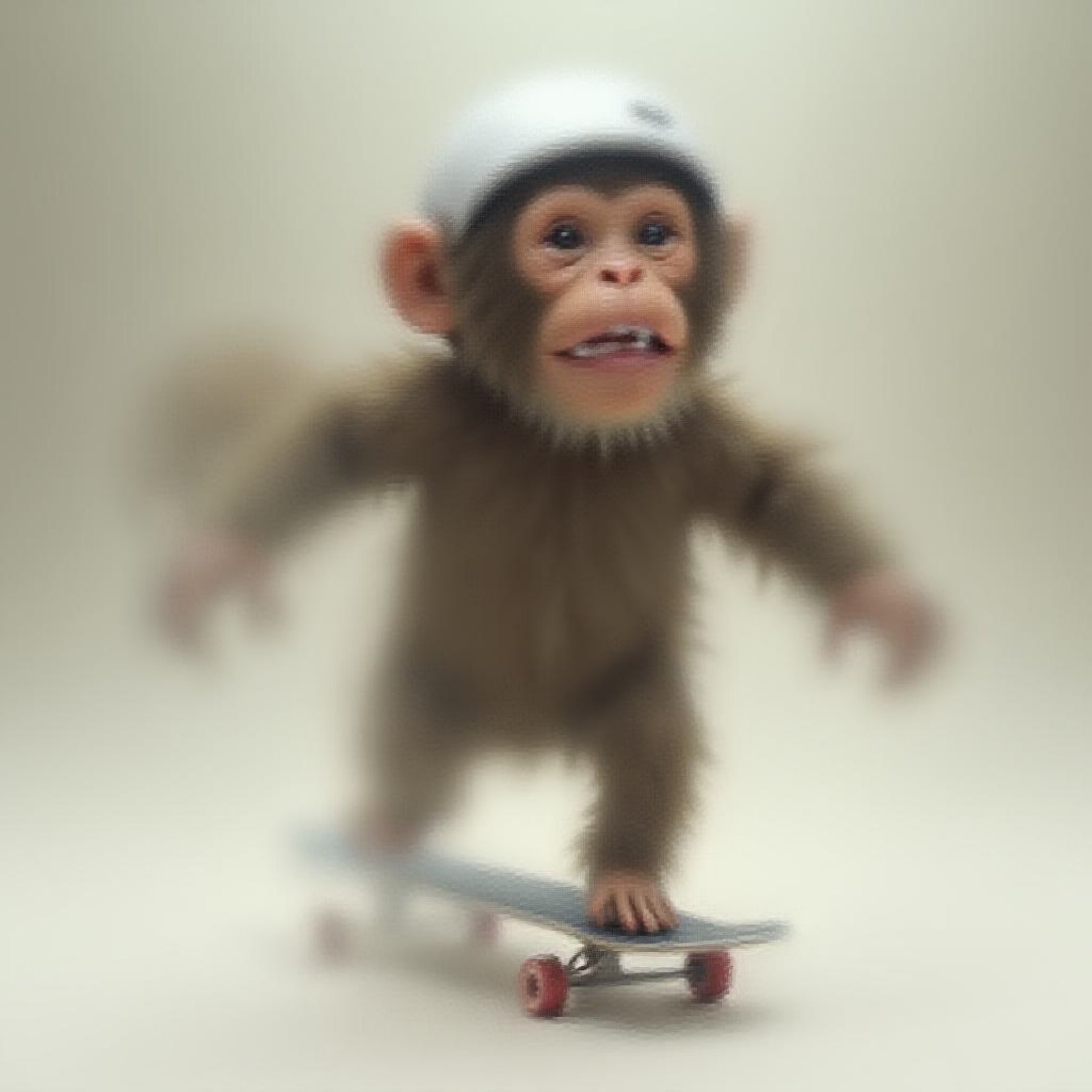}
  \label{fig:fs_r1c2}
\end{subfigure}\hfill
\begin{subfigure}[t]{0.315\linewidth}
  \centering
  \includegraphics[width=\linewidth]{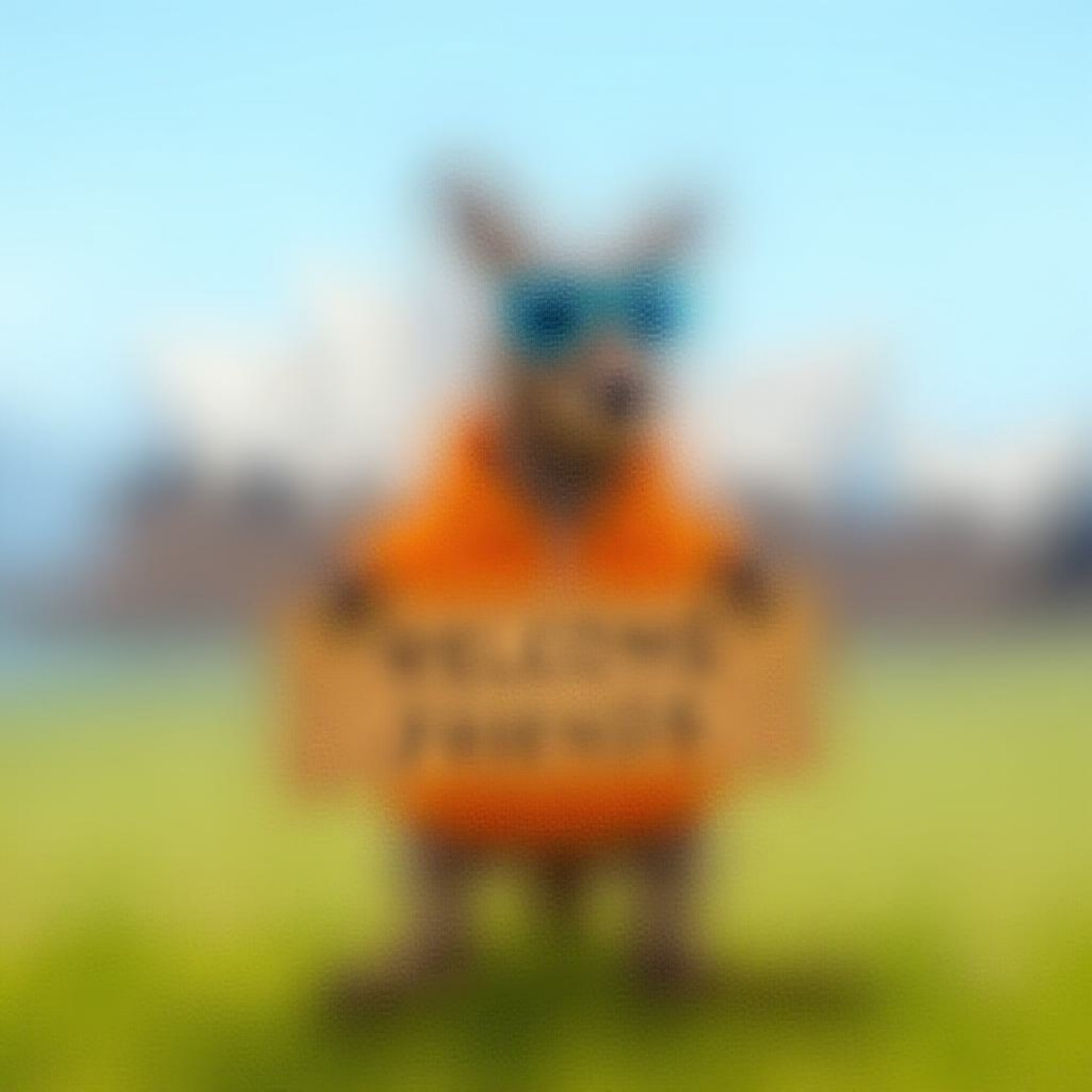}
  \label{fig:fs_r1c3}
\end{subfigure}

\vspace{-10pt} 

\begin{subfigure}[t]{0.315\linewidth}
  \centering
  \includegraphics[width=\linewidth]{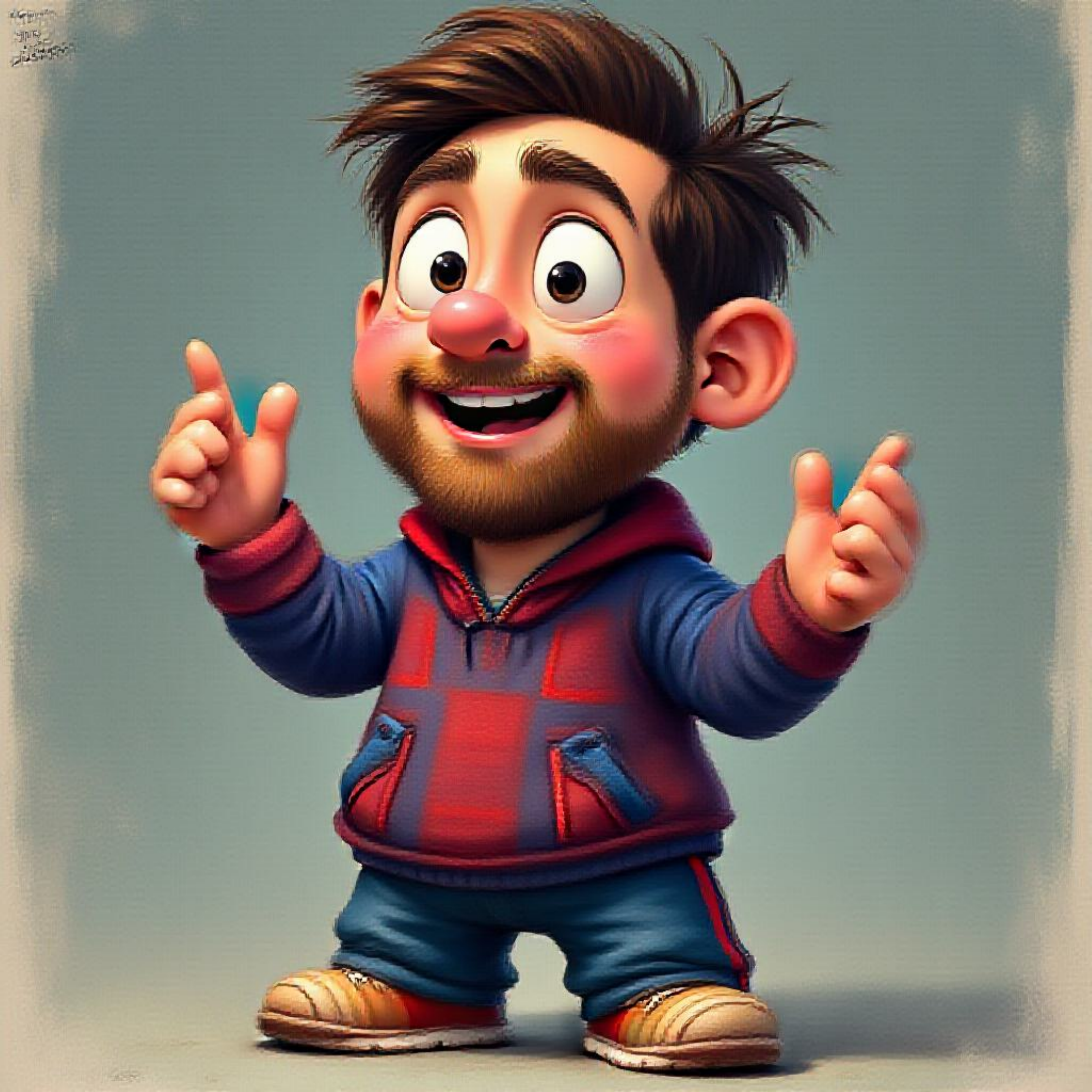}
  \label{fig:fs_r2c1}
\end{subfigure}\hfill
\begin{subfigure}[t]{0.315\linewidth}
  \centering
  \includegraphics[width=\linewidth]{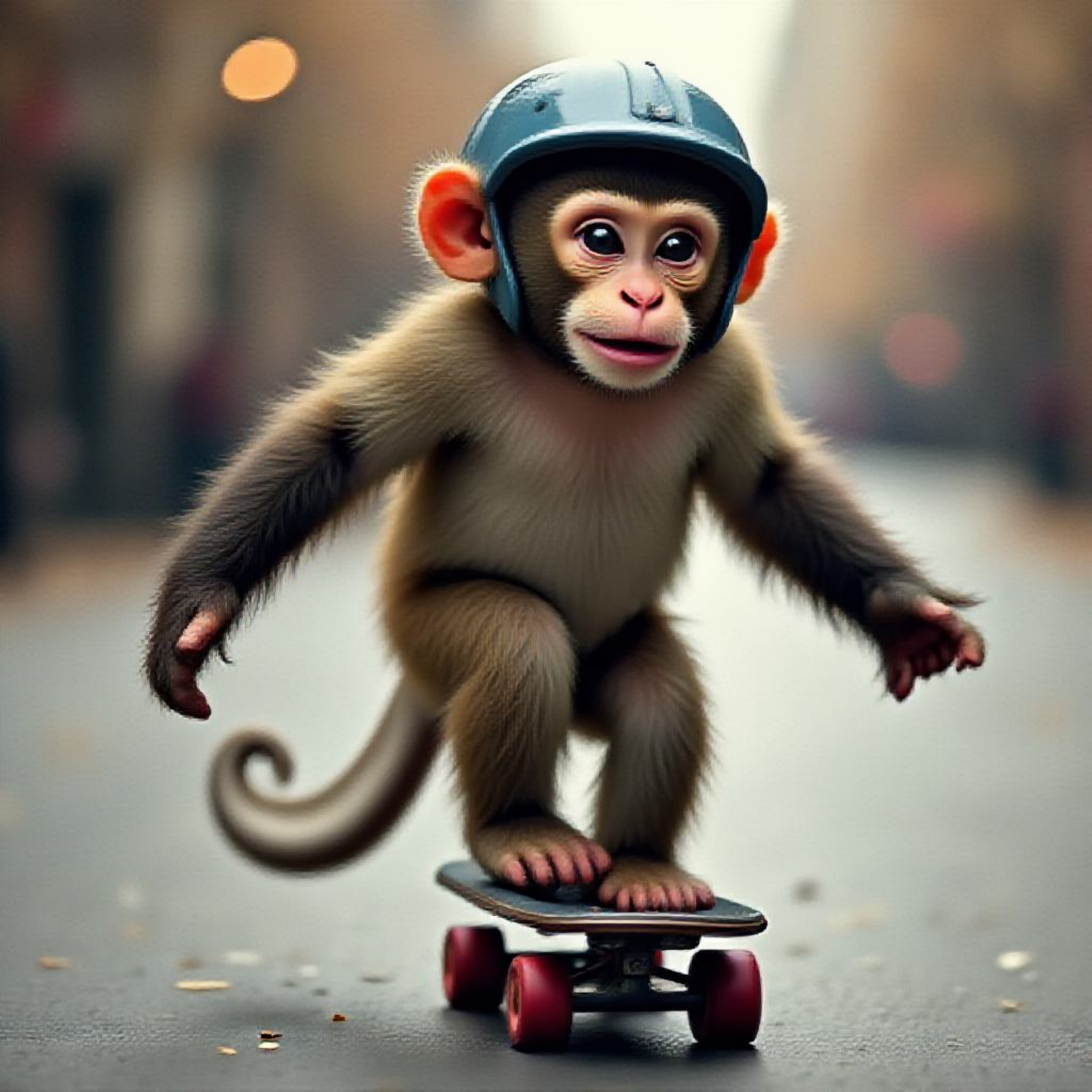}
  \label{fig:fs_r2c2}
\end{subfigure}\hfill
\begin{subfigure}[t]{0.315\linewidth}
  \centering
  \includegraphics[width=\linewidth]{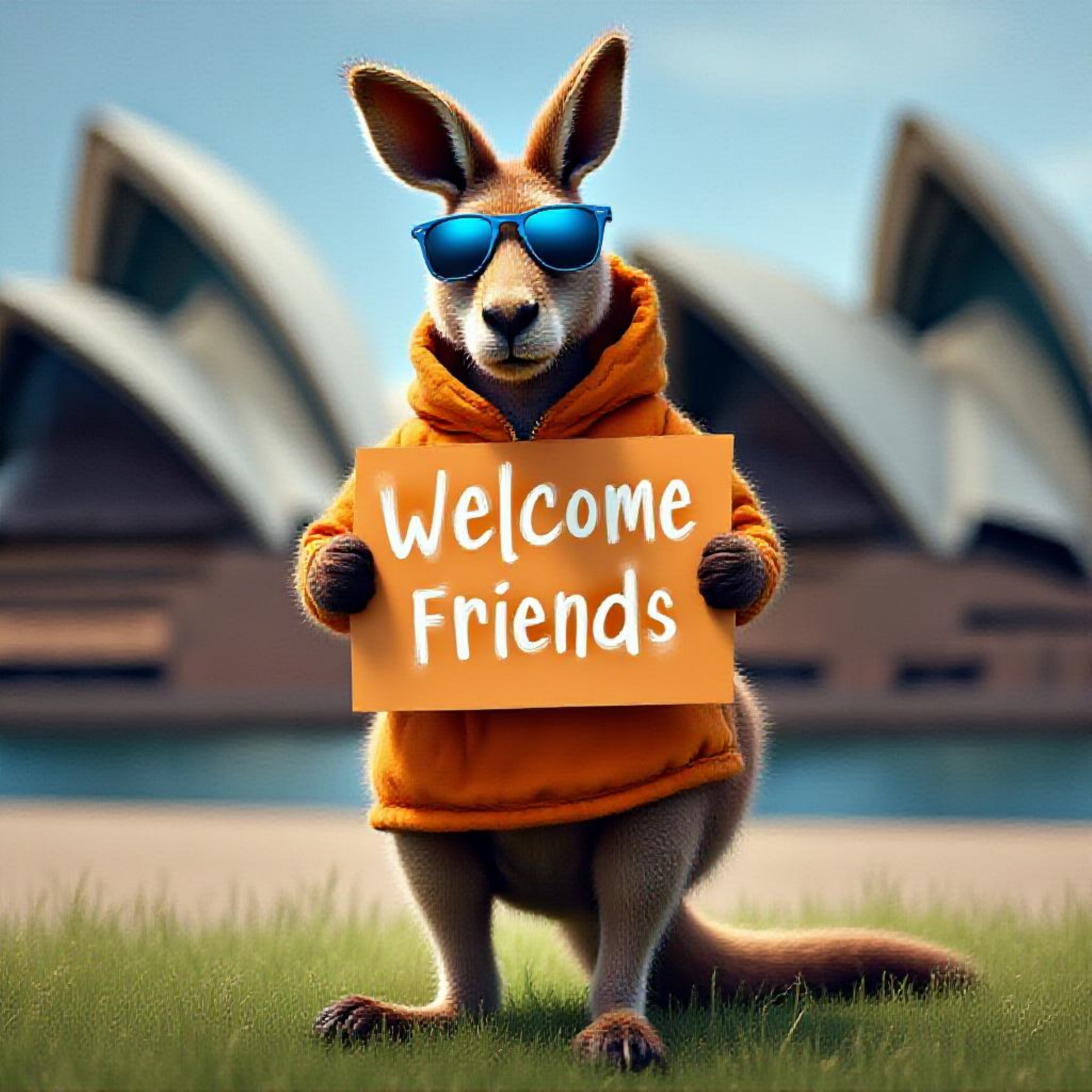}
  \label{fig:fs_r2c3}
\end{subfigure}
\vspace{-10pt}
\caption{\textbf{Rescheduling improves few-step sampling}. Comparisons between images generated with default schedules (upper) and our learned schedules (lower) from Flux-Dev, with 5 steps.}
\vspace{-15pt}
\label{fig:few_steps}
\end{figure}

\subsection{Text Rendering and Fine-Grained Alignment}
\label{subsec:text-rendering}

\paragraph{Dataset and protocol.}
Building on the general text-to-image results in \cref{sec:t2i}, we further assess the fine-grained alignment capability with our method. When the sampling budget is sufficiently large, the gap between the default schedule and our learned schedules may appear to narrow when measured by global preference scores; however, we find that our schedules yield substantially stronger fine-grained alignment capabilities. We focus on two representative scenarios: (i) prompt-specified text rendering, and (ii) object-centric generation tasks following GenEval~\cite{ghosh2023geneval}.

For text rendering, we follow \cite{liu2025flow} and use its 20K text-centric training prompts (alphanumeric strings, short phrases, and proper nouns) and 500 held-out testing prompts. We use again the HPSv2 model as our reward. For GenEval tasks, we use the official script to generate 50K training prompts and 1816 held-out and deduplicated testing prompts. We use a rule-based reward as in \cite{liu2025flow} as our reward. We focus on two representative backbones: SD-3.5M and Flux-Dev, and compare the backbone’s \emph{Default} fixed schedule with \textbf{Ours (JS)} learned schedules. We set the sampling steps to 40 for all experiments. For text rendering, we report \textbf{HPSv2} (↑) for overall perceptual preference and three character-level text-specific OCR metrics: \textbf{OCR-A} (↑, accuracy), \textbf{OCR-P} (↑, precision) and \textbf{OCR-R} (↑, Recall).
Human preference (HPSv2) complements OCR, which can be brittle on stylized glyphs.
For GenEval tasks, we report the corresponding scores on the held-out testing prompts with the official evaluation pipeline. We refer to supplemental materials for more experiment details.

\begin{table}[t]
  \centering
  \caption{\textbf{Text rendering at a fixed step budget ($T$).}
  HPSv2 (↑), OCR-A (↑), OCR-P (↑), OCR-R (↑). Best per backbone in \textbf{bold}.}
  \label{tab:text-rendering-concise}
  \setlength{\tabcolsep}{7pt}
  \renewcommand{\arraystretch}{1.12}
  \scriptsize
  \begin{tabular}{lcccc}
    \toprule
    \textbf{SD3.5-M} & \textbf{HPSv2 (↑)} & \textbf{OCR-A (↑)} & \textbf{OCR-P (↑)} & 
    \textbf{OCR-R (↑)} \\
    \midrule
    Default & \textbf{27.23} & 19.08 & 53.69 & 53.34 \\
    \rowcolor{gray!20}
    \textbf{Ours (JS)}& 27.07 & \textbf{20.04} & \textbf{54.93} & \textbf{58.19} \\
    \midrule
    \textbf{Flux-Dev} & \textbf{HPSv2 (↑)} & \textbf{OCR-A (↑)} & \textbf{OCR-P (↑)} & 
    \textbf{OCR-R (↑)} \\
    \midrule
    Default & 28.70 & 20.06 & 52.63 & 49.77 \\
    \rowcolor{gray!20}
    \textbf{Ours (JS)}& \textbf{28.91} & \textbf{21.30} & \textbf{54.05} & \textbf{58.58} \\
    \bottomrule
  \end{tabular}
\end{table}

\begin{figure}[t]
\centering

\colhead{...{\color{red}``Click Here for Discount''} slogan}\hfill
\colhead{... beaming {\color{red} ``Safe Harbor''}}\hfill
\colhead{... on director chair {\color{red}``Scene 1 Take 2''}}

\begin{subfigure}[t]{0.315\linewidth}
  \centering  \includegraphics[width=\linewidth]{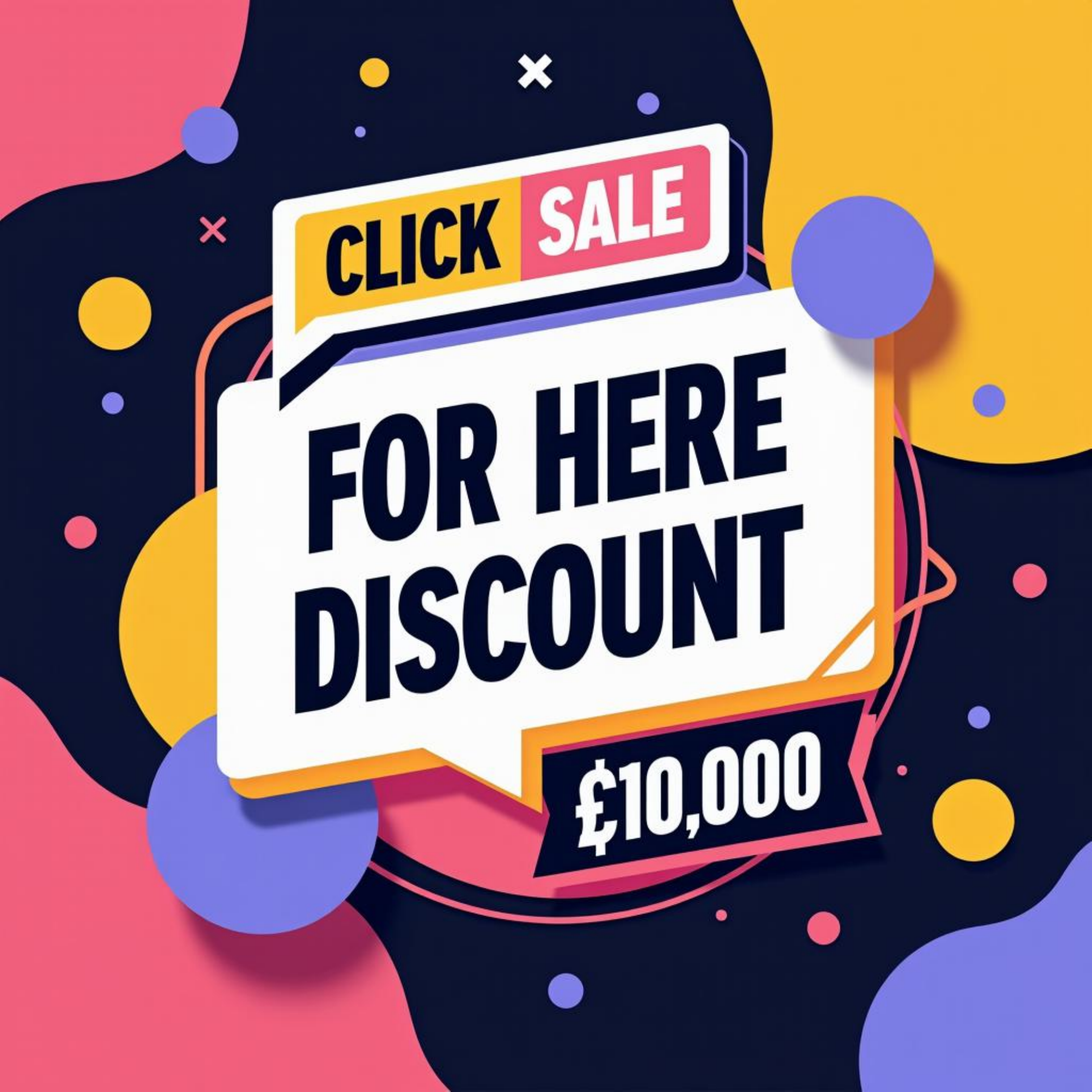}
  \label{fig:tr_r1c1}
\end{subfigure}\hfill
\begin{subfigure}[t]{0.315\linewidth}
  \centering
  \includegraphics[width=\linewidth]{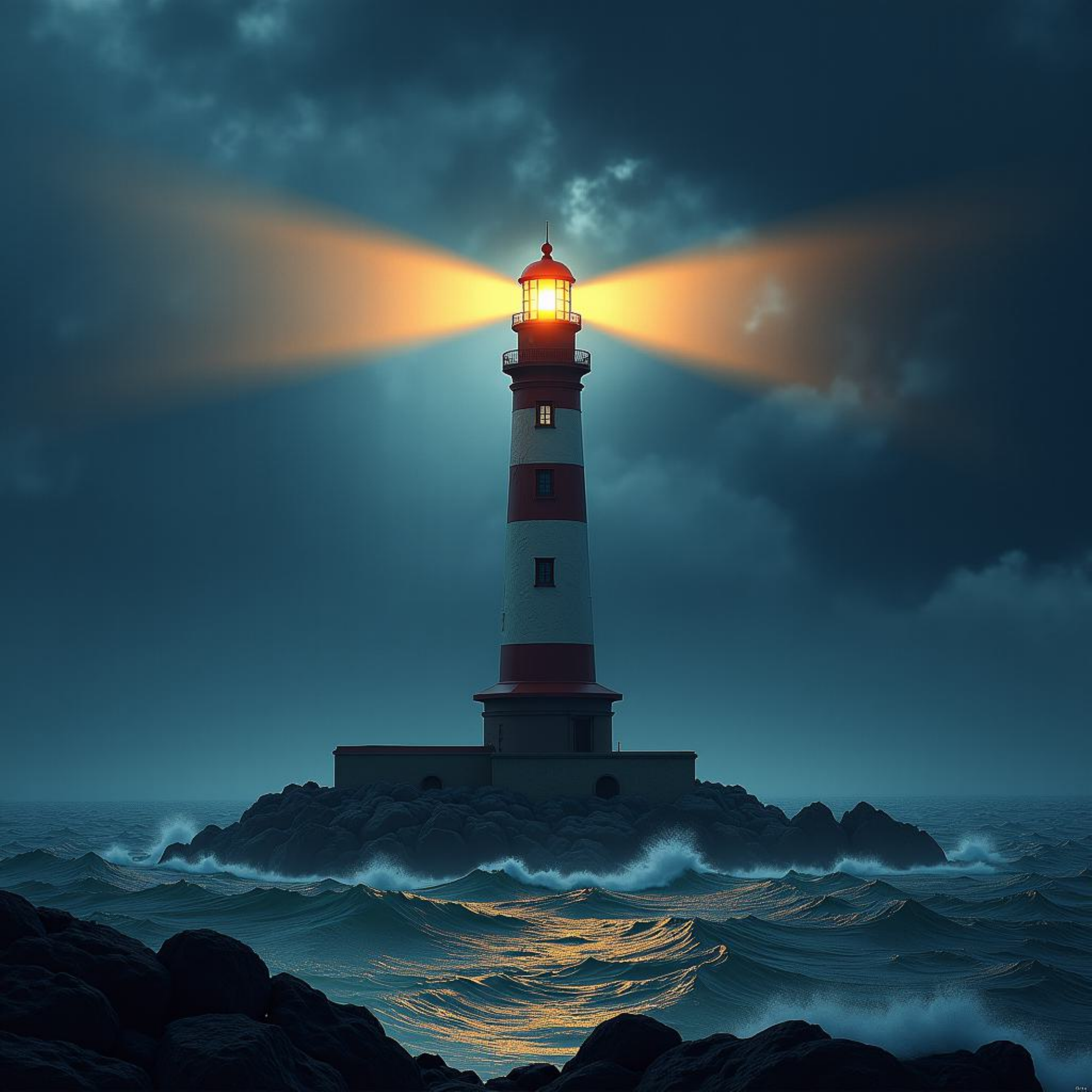}
  \label{fig:tr_r1c2}
\end{subfigure}\hfill
\begin{subfigure}[t]{0.315\linewidth}
  \centering
  \includegraphics[width=\linewidth]{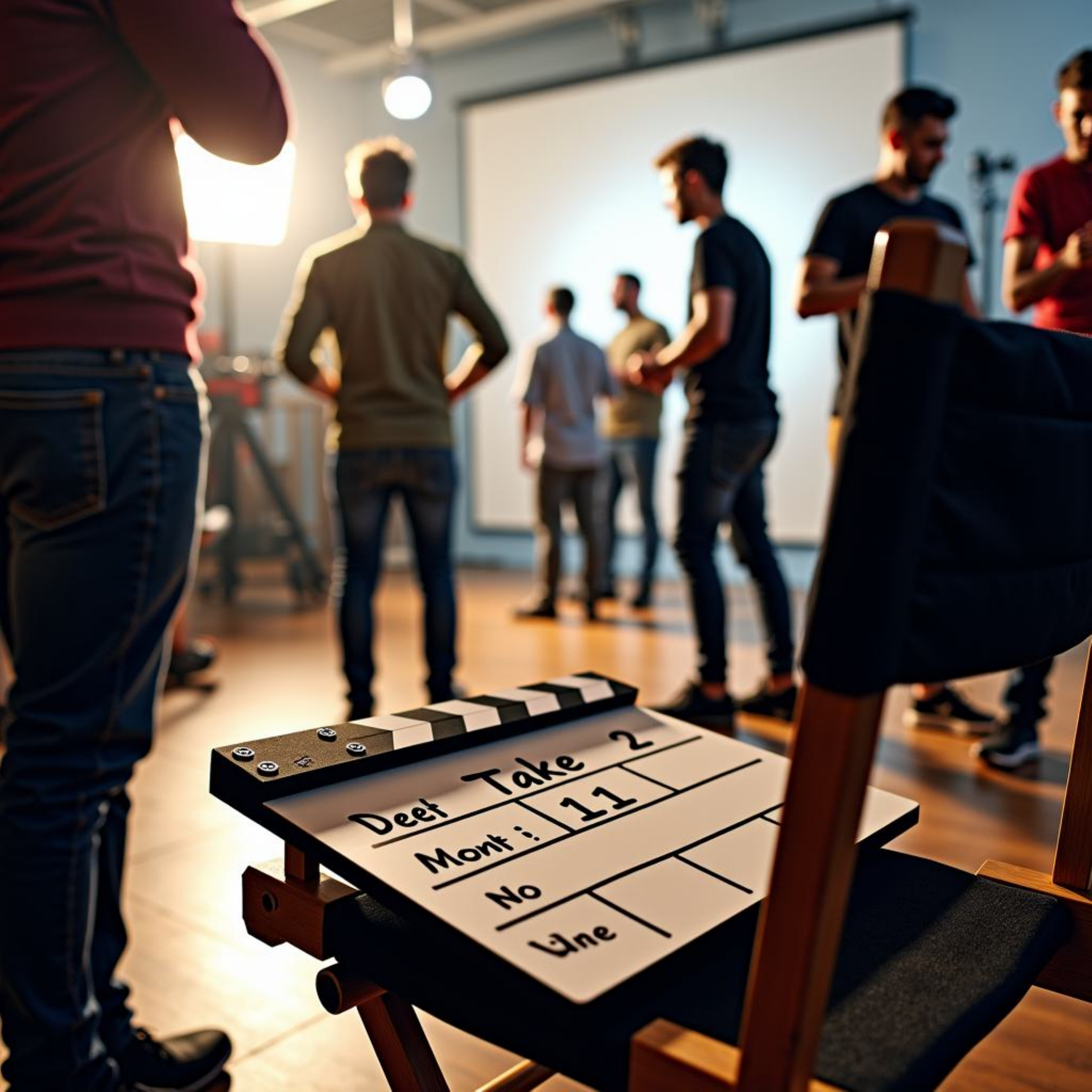}
  \label{fig:tr_r1c3}
\end{subfigure}

\vspace{-10pt} 

\begin{subfigure}[t]{0.315\linewidth}
  \centering
  \includegraphics[width=\linewidth]{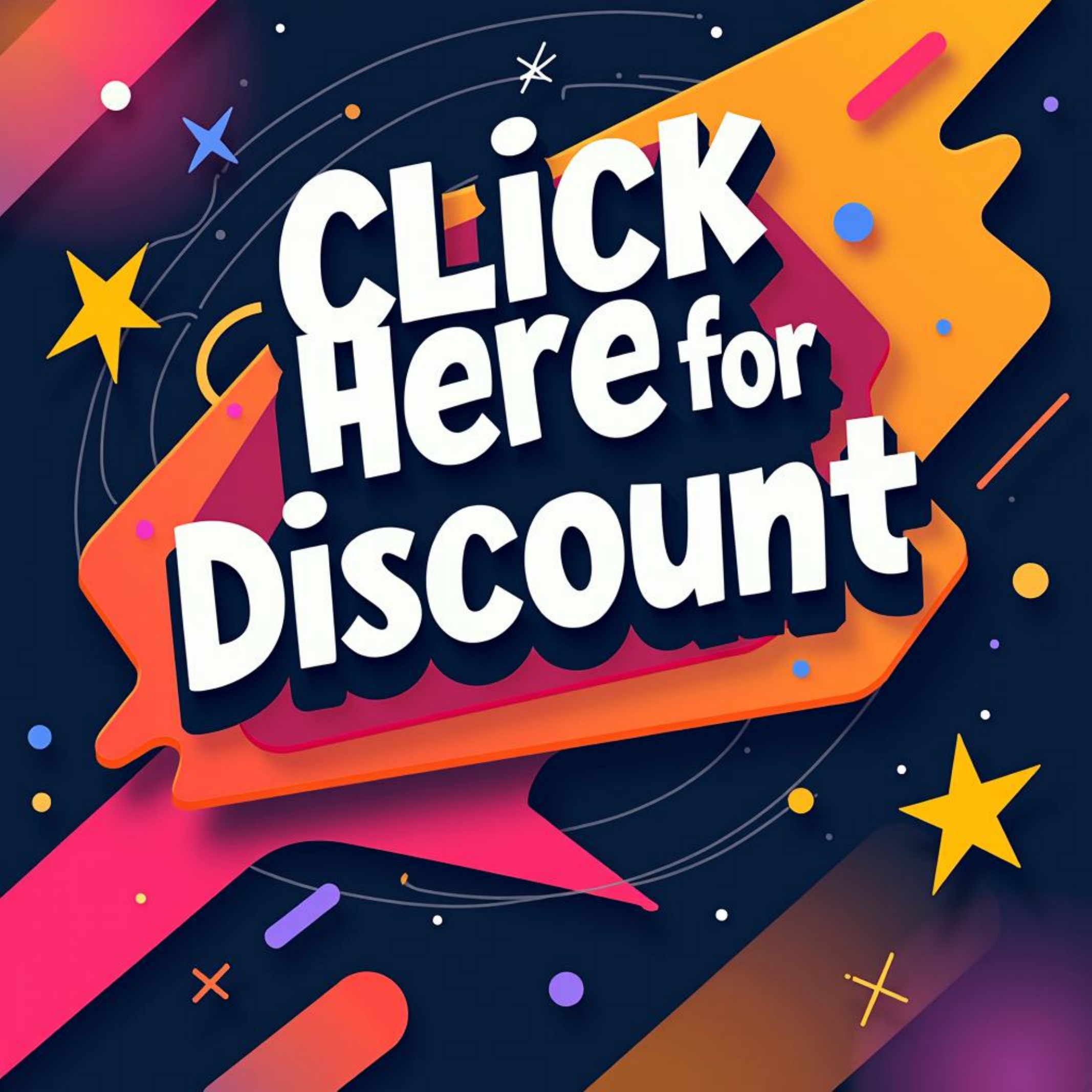}
  \label{fig:tr_r2c1}
\end{subfigure}\hfill
\begin{subfigure}[t]{0.315\linewidth}
  \centering
  \includegraphics[width=\linewidth]{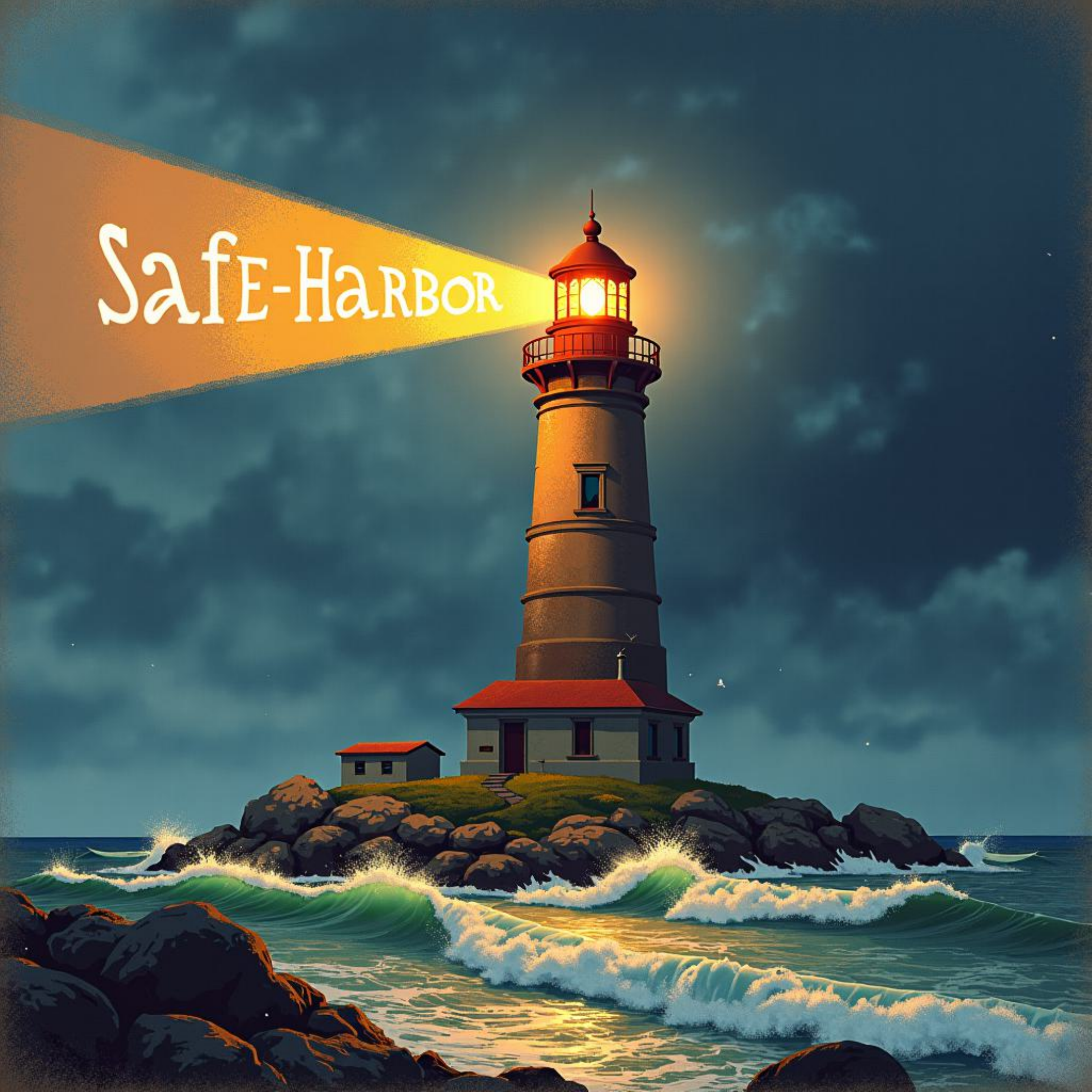}
  \label{fig:tr_r2c2}
\end{subfigure}\hfill
\begin{subfigure}[t]{0.315\linewidth}
  \centering
  \includegraphics[width=\linewidth]{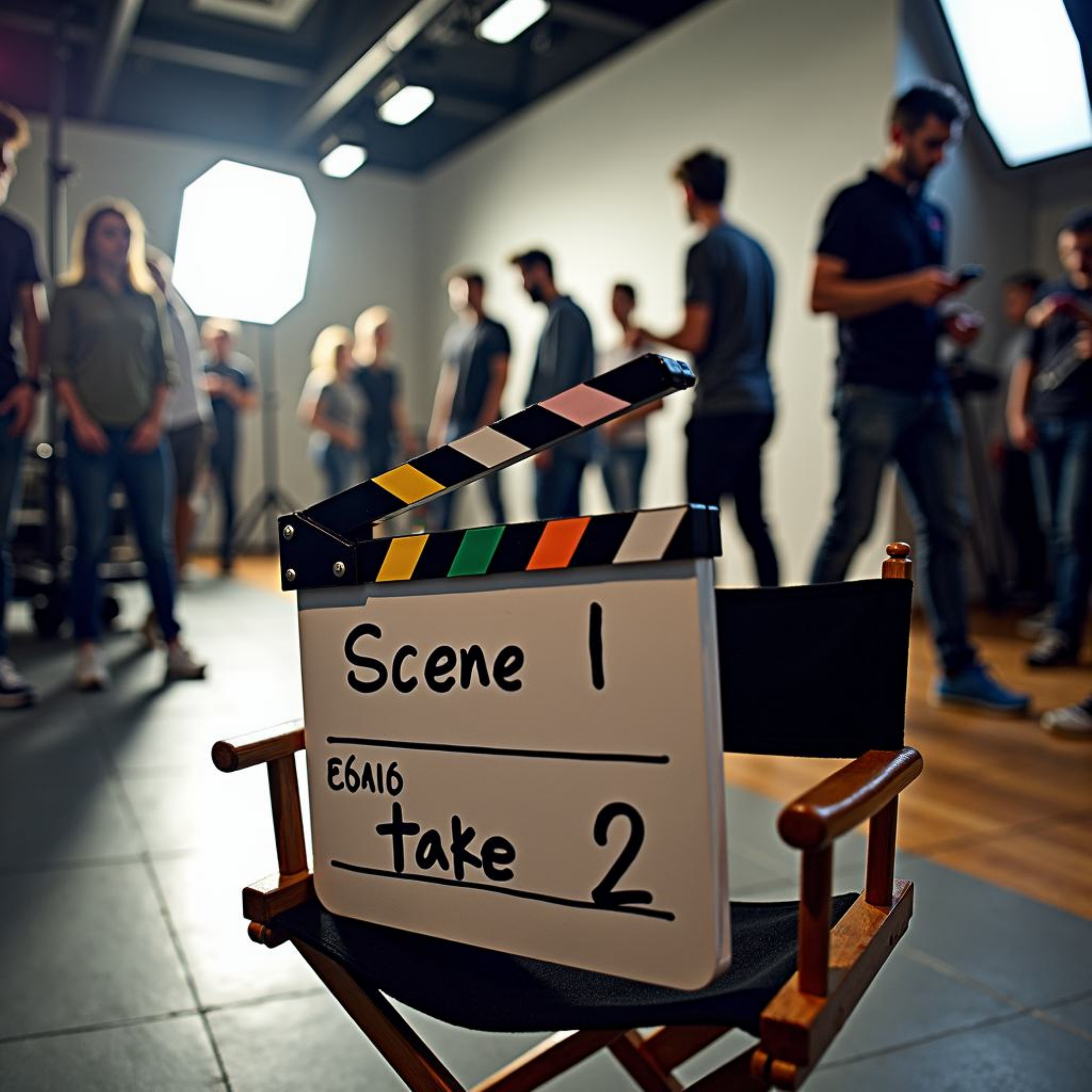}
  \label{fig:tr_r2c3}
\end{subfigure}
\vspace{-10pt}
\caption{\textbf{Rescheduling improves text rendering}. We present comparisons between images generated with default schedules (upper) and our learned schedules (lower) from Flux-Dev.}
\label{fig:text_render}
\vspace{-4pt}
\end{figure}

\begin{figure}[t]
\centering

\colhead{...a cow \color{red}{right of a laptop}}\hfill
\colhead{a photo of {\color{red}three} pizzas}\hfill
\colhead{...bowl and {\color{red}a pink sink}}

\begin{subfigure}[t]{0.315\linewidth}
  \centering  \includegraphics[width=\linewidth]{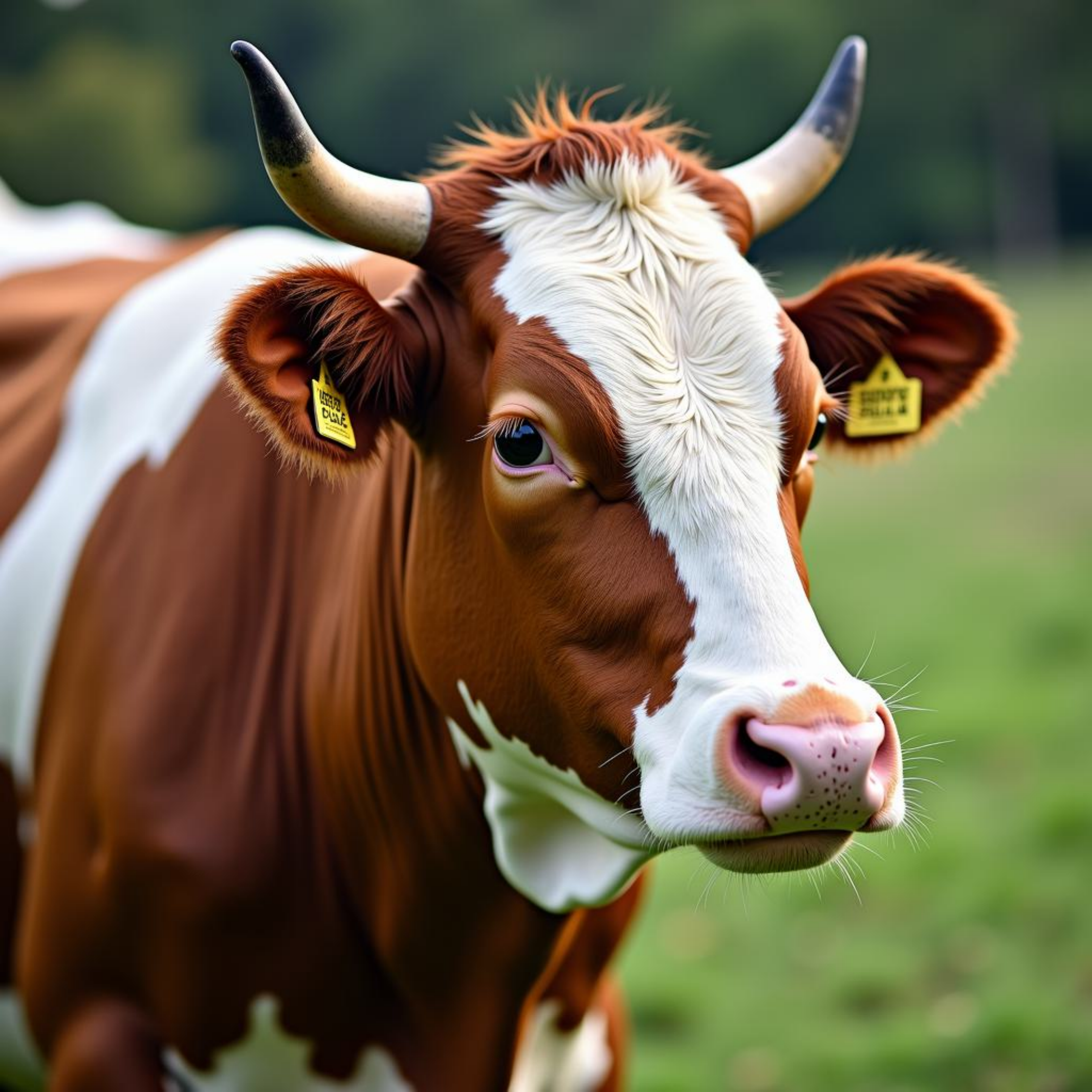}
  \label{fig:ge_r1c1}
\end{subfigure}\hfill
\begin{subfigure}[t]{0.315\linewidth}
  \centering
  \includegraphics[width=\linewidth]{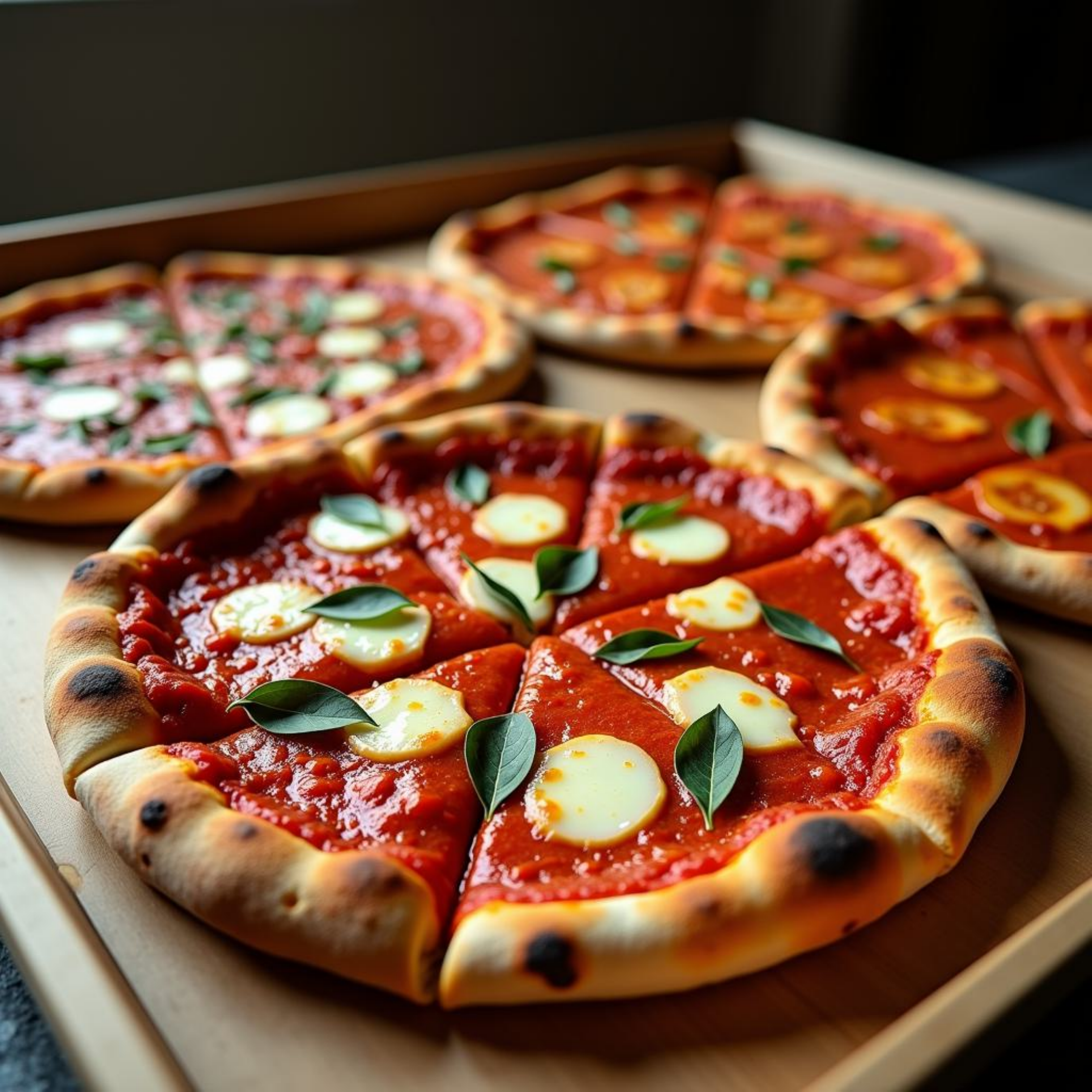}
  \label{fig:ge_r1c2}
\end{subfigure}\hfill
\begin{subfigure}[t]{0.315\linewidth}
  \centering
  \includegraphics[width=\linewidth]{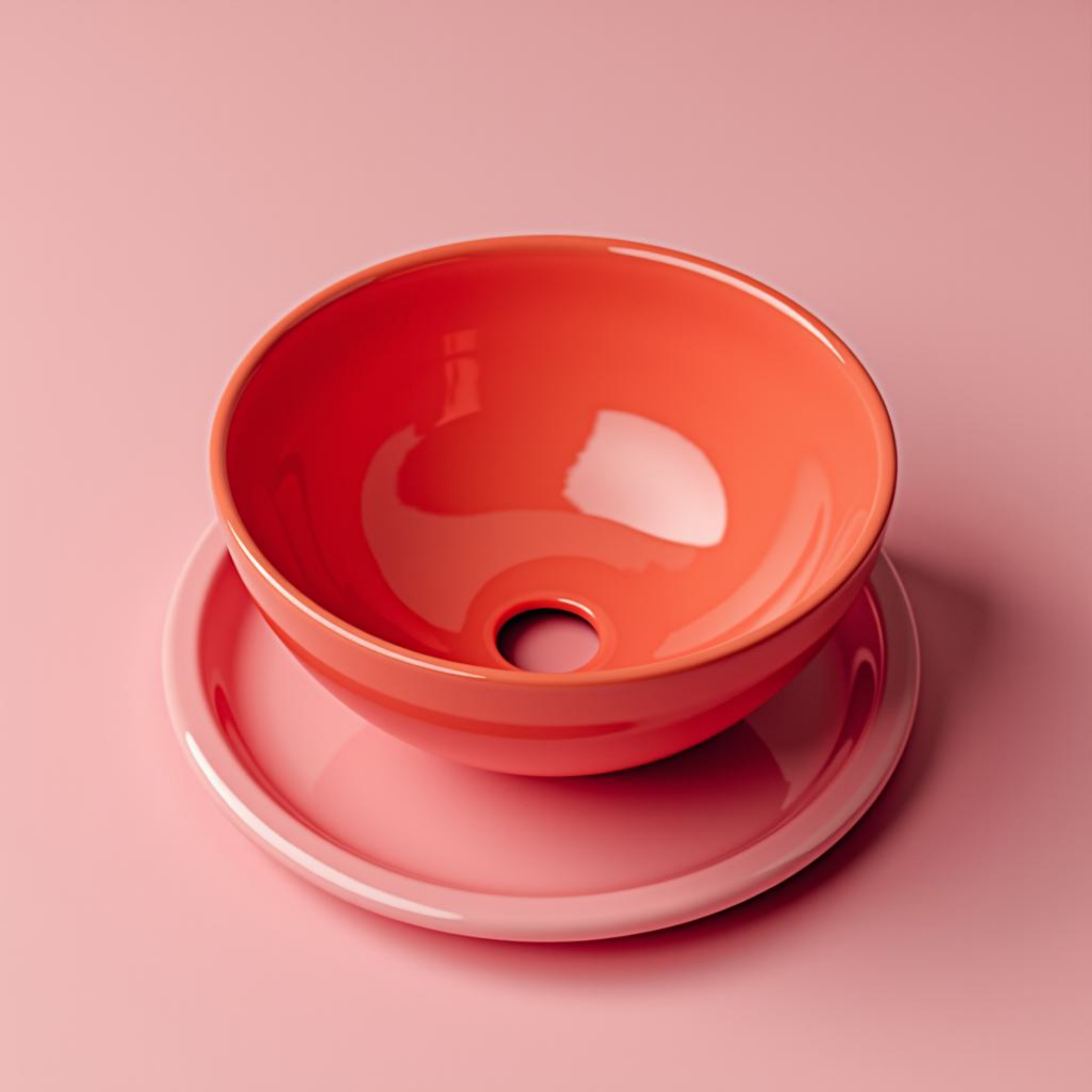}
  \label{fig:ge_r1c3}
\end{subfigure}

\vspace{-10pt} 

\begin{subfigure}[t]{0.315\linewidth}
  \centering
  \includegraphics[width=\linewidth]{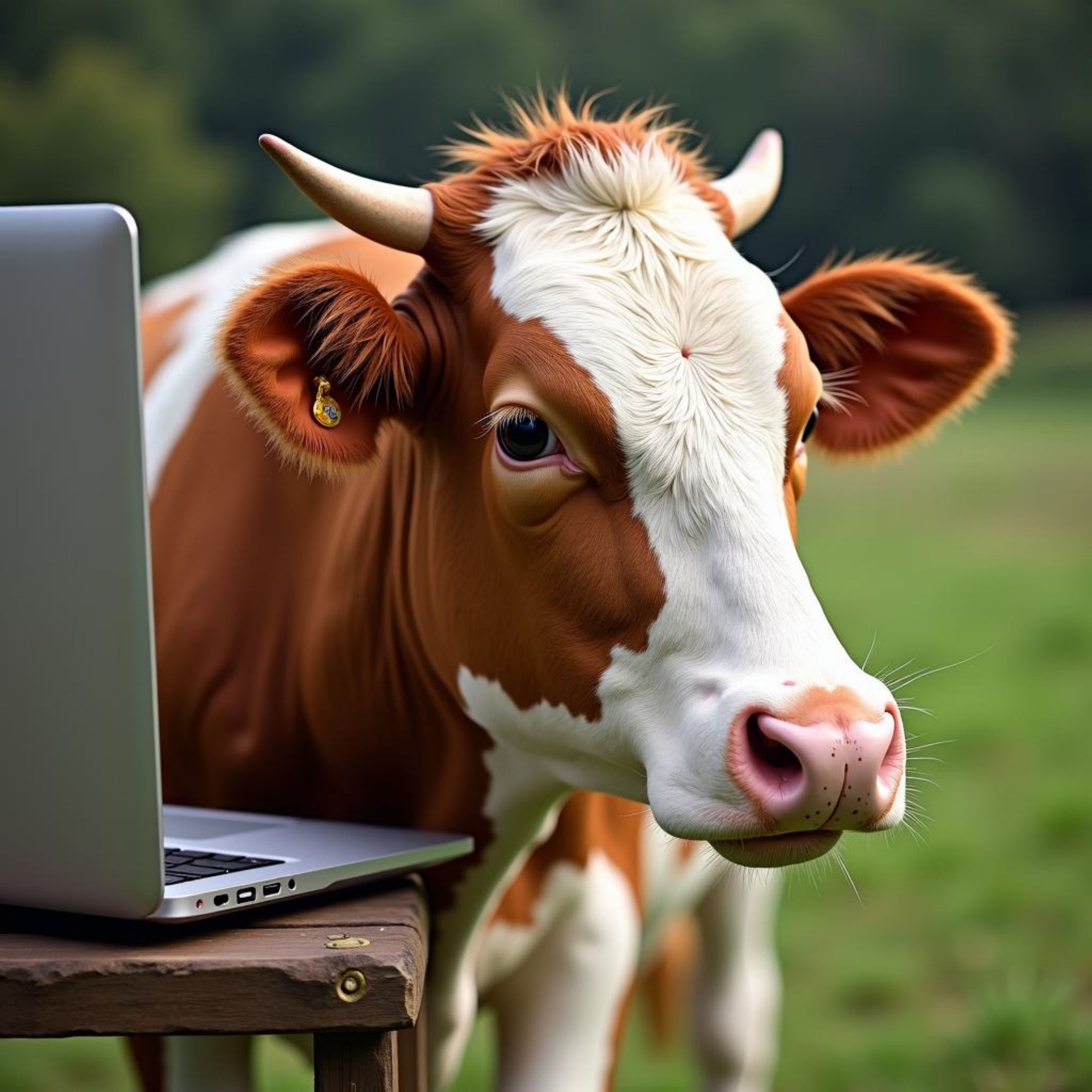}
  \label{fig:ge_r2c1}
\end{subfigure}\hfill
\begin{subfigure}[t]{0.315\linewidth}
  \centering
  \includegraphics[width=\linewidth]{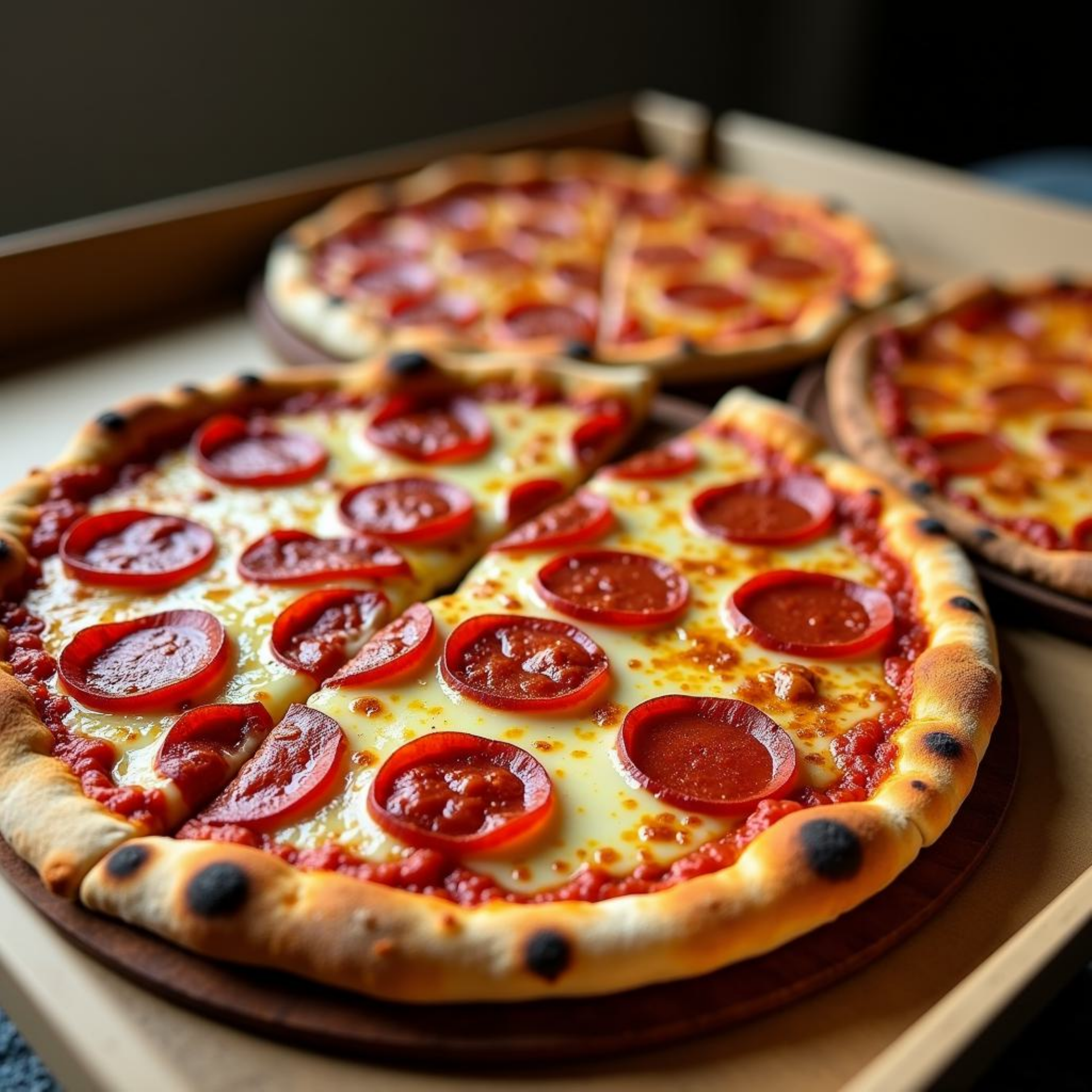}
  \label{fig:ge_r2c2}
\end{subfigure}\hfill
\begin{subfigure}[t]{0.315\linewidth}
  \centering
  \includegraphics[width=\linewidth]{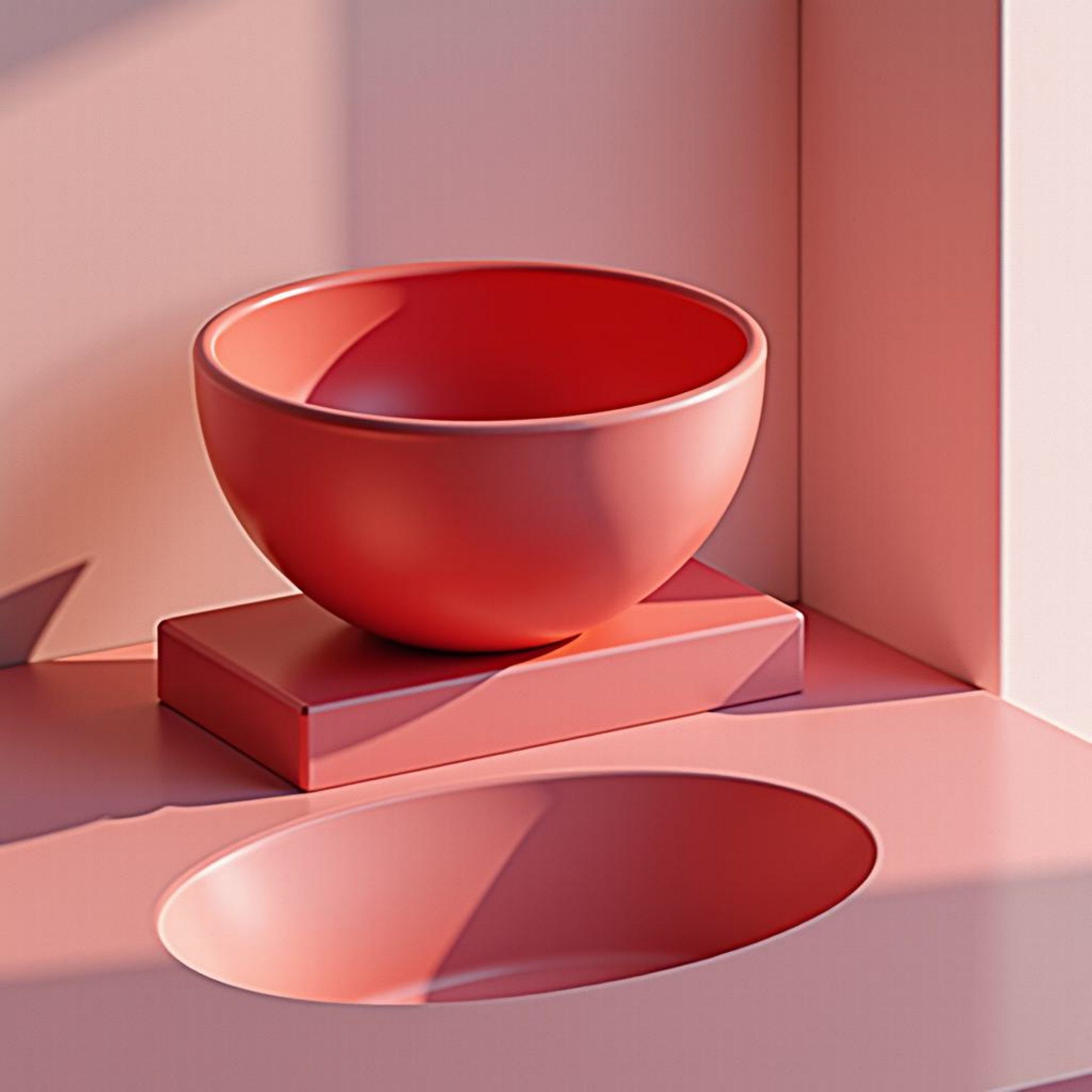}
  \label{fig:ge_r2c3}
\end{subfigure}
\vspace{-10pt}
\caption{\textbf{Rescheduling improves fine grained alignment}. Comparisons between images generated with default schedules (upper) and our learned schedules (lower) from Flux-Dev.}
\label{fig:geneval}
\vspace{-4pt}
\end{figure}

\paragraph{Findings.}
Across both SD-3.5M and Flux-Dev, \emph{Ours (JS)} achieves consistent improvements on fine-grained tasks, as shown in \cref{tab:text-rendering-concise,tab:geneval_two_backbones}. 
The gains are visible in \cref{fig:text_render,fig:geneval}: adaptive timestep reallocation improves stroke completion and reduces character dropout in text rendering, and similarly enhances object-centric correctness in GenEval scenarios. Together, these findings indicate that variance-aware, prompt-adaptive scheduling yields meaningful fine-grained alignment benefits—even when the sampling budget is already sufficient.

\begin{table}[t]
  \centering
  \caption{\textbf{GenEval} results by backbone and schedule. Metrics are fractions in $[0,1]$. c.attrs.\ = Color Attribution.}
  \label{tab:geneval_two_backbones}
  \setlength{\tabcolsep}{4pt}
  \renewcommand{\arraystretch}{1.4}
  \scriptsize
  \begin{tabular}{l c c c c c c c}
    \toprule
    \textbf{Model} & \textbf{Single} & \textbf{Two} & \textbf{Count} & \textbf{Colors} & \textbf{Pos.} & \textbf{C.Attrs.} & \textbf{Overall} \\
    \midrule
    SD3.5-M  & 0.99 & 0.81 & 0.49 & 0.79 & 0.19 & 0.46 & 0.62 \\
    \rowcolor{gray!20}
    SD3.5-M + JS & \textbf{0.99} & \textbf{0.87} & \textbf{0.65} & \textbf{0.86} & \textbf{0.22} & \textbf{0.46} & \textbf{0.68} \\
    \midrule
    Flux-Dev & 0.98 & 0.83 & 0.58 & 0.79 & 0.22 & 0.45 & 0.64 \\
    \rowcolor{gray!20}
    Flux-Dev + JS & \textbf{0.99} & \textbf{0.88} & \textbf{0.77} & \textbf{0.86} & \textbf{0.24} & \textbf{0.46} & \textbf{0.70} \\
    \bottomrule
  \end{tabular}%
\end{table}

\section{Conclusion}
We introduced \emph{instance-level rescheduling} for text-to-image samplers: a policy that outputs the entire denoising schedule in one pass, for each prompt and noise sample. Framing schedule learning as policy optimization, we derived a \emph{James--Stein (JS) shrinkage} baseline that yields an unbiased, variance-reduced gradient estimator to facilitate training. Empirically, our learned schedules further improves text–image alignment.

\paragraph{Limitations and outlook.}
We used simple scheduler architectures and limited reward types (HPSv2, Rule-based GenEval Rewards), leaving richer policies and broader objectives for future work. Promising extensions include: (i) adaptive early stopping for dynamic step budgets; (ii) multi-objective or process-based rewards (e.g., compositionality, aesthetics, safety); (iii) extending JS-based shrinkage to RLHF and other long-horizon policy settings and applications to video, flow, and 3D generative pipelines. These directions can further enhance the stability, efficiency, and controllability of post-training generative systems.

\section*{Acknowledgements}
We would like to thank Yeqing Li and Eugene Ie for reviewing this work and leaving valuable comments. We would also like to thank Keyang Xu and other Google team members for valuable discussions and suggestions during the early stage of this project.

\newpage
\clearpage
{
    \small
    \bibliographystyle{ieeenat_fullname}
    \bibliography{main}

\begin{thebibliography}{31}
\providecommand{\natexlab}[1]{#1}
\providecommand{\url}[1]{\texttt{#1}}
\expandafter\ifx\csname urlstyle\endcsname\relax
  \providecommand{\doi}[1]{doi: #1}\else
  \providecommand{\doi}{doi: \begingroup \urlstyle{rm}\Url}\fi

\bibitem[Ahmadian et~al.(2024)Ahmadian, Cremer, Gall{\'e}, Fadaee, Kreutzer, Pietquin, {\"U}st{\"u}n, and Hooker]{ahmadian2024back}
Arash Ahmadian, Chris Cremer, Matthias Gall{\'e}, Marzieh Fadaee, Julia Kreutzer, Olivier Pietquin, Ahmet {\"U}st{\"u}n, and Sara Hooker.
\newblock Back to basics: Revisiting reinforce-style optimization for learning from human feedback in llms.
\newblock In \emph{Proceedings of the 62nd Annual Meeting of the Association for Computational Linguistics (Volume 1: Long Papers)}, pages 12248--12267, 2024.

\bibitem[Black et~al.(2023)Black, Janner, Du, Kostrikov, and Levine]{black2023training}
Kevin Black, Michael Janner, Yilun Du, Ilya Kostrikov, and Sergey Levine.
\newblock Training diffusion models with reinforcement learning.
\newblock \emph{arXiv preprint arXiv:2305.13301}, 2023.

\bibitem[Esser et~al.(2024)Esser, Kulal, Blattmann, Entezari, M{\"u}ller, Saini, Levi, Lorenz, Sauer, Boesel, et~al.]{esser2024scaling}
Patrick Esser, Sumith Kulal, Andreas Blattmann, Rahim Entezari, Jonas M{\"u}ller, Harry Saini, Yam Levi, Dominik Lorenz, Axel Sauer, Frederic Boesel, et~al.
\newblock Scaling rectified flow transformers for high-resolution image synthesis.
\newblock In \emph{Forty-first international conference on machine learning}, 2024.

\bibitem[Fan et~al.(2023)Fan, Watkins, Du, Liu, Ryu, Boutilier, Abbeel, Ghavamzadeh, Lee, and Lee]{fan2023dpok}
Ying Fan, Olivia Watkins, Yuqing Du, Hao Liu, Moonkyung Ryu, Craig Boutilier, Pieter Abbeel, Mohammad Ghavamzadeh, Kangwook Lee, and Kimin Lee.
\newblock Dpok: Reinforcement learning for fine-tuning text-to-image diffusion models.
\newblock \emph{Advances in Neural Information Processing Systems}, 36:\penalty0 79858--79885, 2023.

\bibitem[Ghosh et~al.(2023)Ghosh, Hajishirzi, and Schmidt]{ghosh2023geneval}
Dhruba Ghosh, Hannaneh Hajishirzi, and Ludwig Schmidt.
\newblock Geneval: An object-focused framework for evaluating text-to-image alignment.
\newblock \emph{Advances in Neural Information Processing Systems}, 36:\penalty0 52132--52152, 2023.

\bibitem[Greensmith et~al.(2004)Greensmith, Bartlett, and Baxter]{greensmith2004variance}
Evan Greensmith, Peter~L Bartlett, and Jonathan Baxter.
\newblock Variance reduction techniques for gradient estimates in reinforcement learning.
\newblock \emph{Journal of Machine Learning Research}, 5\penalty0 (Nov):\penalty0 1471--1530, 2004.

\bibitem[James et~al.(1961)James, Stein, et~al.]{james1961estimation}
William James, Charles Stein, et~al.
\newblock Estimation with quadratic loss.
\newblock In \emph{Proceedings of the fourth Berkeley symposium on mathematical statistics and probability}, pages 361--379. University of California Press, 1961.

\bibitem[Karras et~al.(2022)Karras, Aittala, Aila, and Laine]{karras2022elucidating}
Tero Karras, Miika Aittala, Timo Aila, and Samuli Laine.
\newblock Elucidating the design space of diffusion-based generative models.
\newblock \emph{Advances in neural information processing systems}, 35:\penalty0 26565--26577, 2022.

\bibitem[Kool et~al.(2019)Kool, van Hoof, and Welling]{kool2019buy}
Wouter Kool, Herke van Hoof, and Max Welling.
\newblock Buy 4 reinforce samples, get a baseline for free!
\newblock 2019.

\bibitem[Labs(2024)]{flux2024}
Black~Forest Labs.
\newblock Flux.
\newblock \url{https://github.com/black-forest-labs/flux}, 2024.

\bibitem[Liao et~al.(2020)Liao, Pang, Huang, Hassner, and Bai]{liao2020mask}
Minghui Liao, Guan Pang, Jing Huang, Tal Hassner, and Xiang Bai.
\newblock Mask textspotter v3: Segmentation proposal network for robust scene text spotting.
\newblock In \emph{European conference on computer vision}, pages 706--722. Springer, 2020.

\bibitem[Lipman et~al.(2022)Lipman, Chen, Ben-Hamu, Nickel, and Le]{lipman2022flow}
Yaron Lipman, Ricky~TQ Chen, Heli Ben-Hamu, Maximilian Nickel, and Matt Le.
\newblock Flow matching for generative modeling.
\newblock \emph{arXiv preprint arXiv:2210.02747}, 2022.

\bibitem[Liu et~al.(2025)Liu, Liu, Liang, Li, Liu, Wang, Wan, Zhang, and Ouyang]{liu2025flow}
Jie Liu, Gongye Liu, Jiajun Liang, Yangguang Li, Jiaheng Liu, Xintao Wang, Pengfei Wan, Di Zhang, and Wanli Ouyang.
\newblock Flow-grpo: Training flow matching models via online rl.
\newblock \emph{arXiv preprint arXiv:2505.05470}, 2025.

\bibitem[Liu et~al.(2022)Liu, Gong, and Liu]{liu2022flow}
Xingchao Liu, Chengyue Gong, and Qiang Liu.
\newblock Flow straight and fast: Learning to generate and transfer data with rectified flow.
\newblock \emph{arXiv preprint arXiv:2209.03003}, 2022.

\bibitem[Loshchilov and Hutter(2017)]{loshchilov2017decoupled}
Ilya Loshchilov and Frank Hutter.
\newblock Decoupled weight decay regularization.
\newblock \emph{arXiv preprint arXiv:1711.05101}, 2017.

\bibitem[Lu et~al.(2022)Lu, Zhou, Bao, Chen, Li, and Zhu]{lu2022dpm}
Cheng Lu, Yuhao Zhou, Fan Bao, Jianfei Chen, Chongxuan Li, and Jun Zhu.
\newblock Dpm-solver: A fast ode solver for diffusion probabilistic model sampling in around 10 steps.
\newblock \emph{Advances in neural information processing systems}, 35:\penalty0 5775--5787, 2022.

\bibitem[Podell et~al.(2023)Podell, English, Lacey, Blattmann, Dockhorn, M{\"u}ller, Penna, and Rombach]{podell2023sdxl}
Dustin Podell, Zion English, Kyle Lacey, Andreas Blattmann, Tim Dockhorn, Jonas M{\"u}ller, Joe Penna, and Robin Rombach.
\newblock Sdxl: Improving latent diffusion models for high-resolution image synthesis.
\newblock \emph{arXiv preprint arXiv:2307.01952}, 2023.

\bibitem[Rombach et~al.(2022)Rombach, Blattmann, Lorenz, Esser, and Ommer]{rombach2022high}
Robin Rombach, Andreas Blattmann, Dominik Lorenz, Patrick Esser, and Bj{\"o}rn Ommer.
\newblock High-resolution image synthesis with latent diffusion models.
\newblock In \emph{Proceedings of the IEEE/CVF conference on computer vision and pattern recognition}, pages 10684--10695, 2022.

\bibitem[Sabour et~al.(2024)Sabour, Fidler, and Kreis]{sabour2024align}
Amirmojtaba Sabour, Sanja Fidler, and Karsten Kreis.
\newblock Align your steps: Optimizing sampling schedules in diffusion models.
\newblock \emph{arXiv preprint arXiv:2404.14507}, 2024.

\bibitem[Salimans and Ho(2022)]{salimans2022progressive}
Tim Salimans and Jonathan Ho.
\newblock Progressive distillation for fast sampling of diffusion models.
\newblock \emph{arXiv preprint arXiv:2202.00512}, 2022.

\bibitem[Song et~al.(2020)Song, Meng, and Ermon]{song2020denoising}
Jiaming Song, Chenlin Meng, and Stefano Ermon.
\newblock Denoising diffusion implicit models.
\newblock \emph{arXiv preprint arXiv:2010.02502}, 2020.

\bibitem[Song et~al.(2023)Song, Dhariwal, Chen, and Sutskever]{song2023consistency}
Yang Song, Prafulla Dhariwal, Mark Chen, and Ilya Sutskever.
\newblock Consistency models.
\newblock In \emph{Proceedings of the 40th International Conference on Machine Learning}, pages 32211--32252, 2023.

\bibitem[Wallace et~al.(2024)Wallace, Dang, Rafailov, Zhou, Lou, Purushwalkam, Ermon, Xiong, Joty, and Naik]{wallace2024diffusion}
Bram Wallace, Meihua Dang, Rafael Rafailov, Linqi Zhou, Aaron Lou, Senthil Purushwalkam, Stefano Ermon, Caiming Xiong, Shafiq Joty, and Nikhil Naik.
\newblock Diffusion model alignment using direct preference optimization.
\newblock In \emph{Proceedings of the IEEE/CVF Conference on Computer Vision and Pattern Recognition}, pages 8228--8238, 2024.

\bibitem[Williams(1992)]{williams1992simple}
Ronald~J Williams.
\newblock Simple statistical gradient-following algorithms for connectionist reinforcement learning.
\newblock \emph{Machine learning}, 8\penalty0 (3):\penalty0 229--256, 1992.

\bibitem[Wu et~al.(2023)Wu, Hao, Sun, Chen, Zhu, Zhao, and Li]{wu2023human}
Xiaoshi Wu, Yiming Hao, Keqiang Sun, Yixiong Chen, Feng Zhu, Rui Zhao, and Hongsheng Li.
\newblock Human preference score v2: A solid benchmark for evaluating human preferences of text-to-image synthesis.
\newblock \emph{arXiv preprint arXiv:2306.09341}, 2023.

\bibitem[Xie et~al.(2024)Xie, Xiao, Kingma, Hou, Wu, Murphy, Salimans, Poole, and Gao]{xie2024distillation}
Sirui Xie, Zhisheng Xiao, Diederik~P Kingma, Tingbo Hou, Ying~N Wu, Kevin Murphy, Tim Salimans, Ben Poole, and Ruiqi Gao.
\newblock Em distillation for one-step diffusion models.
\newblock \emph{Advances in Neural Information Processing Systems}, 37:\penalty0 45073--45104, 2024.

\bibitem[Ye et~al.(2025)Ye, Chen, Li, Huang, Luo, and Qi]{ye2025schedule}
Zilyu Ye, Zhiyang Chen, Tiancheng Li, Zemin Huang, Weijian Luo, and Guo-Jun Qi.
\newblock Schedule on the fly: Diffusion time prediction for faster and better image generation.
\newblock In \emph{Proceedings of the Computer Vision and Pattern Recognition Conference}, pages 23412--23422, 2025.

\bibitem[Yin et~al.(2024)Yin, Gharbi, Zhang, Shechtman, Durand, Freeman, and Park]{yin2024one}
Tianwei Yin, Micha{\"e}l Gharbi, Richard Zhang, Eli Shechtman, Fredo Durand, William~T Freeman, and Taesung Park.
\newblock One-step diffusion with distribution matching distillation.
\newblock In \emph{Proceedings of the IEEE/CVF conference on computer vision and pattern recognition}, pages 6613--6623, 2024.

\bibitem[Yu et~al.()Yu, Zhang, He, Ma, Xie, Miao, Lu, Zhang, Kong, Gao, et~al.]{yunoisier}
Peiyu Yu, Dinghuai Zhang, Hengzhi He, Xiaojian Ma, Sirui Xie, Ruiyao Miao, Yifan Lu, Yasi Zhang, Deqian Kong, Ruiqi Gao, et~al.
\newblock ``noisier'’noise contrastive estimation is (almost) maximum likelihood.
\newblock In \emph{The Fourteenth International Conference on Learning Representations}.

\bibitem[Zhao et~al.(2023)Zhao, Bai, Rao, Zhou, and Lu]{zhao2023unipc}
Wenliang Zhao, Lujia Bai, Yongming Rao, Jie Zhou, and Jiwen Lu.
\newblock Unipc: A unified predictor-corrector framework for fast sampling of diffusion models.
\newblock \emph{Advances in Neural Information Processing Systems}, 36:\penalty0 49842--49869, 2023.

\bibitem[Zhou et~al.(2025)Zhou, Shao, Bai, Zhang, Xu, Han, and Xie]{zhou2025golden}
Zikai Zhou, Shitong Shao, Lichen Bai, Shufei Zhang, Zhiqiang Xu, Bo Han, and Zeke Xie.
\newblock Golden noise for diffusion models: A learning framework.
\newblock In \emph{Proceedings of the IEEE/CVF International Conference on Computer Vision}, pages 17688--17697, 2025.

\end{thebibliography}
}

\newpage
\clearpage
\clearpage
\setcounter{page}{1}
\maketitlesupplementary

\setcounter{section}{0}
\renewcommand{\thesection}{\Alph{section}}
\renewcommand{\thesubsection}{\thesection.\arabic{subsection}}
\renewcommand{\thesubsubsection}{\thesubsection.\arabic{subsubsection}}


\section{Proofs}

\subsection{Proof of Proposition~\ref{prop:opt-baseline}.}
\begin{proof}
First, we show the unbiasedness of the baseline.
We abbreviate the policy as $\pi_\theta(\tau) \equiv \pi_\theta(\tau \mid \mathbf{x}_T,\mathbf{c})$,
\[
g(\tau) \;=\; \nabla_\theta \log \pi_\theta(\tau),
\qquad
w(\tau) \;=\; \norm{g(\tau)}^2,
\]
and let $r(\tau)$ denote the scalar reward.
The single–sample REINFORCE estimator with a scalar baseline $b$ is
\begin{equation}
\widehat{G}(b) \;=\; (r(\tau)-b)\,g(\tau).
\end{equation}

\noindent
By swapping integral and gradient operators
\begin{equation}
\E_{\tau\sim\pi_\theta}[g(\tau)]
= \E_{\tau\sim\pi_\theta}[\nabla_\theta \log \pi_\theta(\tau)]
= \nabla_\theta \E_{\tau\sim\pi_\theta}[1]
= 0.
\end{equation}
Hence the estimator \textit{remains unbiased} for any baseline $b$:
\begin{equation}
\E[\widehat{G}(b)]
= \E[(r-b)g]
= \E[r g] - b\,\E[g]
= \E[r g].
\end{equation}

Next, we study the variances of the estimator. The covariance of $\widehat{G}(b)$ is
\[
\Var[\widehat{G}(b)]
= \E[(r-b)^2 g g^\top]
  - \E[(r-b)g]\,\E[(r-b)g]^\top.
\]
Since $\E[(r-b)g]=\E[r g]$ does not depend on $b$, minimizing the total scalar
variance (trace) is equivalent to minimizing
\[
V(b)
\;=\; \E\big[(r-b)^2\,w(\tau)\big]
\;-\; \text{const},
~
{\rm Recall}~w(\tau)=\norm{g(\tau)}^2.
\]

The function $V(b)$ is differentiable and strictly convex in $b$.
Differentiating and setting the derivative to zero yields
\begin{align*}
\frac{\mathrm{d}}{\mathrm{d}b} V(b)
&= \frac{\mathrm{d}}{\mathrm{d}b}\,
   \E\big[(r-b)^2 w\big]
 = \E\big[2(b-r) w\big] \\
&= 2\Big(b\,\E[w] - \E[r w]\Big) = 0,
\end{align*}
which gives the unique minimizer
\begin{equation}
b^{*}
= \frac{\E[r(\tau)\,w(\tau)]}{\E[w(\tau)]}
= \frac{\E\big[r(\tau)\,\norm{\nabla_\theta \log \pi_\theta(\tau)}^2\big]}
       {\E\big[\norm{\nabla_\theta \log \pi_\theta(\tau)}^2\big]}.
\label{eq:opt-baseline-proof}
\end{equation}
Thus $b^{*}$ minimizes the trace of $\Var[\widehat{G}(b)]$.
\end{proof}

\medskip
\paragraph{Remarks.}
(i) Because $b$ is a scalar shared across all components of $g(\tau)$, the same
minimizer $b^{*}$ arises whether one minimizes each component variance or the
trace. (ii) Our derivation shares the spirit of state-dependent baselines derived for GPOMDP in \citet{greensmith2004variance}, which optimize a different variance criterion conditioned on Markov states.
(iii) When the policy becomes nearly deterministic for a fixed context
$(\mathbf{x}_T,\mathbf{c})$, the variation of
$\nabla_\theta \log \pi_\theta(\tau)$ is small. In this regime,
Eq.~\eqref{eq:opt-baseline-proof} reduces to the mean reward
$\E[r(\tau)\mid \mathbf{x}_T,\mathbf{c}]$, which motivates the empirical
approximations used in the main text.

\subsection{Proof of Theorem~\ref{thm:js-mse}.}
\begin{proof}
Now we prove the theoretical results related to the James-Stein reward baseline. 

\paragraph{Bayesian optimality.}
Recall that we have the random–effects model in Eq.~\eqref{eq:random-effects},
\begin{equation}
r^{(c,i)} \mid \mu_c \sim \mathcal{N}(\mu_c,\sigma^2),
\qquad
\mu_c \sim \mathcal{N}(\mu_0,\delta^2),
\end{equation}
the per–context mean satisfies
\begin{equation}
\bar r_c \;=\; \frac{1}{K_c}\sum_{i=1}^{K_c} r^{(c,i)}, 
\qquad
\bar r_c \mid \mu_c \sim \mathcal{N}(\mu_c,\sigma^2/K_c).
\end{equation}
By normal–normal conjugacy, the posterior distribution
$\mu_c \mid \{r^{(c,i)}\}_{i=1}^{K_c}$ is Gaussian with mean
\begin{equation}
\E[\mu_c \mid \{r^{(c,i)}\}]
= (1-\alpha_c^*)\,\bar r_c + \alpha_c^*\,\mu_0,
\label{eq:post-mean-mu}
\end{equation}
where the (population) shrinkage coefficient is
\begin{equation}
\alpha_c^*
= \frac{\sigma^2 / K_c}{\sigma^2 / K_c + \delta^2}.
\label{eq:alpha}
\end{equation}
For any data–dependent predictor $t_c$ of $\mu_c$, the posterior
expected squared error given the observed rewards is
\begin{equation}
\begin{aligned}
\E\big[(\mu_c - t_c)^2 \,\big|\, \{r^{(c,i)}\}\big]
&= \E[\mu_c^2 \mid \{r^{(c,i)}\}] \\
&\quad - 2t_c\,\E[\mu_c \mid \{r^{(c,i)}\}] + t_c^2,
\end{aligned}
\end{equation}
which, as a quadratic in $t_c$, is minimized at
$t_c = \E[\mu_c \mid \{r^{(c,i)}\}]$.
Thus the Bayes–optimal predictor of $\mu_c$ under squared error is the
posterior mean in Eq.~\eqref{eq:post-mean-mu}, i.e., the convex combination
of the within–context mean $\bar r_c$ and the population mean $\mu_0$.

In our REINFORCE setting, we require a baseline that is independent of the
current rollout $(c,i)$ to preserve unbiasedness. We therefore replace
$\bar r_c$ and the global mean $\mu_0$ by their leave–one–out analogues:
\begin{equation}
\begin{aligned}
\bar r_c^{(-i)}
&= \frac{1}{K_c-1}\sum_{j\neq i} r^{(c,j)}, \\
\bar r_{..}^{(-c,-i)}
&= \frac{1}{(\sum_{c'} K_{c'}) - 1}
  \Bigg(
  \sum_{c'=1}^{B}\sum_{j=1}^{K_{c'}} r^{(c',j)} - r^{(c,i)}
  \Bigg).
\end{aligned}
\end{equation}
The ideal Bayes baseline for rollout $(c,i)$ is then
\[
b_{\mathrm{Bayes}}^{(c,i)}
= (1-\alpha_c^*)\,\bar r_c^{(-i)}
  + \alpha_c^*\,\bar r_{..}^{(-c,-i)}.
\]

In practice, $\sigma^2$ and $\delta^2$ are unknown. The method–of–moments
estimators in Eq.~\eqref{eq:var-est} yield empirical counterparts
$\widehat{\sigma}^2$ and $\widehat{\delta}^2$, which in turn define the
empirical shrinkage weights
\[
\widehat{\alpha}_c
= \frac{\widehat{\sigma}^2/(K_c-1)}
       {\widehat{\sigma}^2/(K_c-1)+\widehat{\delta}^2}.
\]
Substituting $\widehat{\alpha}_c$ and the leave–one–out means into the
Bayes form above recovers exactly the empirical James--Stein baseline
\[
b_{\mathrm{JS}}^{(c,i)}
= (1-\widehat{\alpha}_c)\,b_{\mathrm{RLOO}}^{(c,i)}
  + \widehat{\alpha}_c\,b_{\mathrm{xctx}}^{(c,i)}
\]
from Eq.~\eqref{eq:js-baseline}, where both
$b_{\mathrm{RLOO}}^{(c,i)}$ and $b_{\mathrm{xctx}}^{(c,i)}$
are computed in a leave–one–out way that excludes $(c,i)$.
Since $b_{\mathrm{JS}}^{(c,i)}$ does not depend on the current reward
$r^{(c,i)}$, the REINFORCE gradient remains unbiased. 
In summary, $b_{\mathrm{JS}}^{(c,i)}$ by construction coincides with the
empirical Bayes posterior mean of $\mu_c$ under the model
\eqref{eq:random-effects}, and is Bayes–optimal.

\paragraph{MSE improvement.}
Under the same random–effects model, consider the convex estimator
\[
\tilde\mu_c(\alpha)
= (1-\alpha)\,\bar r_c^{(-i)} + \alpha\,\bar r_{..}^{(-c,-i)}.
\]
Assuming independence between the within–context and across–context noise,
its mean–squared error satisfies
\[
\MSE(\tilde\mu_c(\alpha))
= (1-\alpha)^2\,\frac{\sigma^2}{K_c-1}
  + \alpha^2\,\delta^2,
\]
which is minimized at
\[
\alpha^\star
= \frac{\sigma^2/(K_c-1)}{\sigma^2/(K_c-1)+\delta^2}.
\]
Thus the same shrinkage coefficient $\alpha^\star$ that defines the
posterior mean also uniquely minimizes the frequentist MSE of the baseline,
strictly improving not only over the purely contextual baseline (RLOO), but also over all the convex combination between RLOO and the XCTX baselines including themselves.
\end{proof}

\paragraph{Remarks.}
If we collect the context-specific means into the vector
$\boldsymbol{\mu} = (\mu_1,\dots,\mu_B)$ and likewise stack their noisy
estimators, we obtain the classical homoscedastic normal-mean problem in
dimension $B$.  In that setting, James--Stein shrinkage estimators that
shrink the vector of means toward a common target are known to uniformly
dominate the unshrunk sample mean in total squared error risk whenever
the dimension is at least three; see, e.g., \citet{james1961estimation}.
In our notation, the ``dimension'' plays the role of the number of
contexts $B$, which is the reason for the $B \ge 3$ condition mentioned
in the main text.  Our empirical Bayes baseline has the same
shrinkage-to-a-global-mean structure as these classical estimators, but
here we focus on the random--effects and REINFORCE viewpoints rather than
re-proving the classical domination result.

\section{Scheduler Network Architectures}
We summarize the network architecture we used for implementing the scheduler network in \cref{tab:arch_scheduler_dirichlet}. The scheduler is $0.1–1\%$ the size ($\sim$20M parameters) of the backbone networks used in Stable Diffusion and Flux model families, and incurs at most $\sim0.1–0.3\%$ extra end-to-end sampling compute in the step counts we used ranging from 5 to 80.

\begin{table*}[t]
    \caption{\textbf{Network architecture for the Dirichlet schedule policy $\alpha_\theta(x, c)$.} 
    The network takes an image-shaped noise tensor $x$, a sequence of text embeddings $e_{\text{text}}$, 
    and an optional pooled text embedding $\bar e_{\text{text}}$ (consistent with SD-XL, SD3.5M/L and FLUX models), and outputs Dirichlet parameters 
    $\boldsymbol{\alpha} \in \mathbb{R}_+^{L+1}$ over $L$ schedule intervals plus one skipped interval.
    Default network architecture hyperparameters: \texttt{text\_embed\_dim} $d_{\rm text}=2048$, \texttt{pooled\_text\_embed\_dim} $d_{\rm pool}=1280$, \texttt{image\_encoder\_depth} $n_{\rm conv}=2$, \texttt{image\_encoder\_width} $d_{\rm conv}=32$, 
    \texttt{attention\_dim} $d_{\rm attn}=256$, \texttt{cross\_attention\_heads} $n_{\rm attn}=4$, \texttt{number\_of\_transformer\_blocks} $N_b=4$, 
    \texttt{hidden\_dim} $d_{h}=256$, 
    \texttt{num\_mlp\_layers} $n_{\rm mlp}=2$.}
    \label{tab:arch_scheduler_dirichlet}
    \vskip 0.1in
    \centering
    \small
    \begin{tabular}{ccc}
        \toprule
        {\bf Layers} & {\bf Output size} & {\bf Note} \\
        \cmidrule(r){1-1} \cmidrule(r){2-2} \cmidrule(r){3-3}

        \multicolumn{3}{c}{\bf Inputs} \\
        \hline
        Input: $x$ & $B \times H \times W \times C$ & image-shaped noise \\
        Input: $e_{\text{text}}$ & $B \times L_{\text{text}} \times d_{\text{text}}$ & token text emb.; $d_{\text{text}}{=}2048$ \\
        Input: $\bar e_{\text{text}}$ & $B \times d_{\text{pool}}$ & optional pooled emb.; $d_{\text{pool}}{=}1280$ \\

        \hline
        \multicolumn{3}{c}{\bf Image encoder \& cross-attention blocks} \\
        \hline
        \tabincell{c}{Image encoder conv stack \\ (inside block $i$, layers $j=0,\dots,n_{\rm conv}{-}1$)} 
        & $B \times H_i \times W_i \times C_i$ 
        & \tabincell{c}{$n_{\rm conv}$ conv layers, $3{\times}3$, stride $1$ \\ 
                      GroupNorm, SiLU \\ 
                      $C_i = d_{\rm conv} \cdot 2^{\min(4,\,i+j)}$} \\[0.6ex]
        Flatten spatial dims 
        & $B \times (H_i W_i) \times C_i$ 
        & prepare for attention \\
        Query projection (Dense) 
        & $B \times (H_i W_i) \times C_i$ 
        & \tabincell{c}{applied to image features \\ (\texttt{query\_projection})} \\
        Key/Value projection (Dense) 
        & $B \times L_{\text{text}} \times C_i$ 
        & \tabincell{c}{applied to $e_{\text{text}}$ \\ (\texttt{key\_value\_projection})} \\
        MultiHeadDotProductAttention 
        & $B \times (H_i W_i) \times C_i$ 
        & \tabincell{c}{cross-attn: query from image, \\ key/value from text, $n_{\rm attn}$ heads} \\
        Residual + LayerNorm 
        & $B \times (H_i W_i) \times C_i$ 
        & \tabincell{c}{$\text{LayerNorm}(\text{query}+\text{attn\_output})$ \\ (\texttt{cross\_attn\_norm})} \\
        Reshape to image grid 
        & $B \times H_i \times W_i \times C_i$ 
        & output of \texttt{SchedulerTransformerBlock} \\[0.6ex]

        \tabincell{c}{Global avg. pooling \\ (per block)} 
        & $B \times C_i$ 
        & mean over $(H_i, W_i)$ for feature fusion \\
        \tabincell{c}{Downsample Conv $3{\times}3$, stride 2 \\ + GroupNorm + SiLU \\ (between blocks)} 
        & $B \times H_{i+1} \times W_{i+1} \times C_{i+1}$ 
        & \tabincell{c}{from reshaped image grid, for $i < N_b{-}1$; \\ $C_{i+1} = d_{\rm conv} \cdot 2^{\min(4,\,(i+1)\,n_{\rm conv}-1)}$} \\

        \hline
        \multicolumn{3}{c}{\bf Feature pyramid fusion} \\
        \hline
        Concat. over blocks 
        & $B \times \sum_i C_i$ 
        & feature pyramid $\{\text{block }i\}_{i=0}^{N_b-1}$ \\
        Concat. with $\bar e_{\text{text}}$ 
        & $B \times \big(\sum_i C_i + d_{\text{pool}}\big)$ 
        & only if pooled emb. used \\

        \hline
        \multicolumn{3}{c}{\bf MLP Dirichlet head} \\
        \hline
        Dense, SiLU $\times$ $(n_{\rm mlp}-1)$ 
        & $B \times d_h$ 
        & default $d_h = 256$ \\
        Dense (Dirichlet logits head) 
        & $B \times (L{+}1)$ 
        & $L = \texttt{num\_timesteps}$ \\
        Softplus $+ 10^{-3}$ 
        & $B \times (L{+}1)$ 
        & $\boldsymbol{\alpha} = \text{softplus}(\cdot){+}10^{-3}$ \\
        \bottomrule
    \end{tabular}
\end{table*}

\section{Experiment Settings}
\subsection{Synthetic Sanity Check for Baseline Variance}
\label{subsec:synthetic-sanity}

RLOO is an unbiased baseline but fragile in exactly the regime most relevant to our policy learning scenarios: limited number of rollouts per context and highly heterogeneous reward scales.
To isolate this failure mode and evaluate variance reduction in a controlled setting, we construct a synthetic experiment where the policy is analytic, rewards follow a spiky random–effects model. We measure the variance of the REINFORCE gradient estimators since they are all unbiased.
This setup cleanly reveals the effect of the proposed James--Stein (JS) baseline.

\paragraph{Setup.}
We adopt an analytic Dirichlet policy
$\pi(\tau)=\mathrm{Dirichlet}(\boldsymbol{\alpha})$
with concentration $\boldsymbol{\alpha}=2\mathbf{1}_T$ and horizons
$T\in\{4,16,64\}$.
Its score function has a closed form:
\[
\nabla_{\alpha_j}\log\pi(\tau)
= \psi\!\Big(\sum_t \alpha_t\Big)-\psi(\alpha_j)+\log\tau_j,
\]
where $\psi(\cdot)$ denotes the digamma function and $\log\tau_j$ is the elementwise logarithm of the $j$-th coordinate of $\tau$.
To mimic the two-level random–effects model used in our theoretical analysis,
we generate synthetic rewards as
\begin{equation}
\begin{aligned}
&r^{(c,i)} = \mu_c + \varepsilon^{(c,i)} + \delta^{(c,i)},\\
&\mu_c \sim \mathcal{N}(0,1), \quad
\varepsilon^{(c,i)} \sim \mathcal{N}\big(0, s_c^2\big), \\
&s_c \sim \mathrm{LogNormal}(0,1), \quad
\delta^{(c,i)} \sim \mathrm{Bernoulli}(0.15)\times 8,
\end{aligned}
\label{eq:spiky-model}
\end{equation}
where $\mu_c$ acts as the context-level effect (analogous to $\mu_0+\xi^{(c)}$ in Eq.~\eqref{eq:random-effects})
and $\varepsilon^{(c,i)}$ represents within-context noise.
The additional sparse term $\delta^{(c,i)}$ introduces occasional outliers,
creating a heavy-tailed, heteroskedastic reward landscape that stresses RLOO in the small-$K$ regime.

We sweep through different settings for $(B, K, L)$. For each $(B,K,L)$ we generate $500$ i.i.d.\ batches, compute
\[
\hat g(b)
= \frac{1}{BK}\sum_{c=1}^B \sum_{i=1}^K
\big(r^{(c,i)} - b^{(c,i)}\big)\,
\nabla_{\alpha}\log \pi(\tau^{(c,i)}),
\]
once with RLOO and once with JS, and report the empirical per-dimension variance
$\frac{1}{L}\sum_{j=1}^L \Var(\hat g_j)$.
We sweep $B\in\{8,16,32\}$ and $K\in\{2,4,8,16\}$.

\begin{table}[t]
    \centering
    \footnotesize
    \caption{Synthetic variance sanity check.
    Average per-dimension variance of REINFORCE for an analytic Dirichlet policy.
    Rewards follow Eq.~\eqref{eq:spiky-model}.
    Entries are RLOO $\to$ JS.
    JS consistently gives gain over RLOO baselines for various settings.}
    \label{tab:synthetic-variance}
    \begin{tabular}{llccc}
    \toprule
    $K$ & $B$ & $L{=}4$ & $L{=}16$ & $L{=}64$ \\
    \midrule
    \multirow{3}{*}{$2$}
      & $8$  & 1.055 $\to$ \textbf{0.612} & 1.445 $\to$ \textbf{0.725} & 1.148 $\to$ \textbf{0.625} \\
      & $16$ & 0.479 $\to$ \textbf{0.285} & 0.663 $\to$ \textbf{0.339} & 0.606 $\to$ \textbf{0.339} \\
      & $32$ & 0.261 $\to$ \textbf{0.140} & 0.328 $\to$ \textbf{0.174} & 0.310 $\to$ \textbf{0.165} \\
    \midrule
    \multirow{3}{*}{$4$}
      & $8$  & 0.393 $\to$ \textbf{0.284} & 0.402 $\to$ \textbf{0.307} & 0.386 $\to$ \textbf{0.299} \\
      & $16$ & 0.175 $\to$ \textbf{0.143} & 0.193 $\to$ \textbf{0.154} & 0.210 $\to$ \textbf{0.163} \\
      & $32$ & 0.086 $\to$ \textbf{0.066} & 0.099 $\to$ \textbf{0.079} & 0.106 $\to$ \textbf{0.084} \\
    \midrule
    \multirow{3}{*}{$8$}
      & $8$  & 0.140 $\to$ \textbf{0.126} & 0.191 $\to$ \textbf{0.170} & 0.182 $\to$ \textbf{0.166} \\
      & $16$ & 0.071 $\to$ \textbf{0.064} & 0.099 $\to$ \textbf{0.090} & 0.086 $\to$ \textbf{0.078} \\
      & $32$ & 0.038 $\to$ \textbf{0.034} & 0.042 $\to$ \textbf{0.038} & 0.045 $\to$ \textbf{0.040} \\
    \midrule
    \multirow{3}{*}{$16$}
      & $8$  & 0.072 $\to$ \textbf{0.070} & 0.081 $\to$ \textbf{0.078} & 0.094 $\to$ \textbf{0.089} \\
      & $16$ & 0.031 $\to$ \textbf{0.030} & 0.038 $\to$ \textbf{0.037} & 0.039 $\to$ \textbf{0.038} \\
      & $32$ & 0.017 $\to$ \textbf{0.016} & 0.020 $\to$ \textbf{0.019} & 0.020 $\to$ \textbf{0.019} \\
    \bottomrule
    \end{tabular}
\end{table}

\paragraph{Observations.}
As shown in \cref{tab:synthetic-variance}, the James--Stein (JS) baseline delivers uniformly lower gradient variance than RLOO across all batch sizes, rollout counts, and horizons.
For the most challenging regime ($K\!=\!2$), JS reduces variance by roughly $45$--$50\%$, effectively matching the stability one would expect from doubling the rollout count—without any additional computation.
Even with moderate rollouts ($K\!=\!4$), variance drops by $20$--$25\%$, demonstrating that the improvement is systematic rather than case-specific.
Across all settings, JS remains consistently superior, reflecting its adaptive shrinkage across both within- and cross-context variability.
In practical terms, this means JS acts as a computationally free variance amplifier: it increases the \emph{effective} number of rollouts available to policy-gradient estimation, providing a principled and efficient replacement for RLOO in rollout-expensive training regimes. We visualize a side-by-side comparison in \cref{fig:js_variance} for the $B=32,L=16$ case with different number of rollouts $K$.

\subsection{General Text-to-Image Experiments on HPDv2}
\label{subsec:general-t2i}
\paragraph{General settings.} For all the experiments, we consistently use the AdamW \cite{loshchilov2017decoupled} optimizer for learning scheduler parameters. We set the \texttt{weight\_decay} to $1e-4$ throughout experiments, tune the learning rates (see \cref{tab:gt2i_lr}), and use default values for the remaining hyperparameters. We apply the gradient clipping with a maximum norm of $1$. For all models we consistently use a rollout number of 2, and set the output resolution to $1024\times1024$. Due to memory limit, we use a batch size of 32 for smaller models like SD-XL or SD-3.5M, and a batch size of 16 for larger ones like SD-3.5L and Flux-Dev. We use a guidance weight of $7.5$ for SD-XL and $5.0$ for SD3.5-M/L and Flux models. For SD-XL model we use the DDIM solver \cite{song2020denoising} for per step denoising update; for the flow-matching-based SD3.5 and Flux models we use the Euler solver.

\begin{table}[t]
  \centering
  \footnotesize
  \caption{Learning rates for schedule training on HPDv2 prompts.}
  \label{tab:gt2i_lr}

  \setlength{\tabcolsep}{6pt}
  \renewcommand{\arraystretch}{1.1}

  \begin{tabular}{lccccc}
    \toprule
    \textbf{Backbone} & \textbf{$L{=}5$} & \textbf{$L{=}10$} & \textbf{$L{=}20$} & \textbf{$L{=}40$} & \textbf{$L{=}80$} \\
    \midrule
    SD\mbox{-}XL      & $5\mathrm{e}{-5}$ & $5\mathrm{e}{-5}$ & $5\mathrm{e}{-5}$ & $5\mathrm{e}{-5}$ & $5\mathrm{e}{-5}$ \\
    SD3.5\mbox{-}M    & $5\mathrm{e}{-5}$ & $2.5\mathrm{e}{-5}$ & $2.5\mathrm{e}{-5}$ & $1.25\mathrm{e}{-5}$ & $5\mathrm{e}{-5}$ \\
    SD3.5\mbox{-}L    & $5\mathrm{e}{-5}$ & $5\mathrm{e}{-5}$ & $5\mathrm{e}{-5}$ & $5\mathrm{e}{-5}$ & $5\mathrm{e}{-5}$ \\
    Flux\mbox{-}Dev   & $2.5\mathrm{e}{-5}$ & $5\mathrm{e}{-5}$ & $2.5\mathrm{e}{-5}$ & $5\mathrm{e}{-5}$ & $5\mathrm{e}{-5}$ \\
    \bottomrule
  \end{tabular}
\end{table}

\paragraph{TPDM baseline.} We reimplement the TPDM baseline using (i) the same network architecture we used, and (ii) the same set of seeds, hyperparmeters and base model settings for training. We follow the customized PPO implementation in TPDM with no ratio clipping. 

We would like to explicitly mention that (i) we \emph{exclude} any regularizer including KL constraint between a reference schedule and the learned schedule during training to be consistent with our settings. This is different from the vanilla implementation of TPDM, where KL constraint is used. (ii) TPDM in its basic form \emph{does not} involve text condition as its input. We experiment with different settings: (i) we faithfully nullify the textual inputs by setting the embeddings to zero vectors to train the model, and (ii) we include the textual inputs as in our settings. We report in \cref{tab:hpdt2i_steps} the results for (i). For (ii) the variant, especially in our setting, reduces to the RLOO baseline, which we also report in \cref{tab:hpdt2i_steps}.

\subsection{Text Rendering and Fine-Grained Alignment}
We use the same settings as in \cref{subsec:general-t2i}, except for tuning the learning rates (see \cref{tab:fg_t2i_lr}). For the OCR metrics, we adopt the commonly used mask textspotter v3 model \cite{liao2020mask} to calculate the character level accuracy, precision and recall scores for each image, and then calculate the average across the testing samples as the final results in \cref{tab:text-rendering-concise}.

\begin{table}[t]
  \centering
  \footnotesize
  \caption{Learning rates for schedule training on text rendering and geneval-style prompts. Num. of sampling steps is 40.}
  \label{tab:fg_t2i_lr}

  \setlength{\tabcolsep}{6pt}
  \renewcommand{\arraystretch}{1.1}

  \begin{tabular}{lcc}
    \toprule
    \textbf{Backbone} & \textbf{SD3.5-M} & \textbf{Flux-Dev} \\
    \midrule
    Text Render      & $1.25\mathrm{e}{-5}$ & $5\mathrm{e}{-5}$ \\
    Geneval    & 
    $1.25\mathrm{e}{-6}$ & $5\mathrm{e}{-5}$ \\
    \bottomrule
  \end{tabular}
\end{table}

\section{Additional qualitative results}
\begin{figure*}[t]
\centering

\colhead{A white-haired girl in a pink sweater looks out a window in her bedroom.}\hfill
\colhead{A hamster resembling a horse.}\hfill
\colhead{A lemon wearing sunglasses on the beach.}

\begin{subfigure}[t]{0.315\linewidth}
  \centering
  \includegraphics[width=\linewidth]{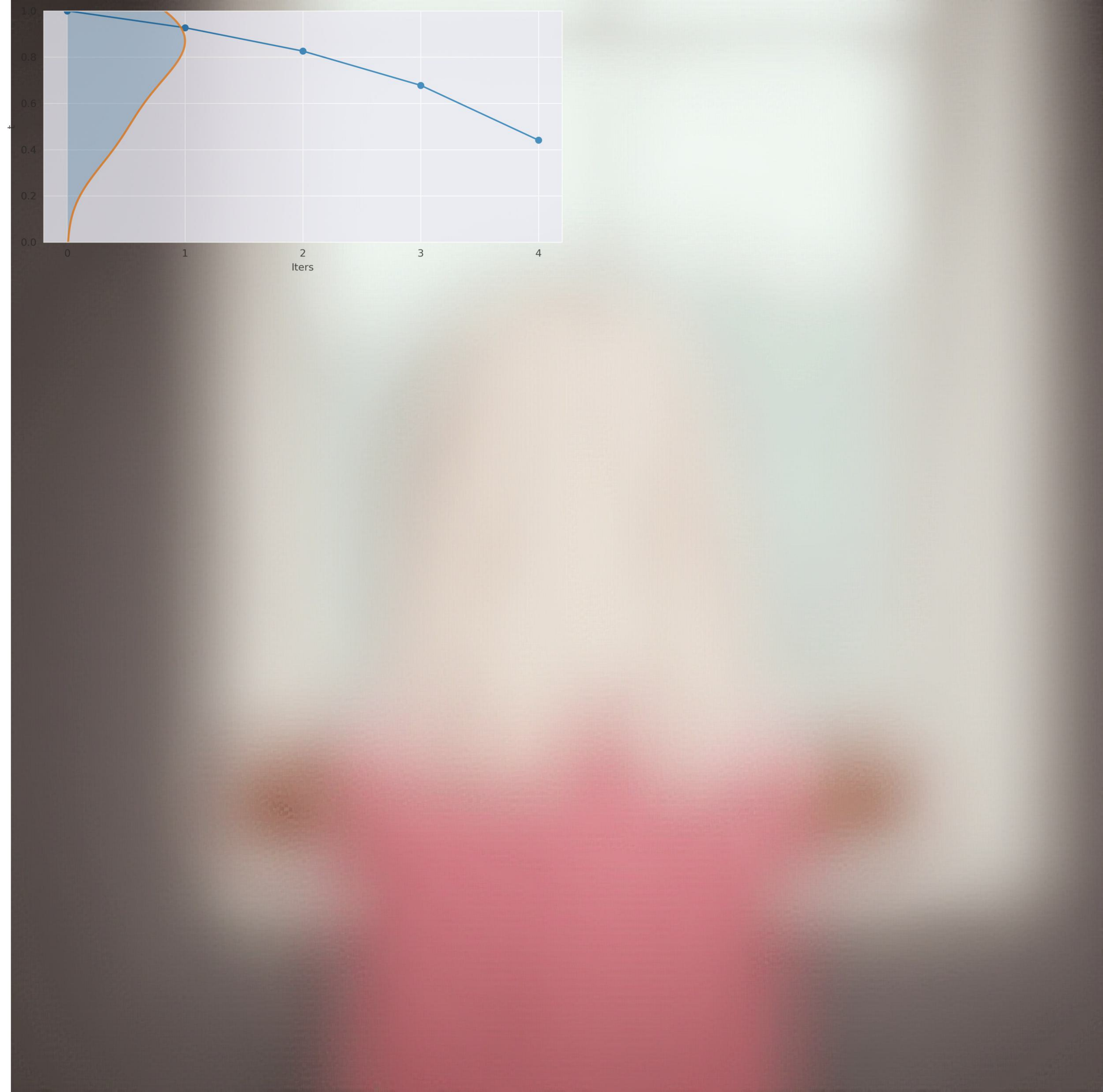}
  \label{fig:appx_fs_r1c1}
\end{subfigure}\hfill
\begin{subfigure}[t]{0.315\linewidth}
  \centering
  \includegraphics[width=\linewidth]{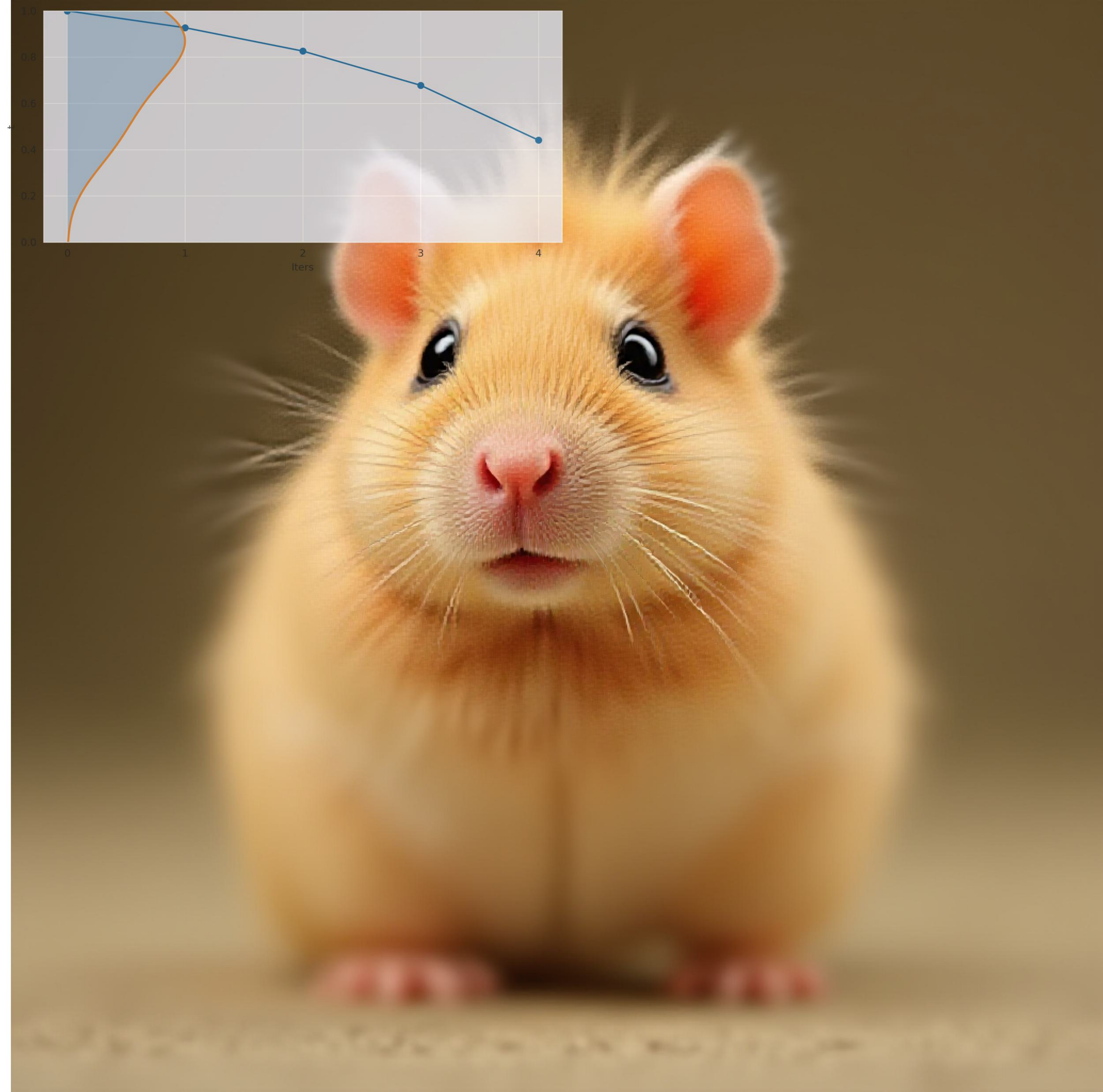}
  \label{fig:appx_fs_r1c2}
\end{subfigure}\hfill
\begin{subfigure}[t]{0.315\linewidth}
  \centering
  \includegraphics[width=\linewidth]{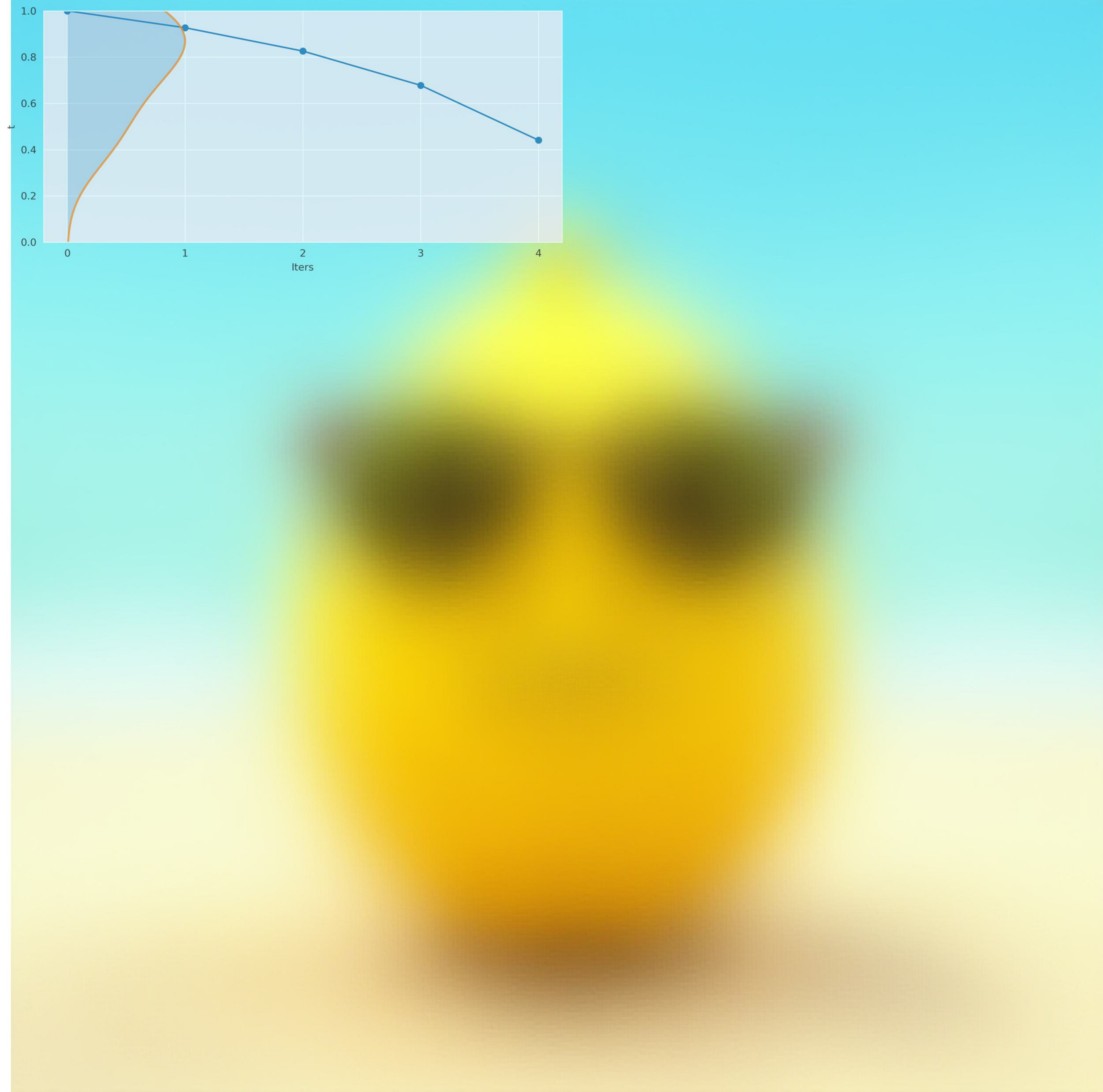}
  \label{fig:appx_fs_r1c3}
\end{subfigure}

\vspace{-10pt} 

\begin{subfigure}[t]{0.315\linewidth}
  \centering
  \includegraphics[width=\linewidth]{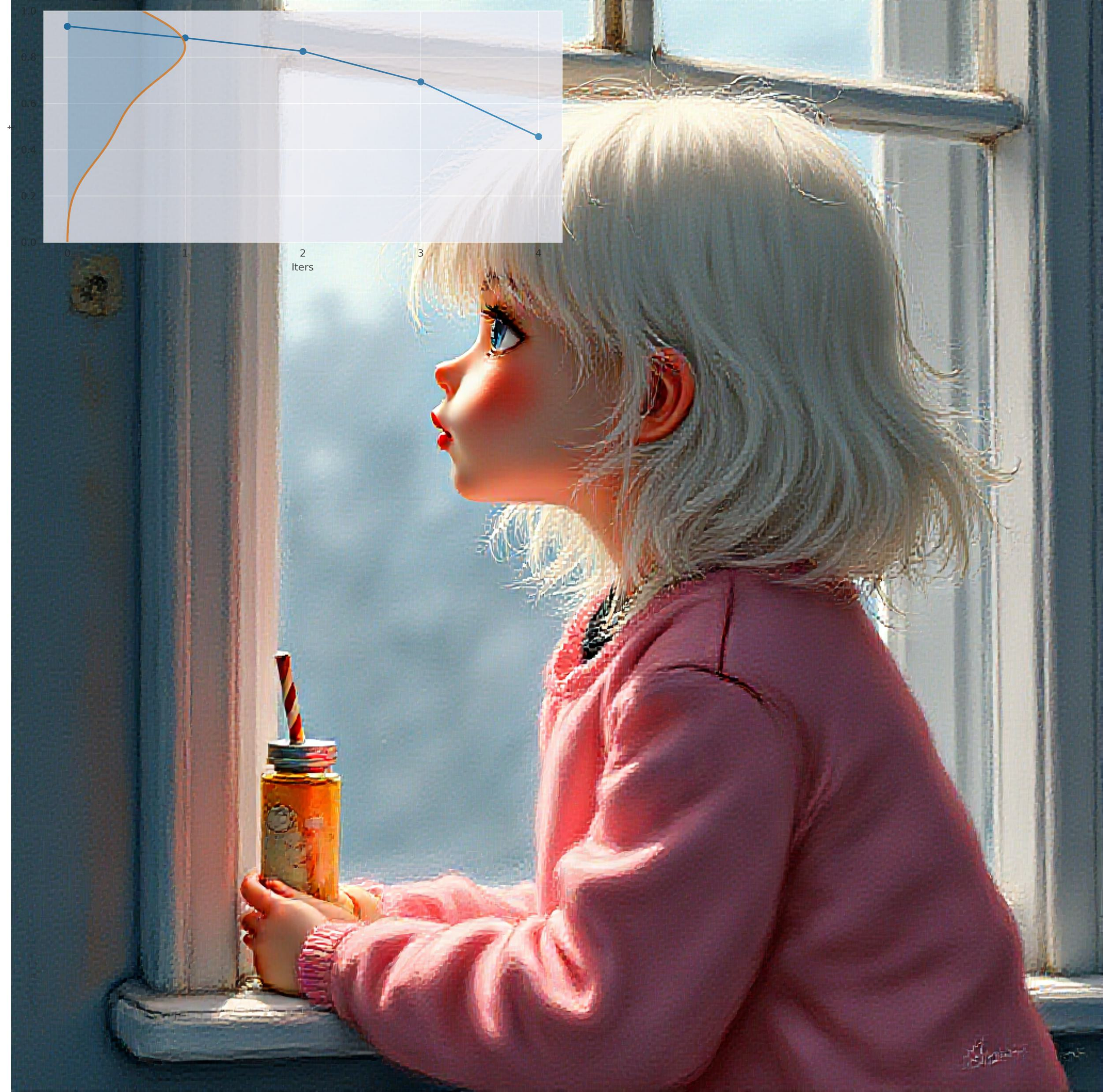}
  \label{fig:appx_fs_r2c1}
\end{subfigure}\hfill
\begin{subfigure}[t]{0.315\linewidth}
  \centering
  \includegraphics[width=\linewidth]{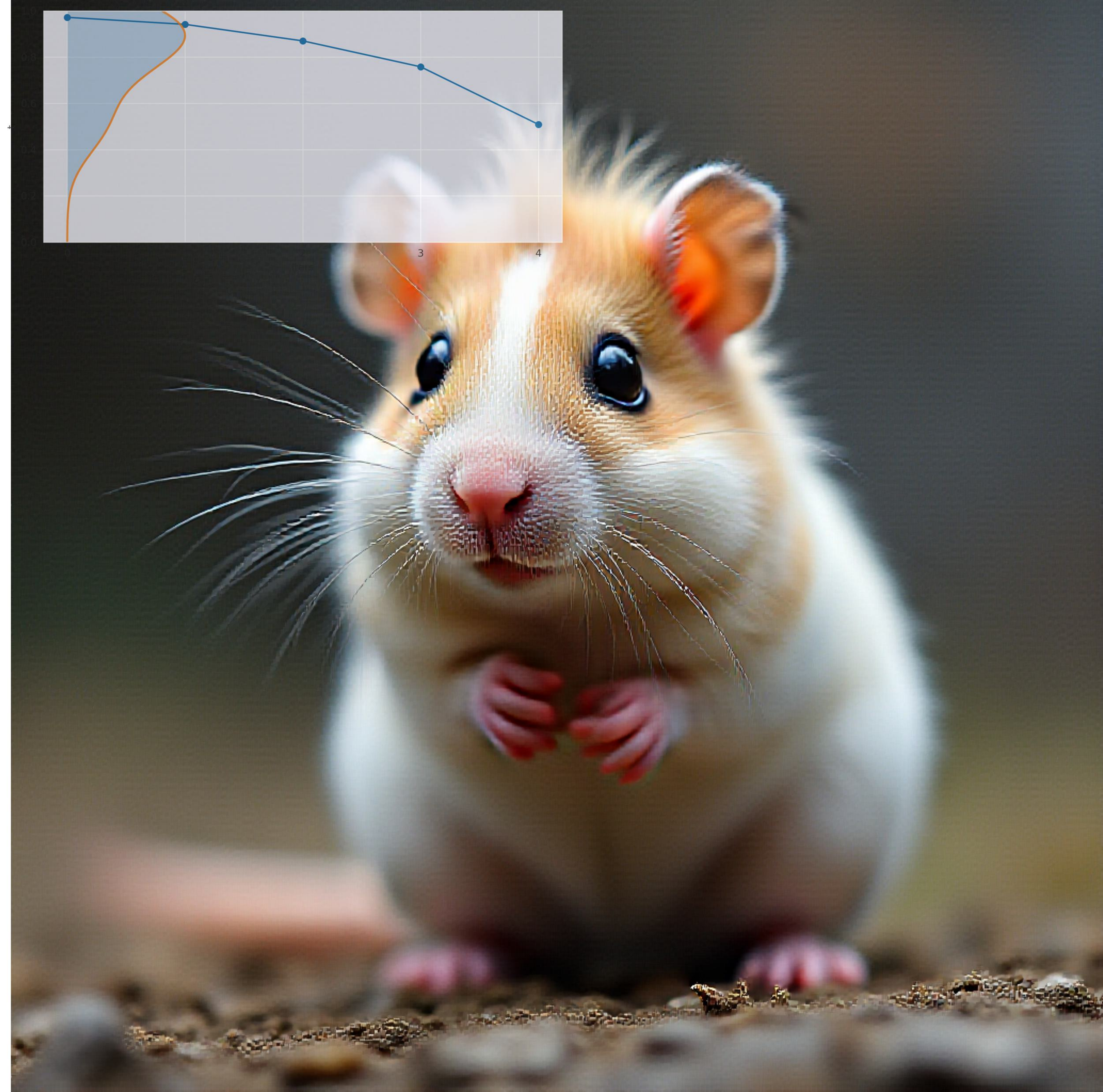}
  \label{fig:appx_fs_r2c2}
\end{subfigure}\hfill
\begin{subfigure}[t]{0.315\linewidth}
  \centering
  \includegraphics[width=\linewidth]{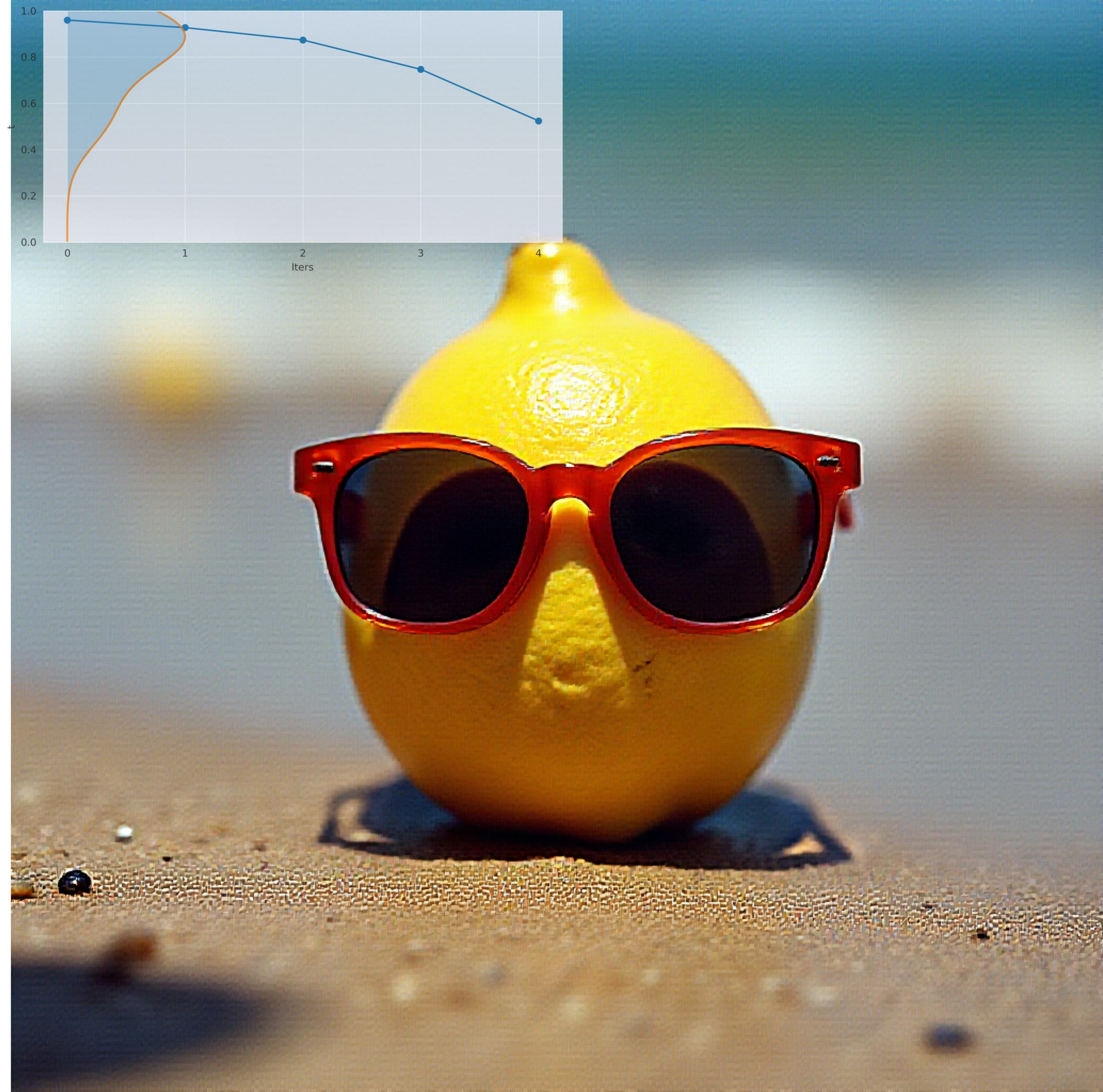}
  \label{fig:appx_fs_r2c3}
\end{subfigure}
\vspace{-10pt}
\caption{\textbf{Rescheduling improves few-step sampling}. Comparisons between images generated with default schedules (upper) and our learned schedules (lower) from Flux-Dev, with 5 steps.}
\vspace{-15pt}
\label{fig:appx_few_steps_1}
\end{figure*}

\begin{figure*}[t]
\centering

\colhead{A goth anime woman with a symmetrical and attractive face in a black and white watercolor headshot art on ArtStation.}\hfill
\colhead{A girl in school uniform standing in the city.}\hfill
\colhead{A photo of Big Chungus from Looney Tunes.}

\begin{subfigure}[t]{0.315\linewidth}
  \centering
  \includegraphics[width=\linewidth]{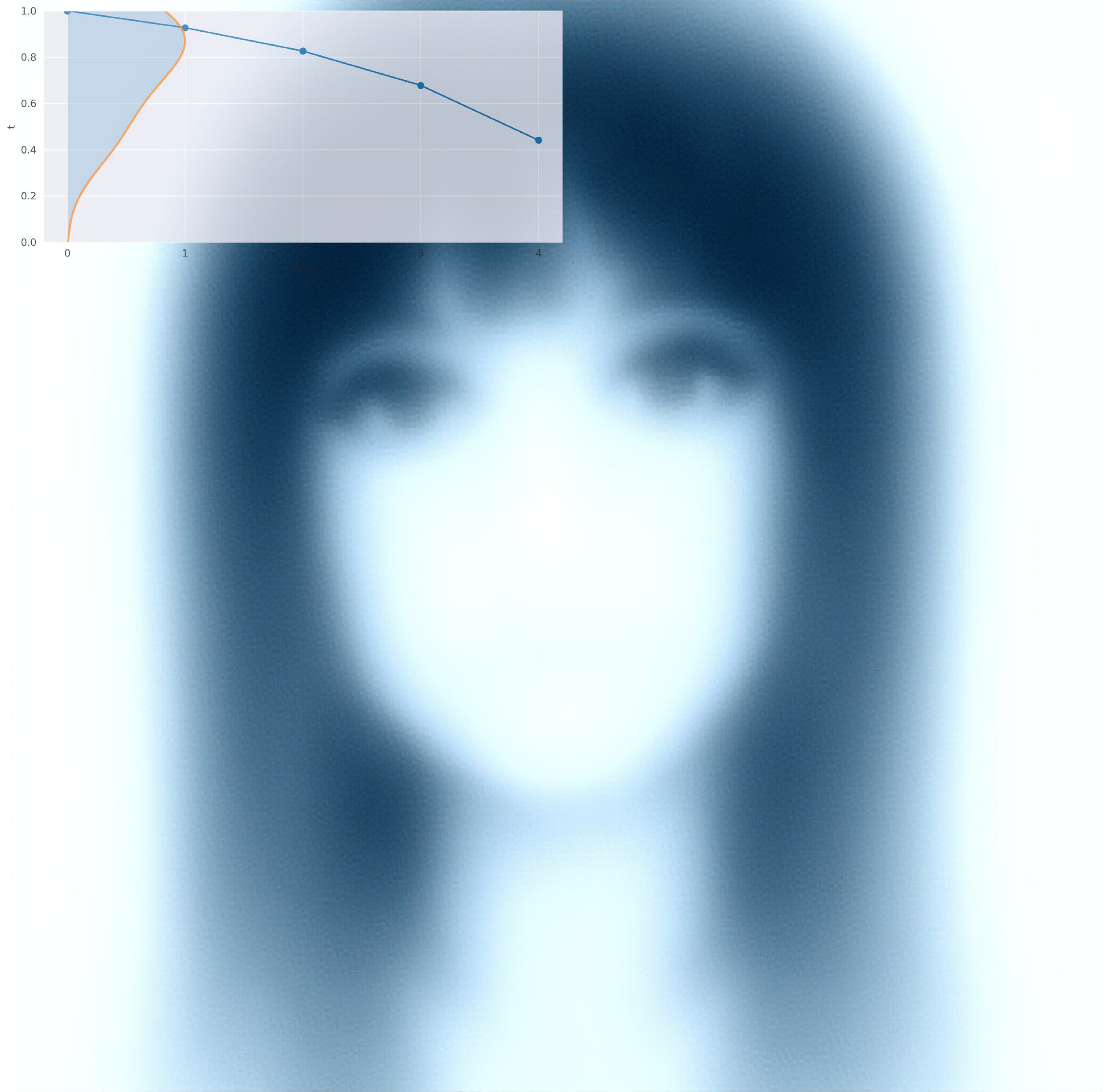}
  \label{fig:appx_fs2_r1c1}
\end{subfigure}\hfill
\begin{subfigure}[t]{0.315\linewidth}
  \centering
  \includegraphics[width=\linewidth]{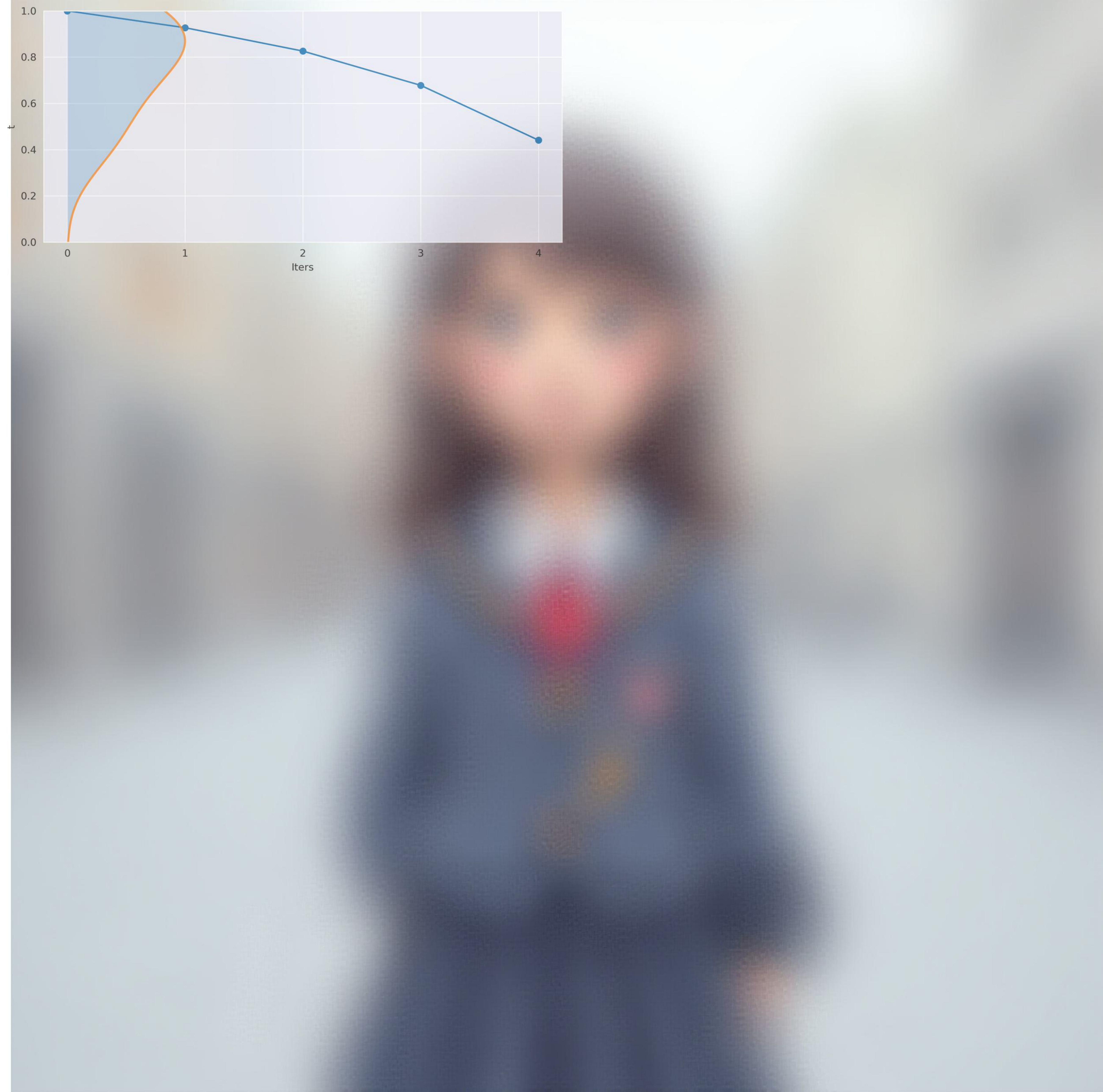}
  \label{fig:appx_fs2_r1c2}
\end{subfigure}\hfill
\begin{subfigure}[t]{0.315\linewidth}
  \centering
  \includegraphics[width=\linewidth]{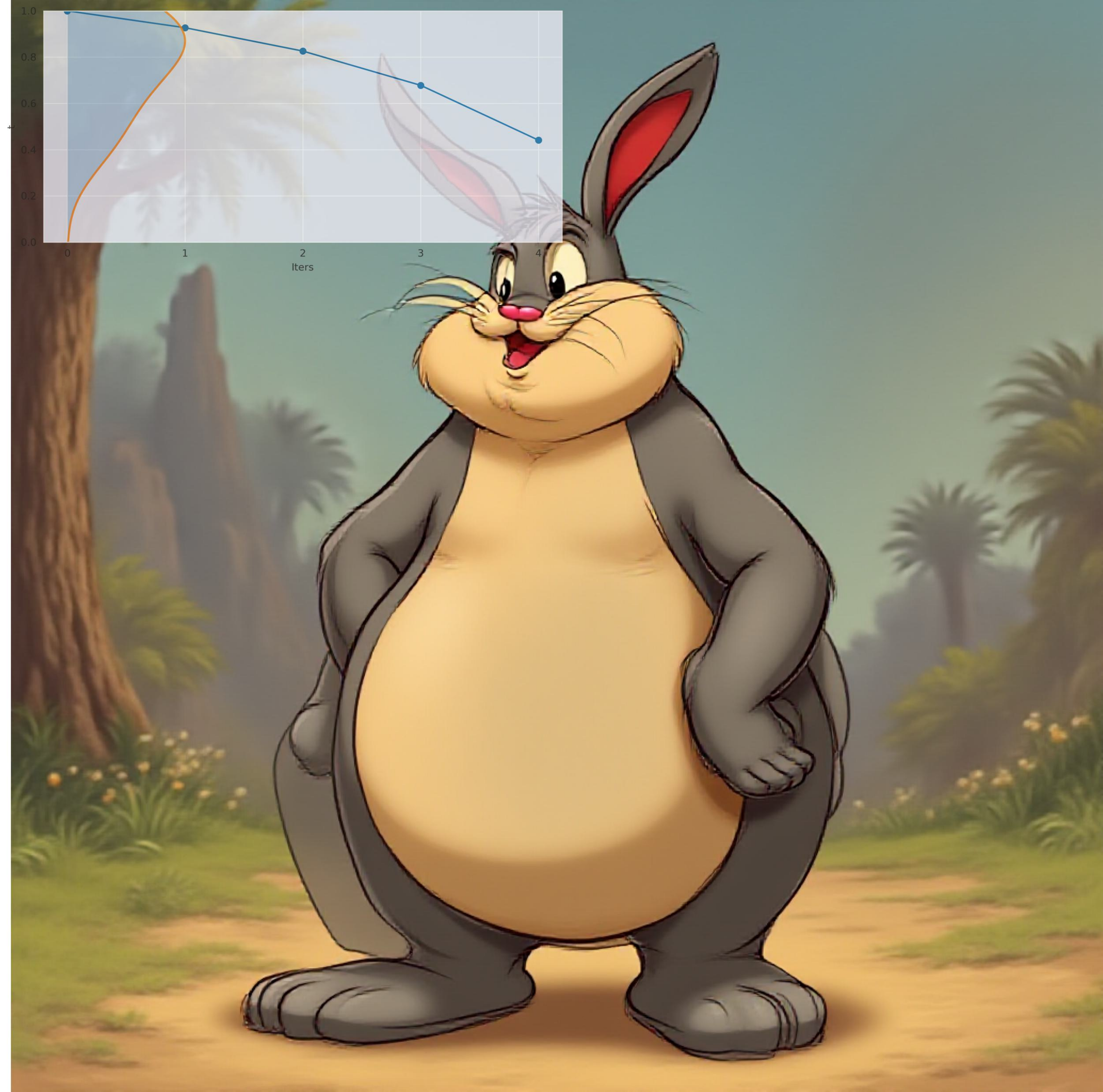}
  \label{fig:appx_fs2_r1c3}
\end{subfigure}

\vspace{-10pt} 

\begin{subfigure}[t]{0.315\linewidth}
  \centering
  \includegraphics[width=\linewidth]{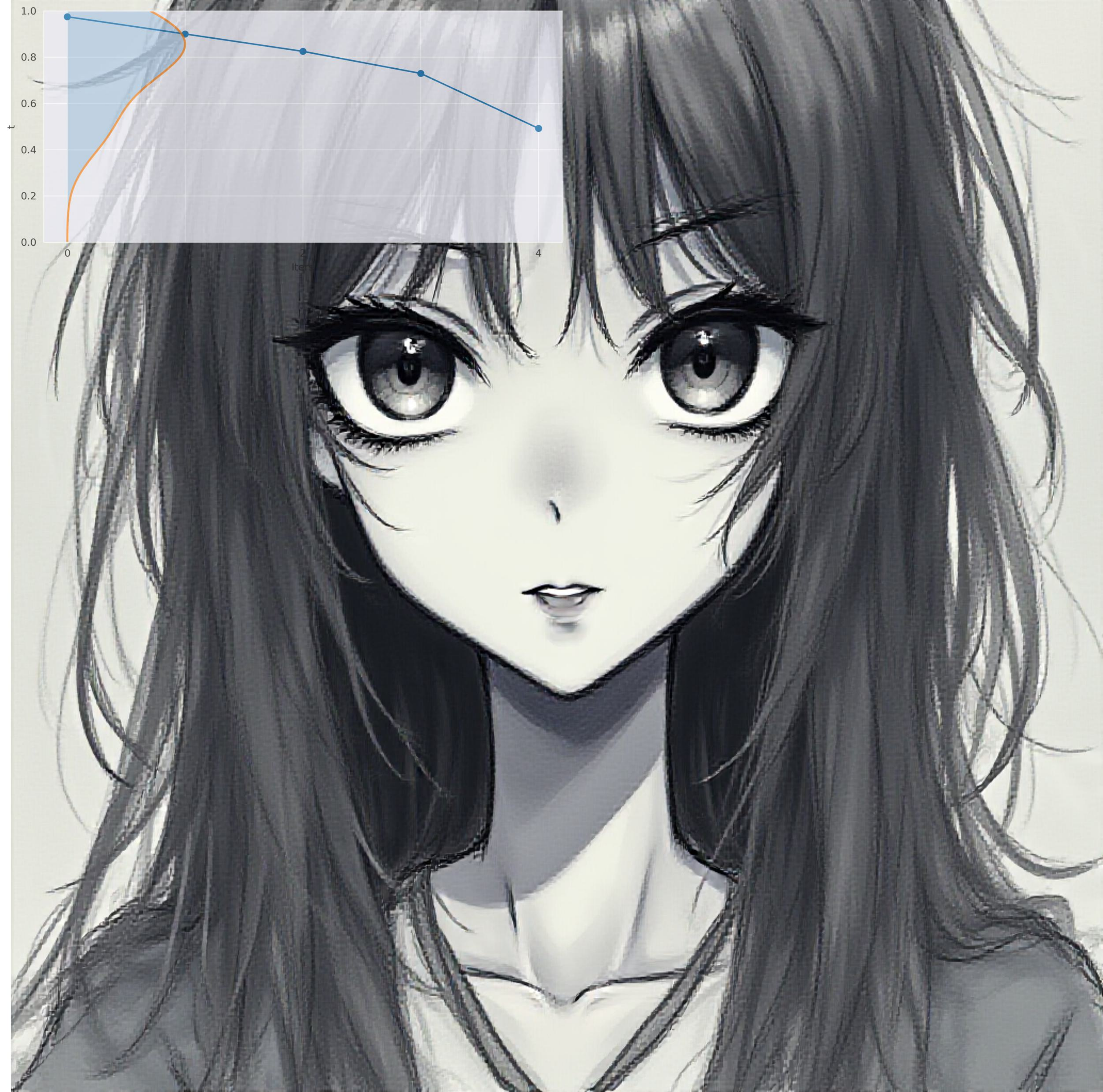}
  \label{fig:appx_fs2_r2c1}
\end{subfigure}\hfill
\begin{subfigure}[t]{0.315\linewidth}
  \centering
  \includegraphics[width=\linewidth]{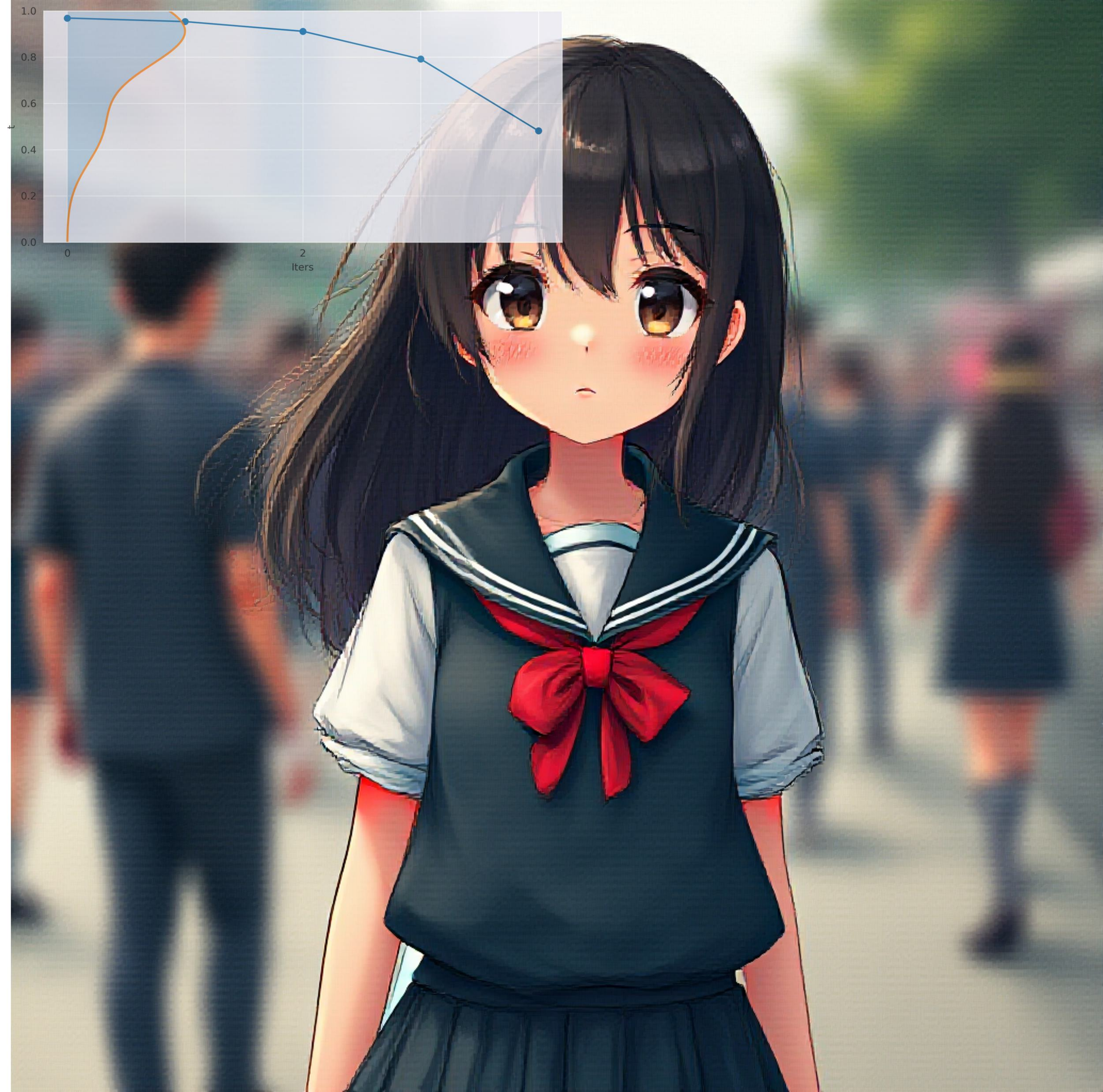}
  \label{fig:appx_fs2_r2c2}
\end{subfigure}\hfill
\begin{subfigure}[t]{0.315\linewidth}
  \centering
  \includegraphics[width=\linewidth]{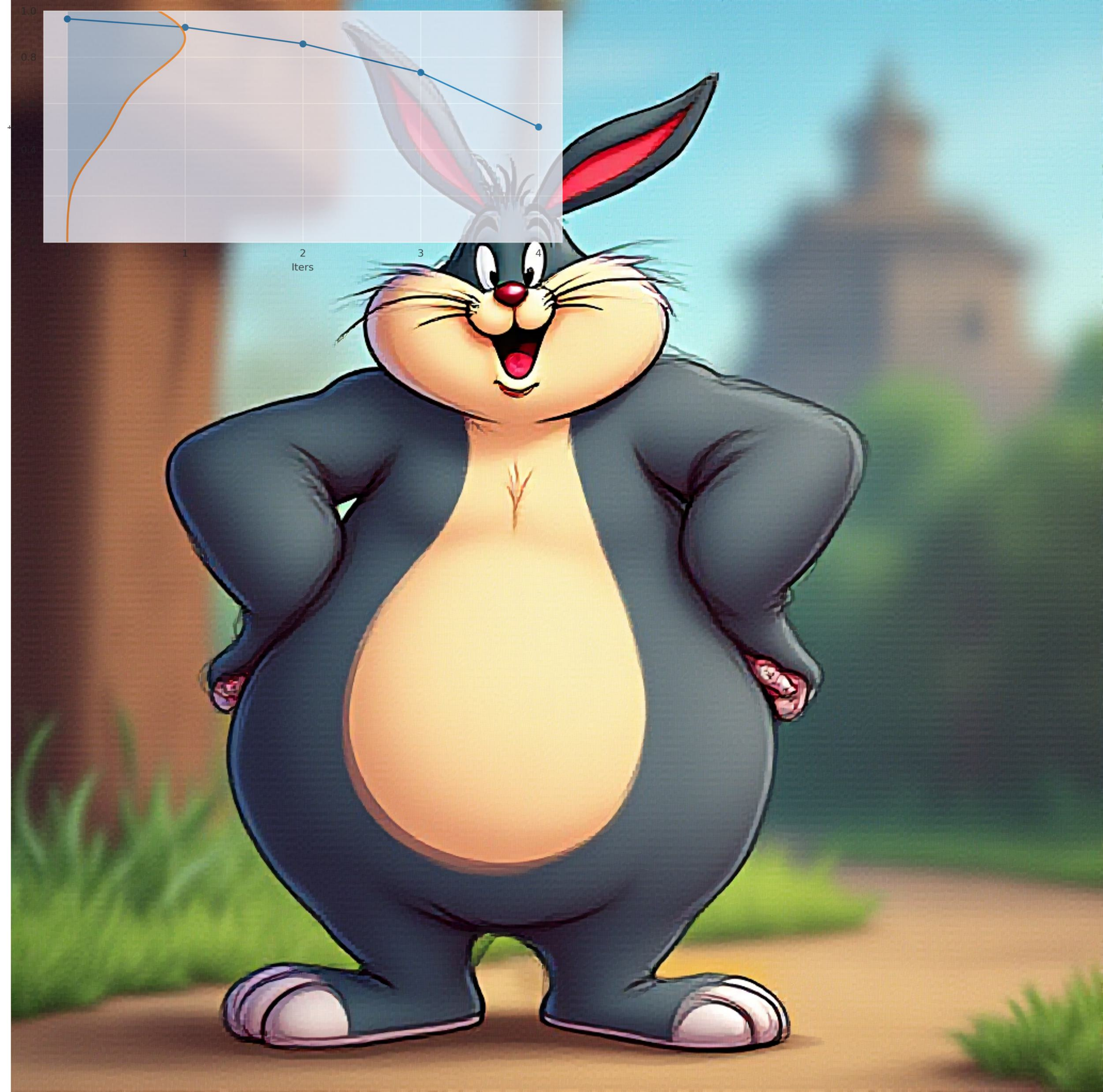}
  \label{fig:appx_fs2_r2c3}
\end{subfigure}
\vspace{-10pt}
\caption{\textbf{Rescheduling improves few-step sampling}. Comparisons between images generated with default schedules (upper) and our learned schedules (lower) from Flux-Dev, with 5 steps.}
\vspace{-15pt}
\label{fig:appx_few_steps_2}
\end{figure*}

\begin{figure*}[t]
\centering

\colhead{A {\color{red} banana spaceship} reminiscent of Homeworld}\hfill
\colhead{An anime girl with an athletic build poses confidently while holding an assault rifle...}\hfill
\colhead{{\color{red} Clint Eastwood} fighting in a {\color{red} white Michelin man costume} with hippo...}

\begin{subfigure}[t]{0.315\linewidth}
  \centering
  \includegraphics[width=\linewidth]{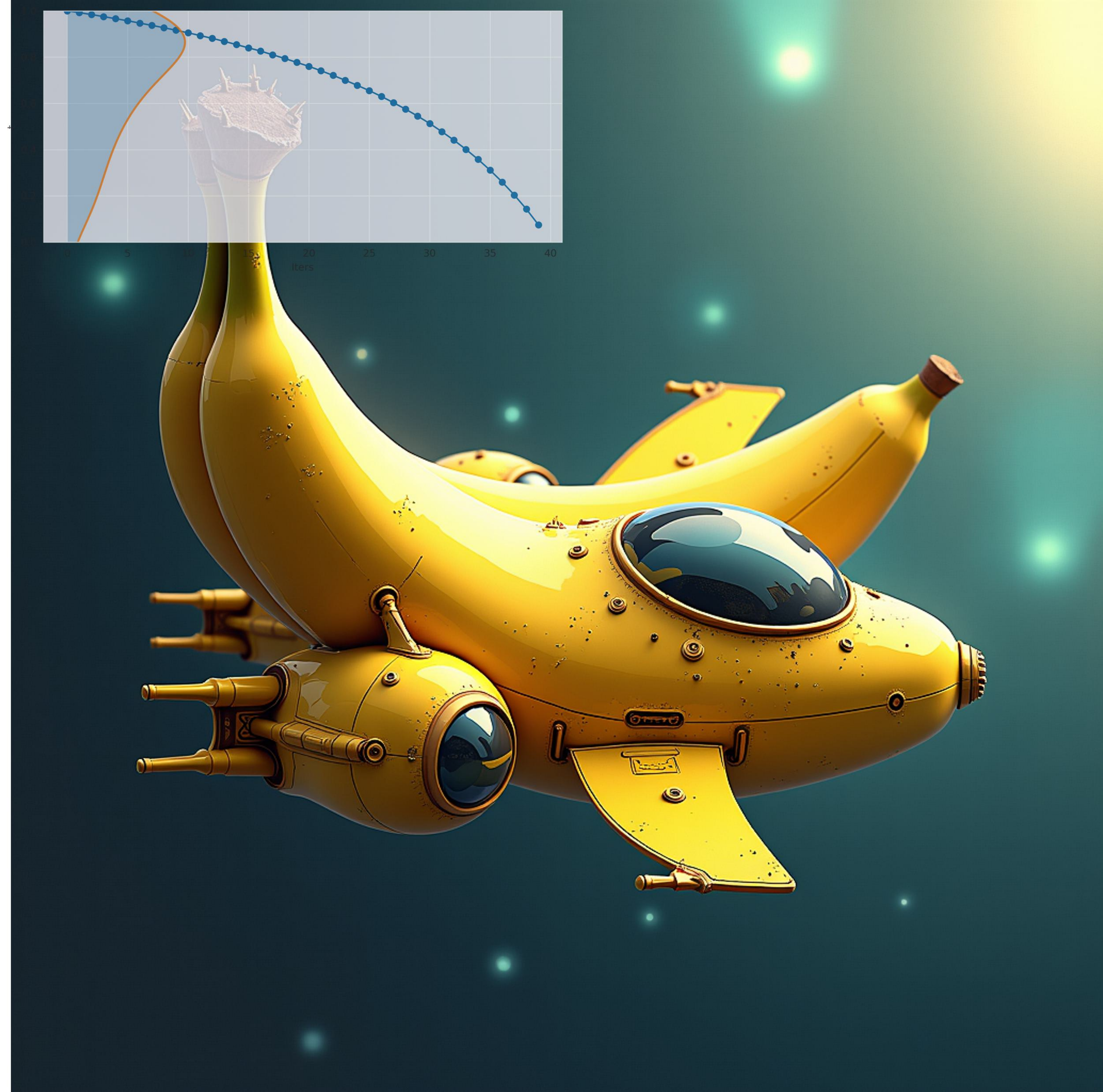}
  \label{fig:appx_gt_r1c1}
\end{subfigure}\hfill
\begin{subfigure}[t]{0.315\linewidth}
  \centering
  \includegraphics[width=\linewidth]{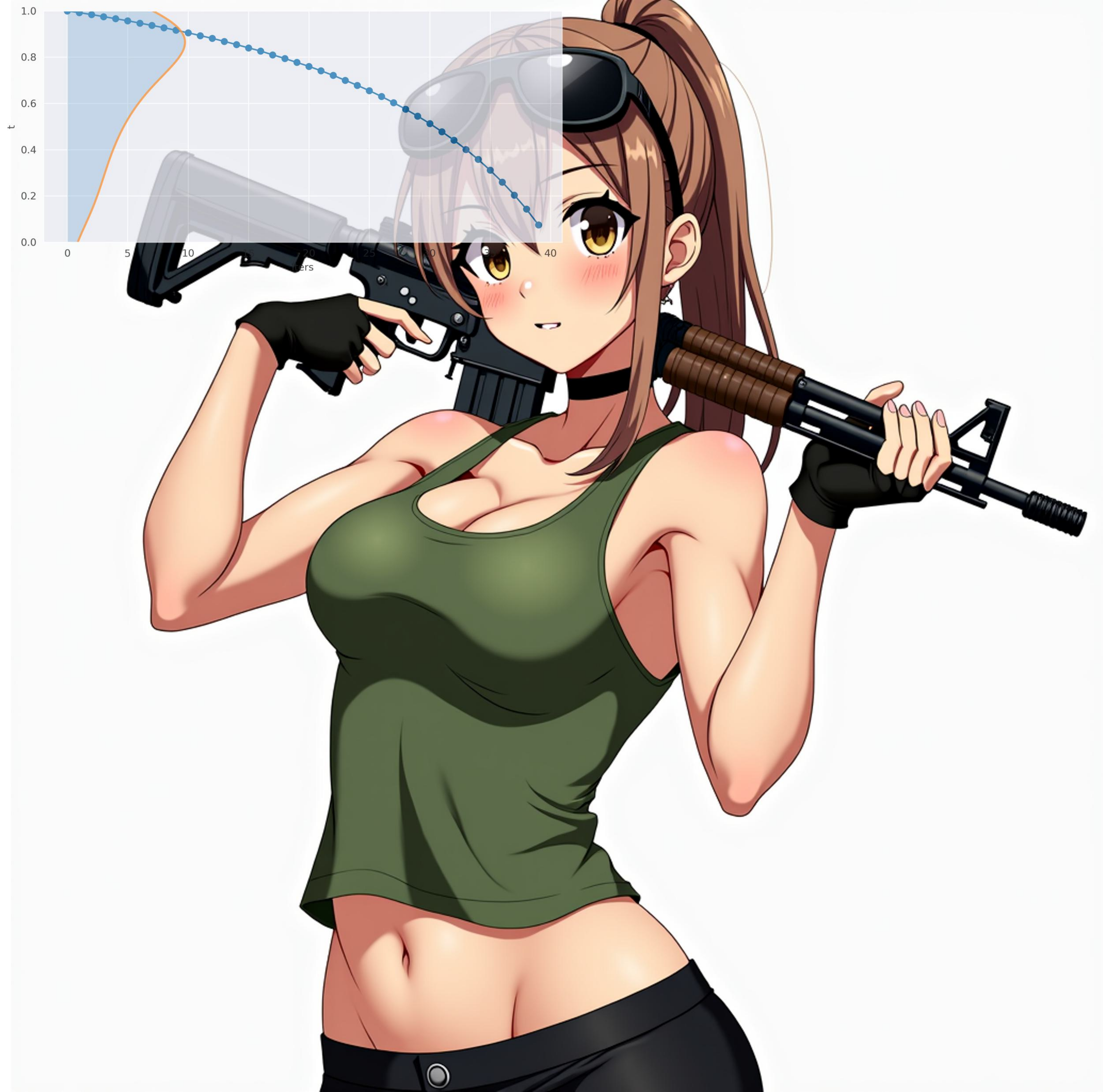}
  \label{fig:appx_gt_r1c2}
\end{subfigure}\hfill
\begin{subfigure}[t]{0.315\linewidth}
  \centering
  \includegraphics[width=\linewidth]{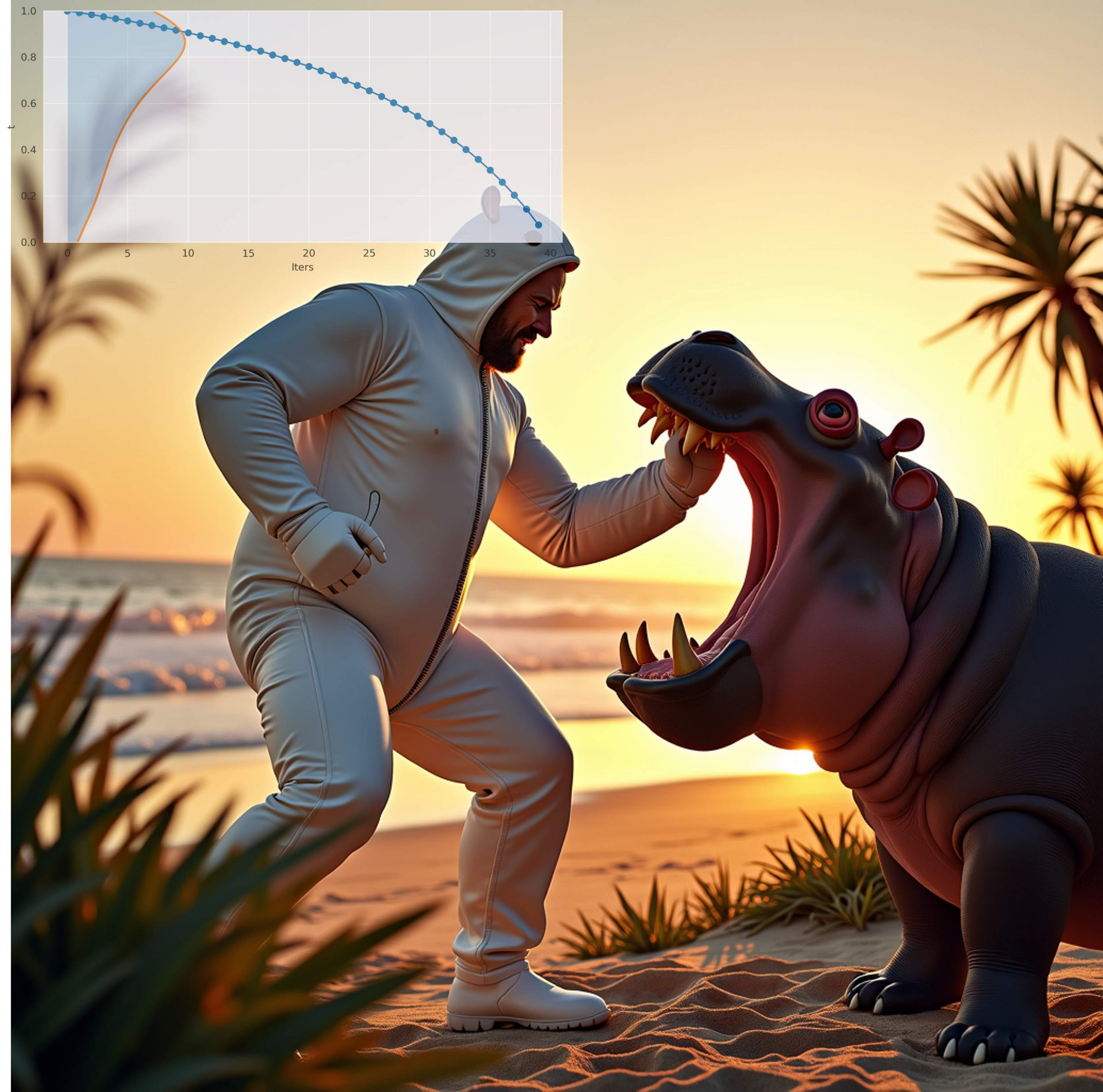}
  \label{fig:appx_gt_r1c3}
\end{subfigure}

\vspace{-10pt} 

\begin{subfigure}[t]{0.315\linewidth}
  \centering
  \includegraphics[width=\linewidth]{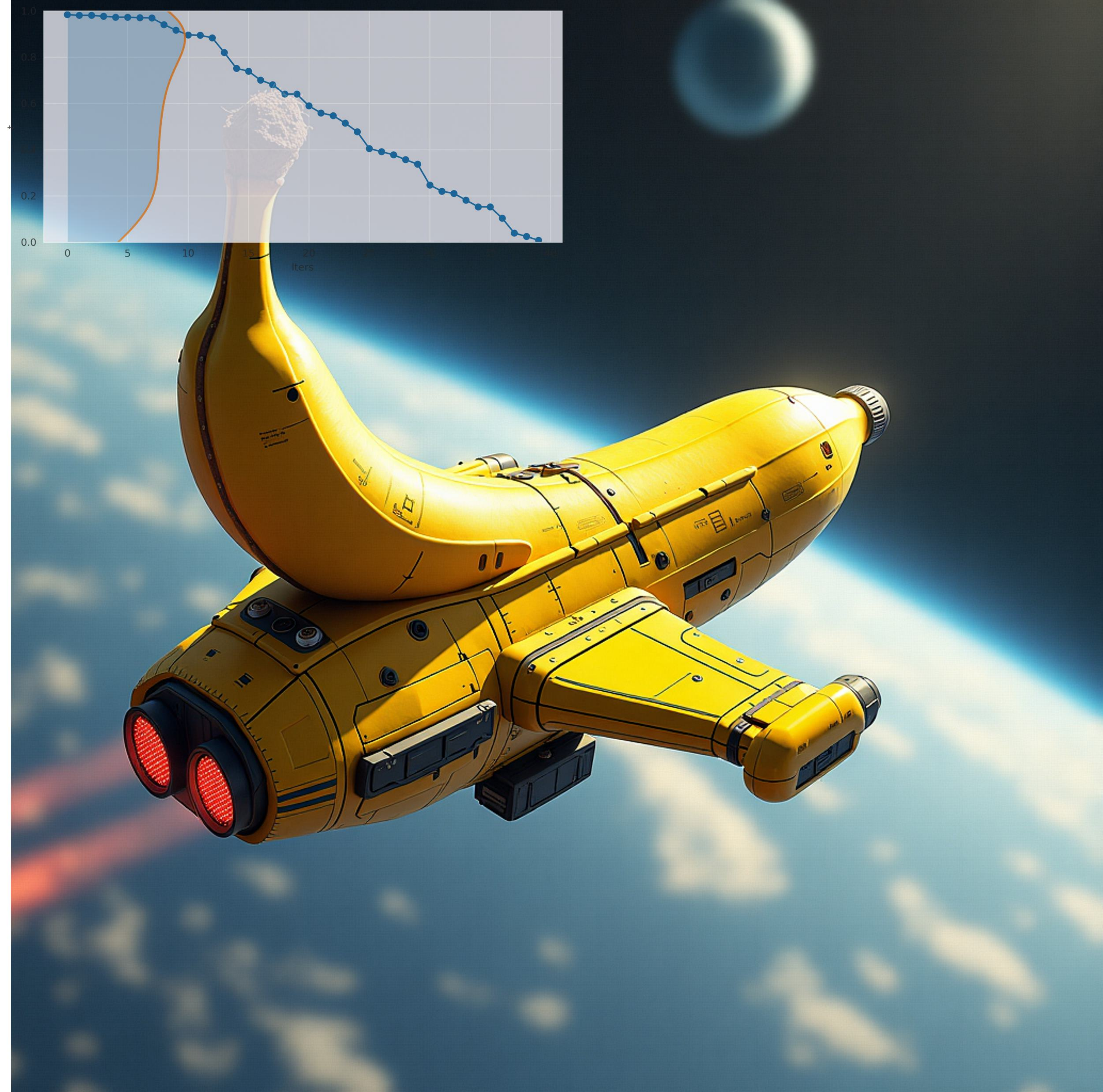}
  \label{fig:appx_gt_r2c1}
\end{subfigure}\hfill
\begin{subfigure}[t]{0.315\linewidth}
  \centering
  \includegraphics[width=\linewidth]{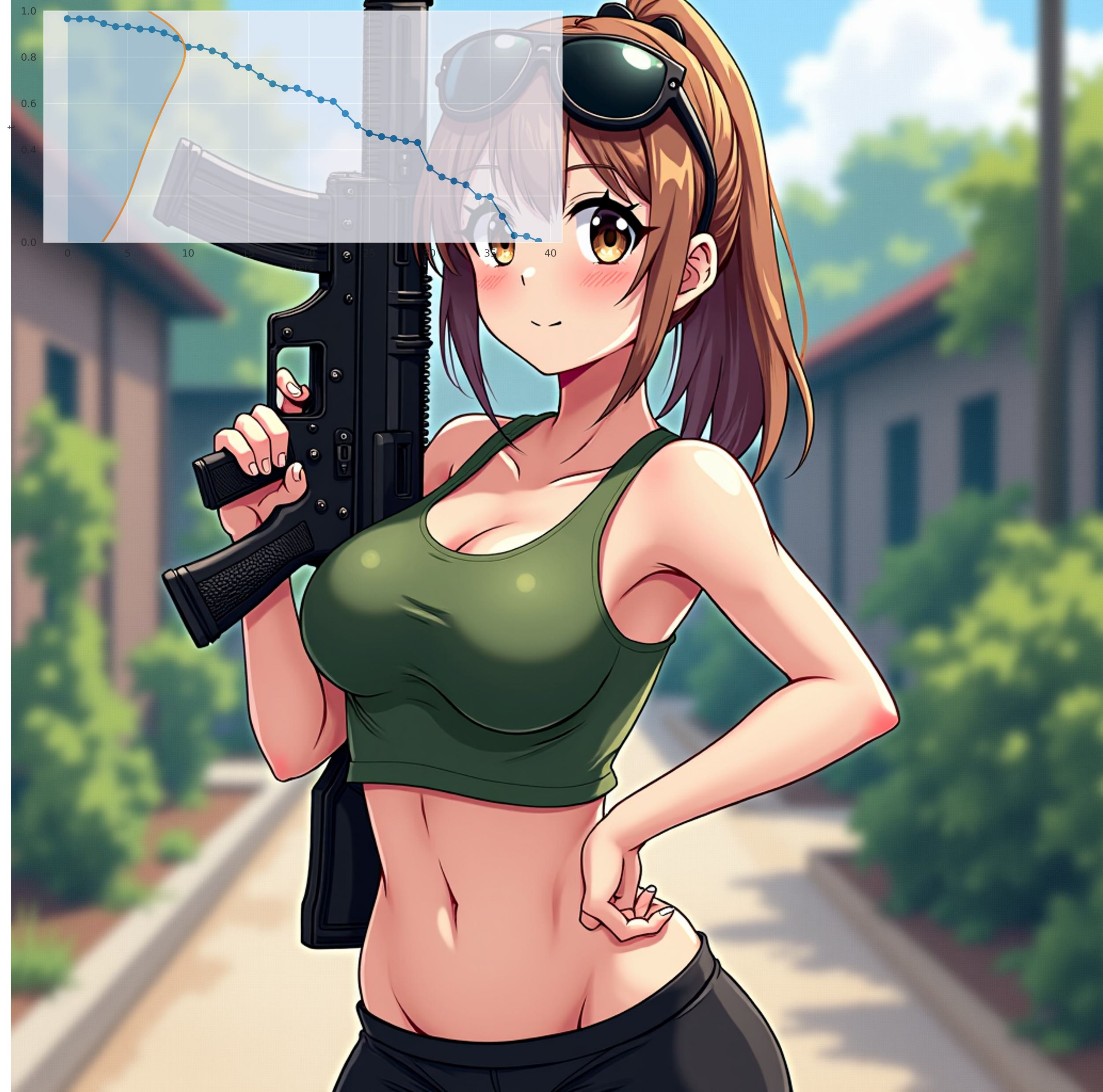}
  \label{fig:appx_gt_r2c2}
\end{subfigure}\hfill
\begin{subfigure}[t]{0.315\linewidth}
  \centering
  \includegraphics[width=\linewidth]{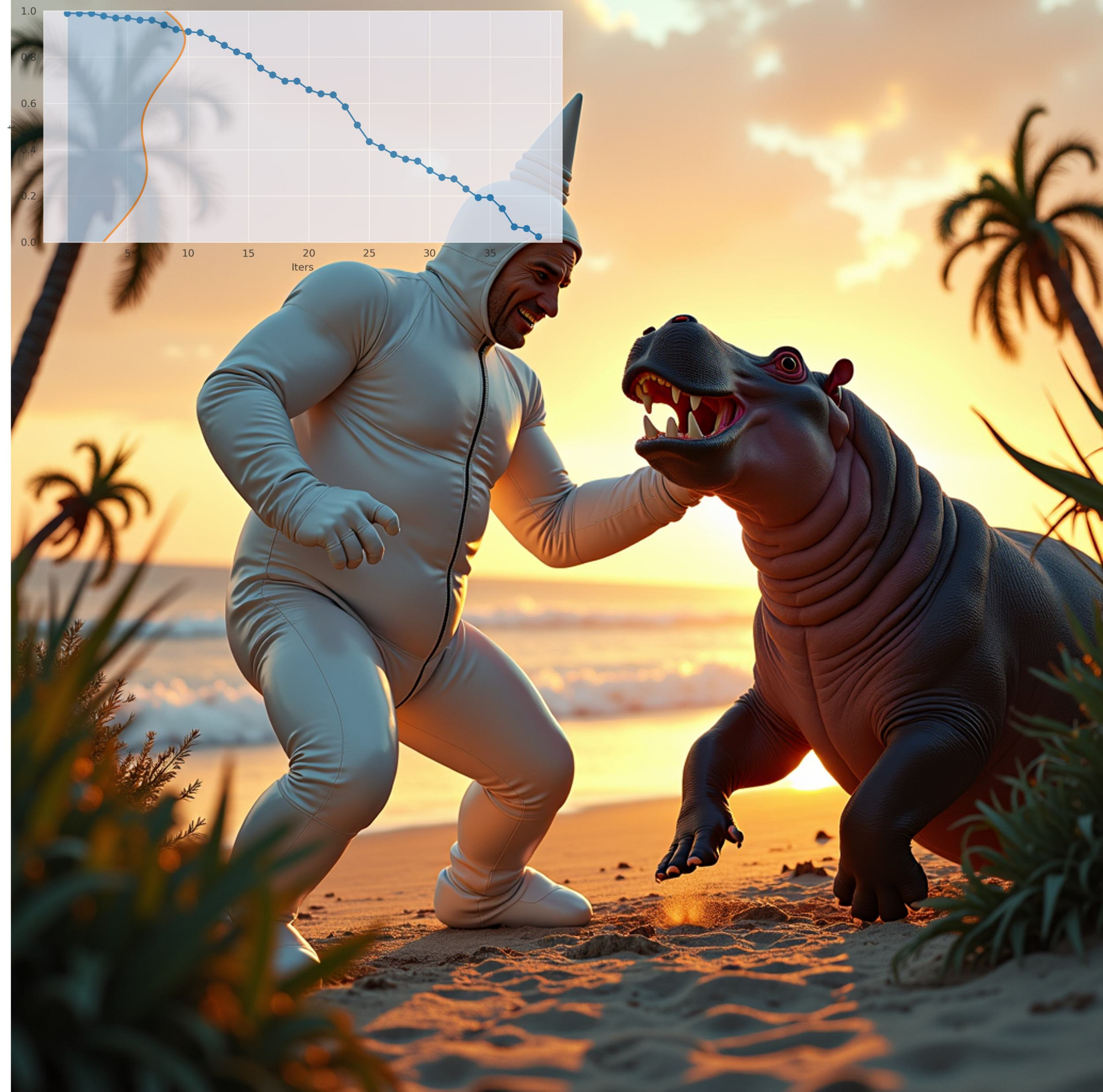}
  \label{fig:appx_gt_r2c3}
\end{subfigure}
\vspace{-10pt}
\caption{\textbf{Rescheduling improves general T2I alignment.} Head-to-head comparisons between images generated with default schedules (upper) and our learned schedules (lower) from Flux-Dev with 40 steps.}
\label{fig:appx_gen_t2i_1}
\end{figure*}

\begin{figure*}[t]
\centering

\colhead{The image depicts a muscular mouse wielding assault rifles, in a Disney art style.}\hfill
\colhead{Yoda {\color{red} performing at Woodstock}.}\hfill
\colhead{Photo of a {\color{red} chocolate-type Pokemon card}.}

\begin{subfigure}[t]{0.315\linewidth}
  \centering
  \includegraphics[width=\linewidth]{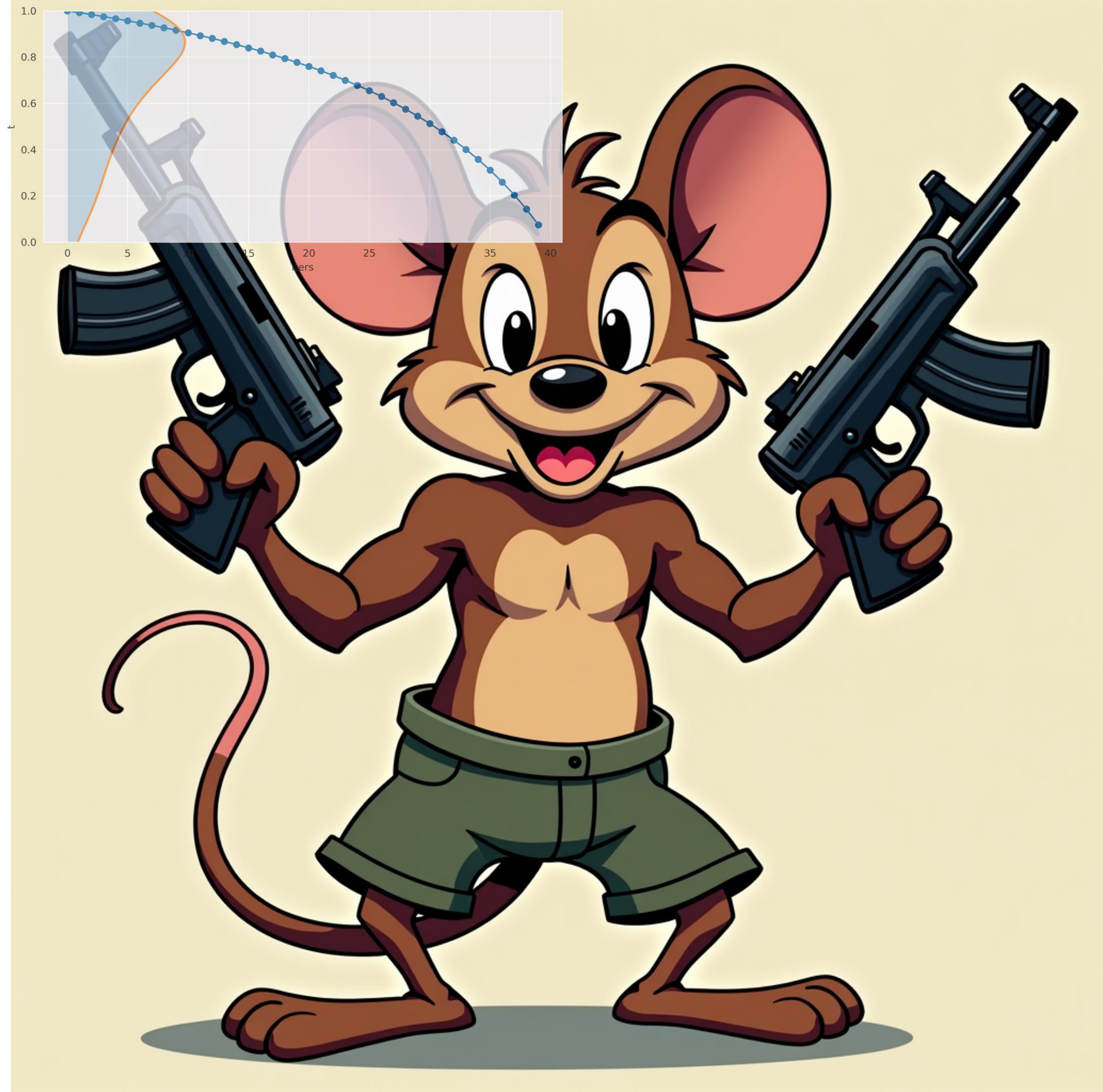}
  \label{fig:appx_gt2_r1c1}
\end{subfigure}\hfill
\begin{subfigure}[t]{0.315\linewidth}
  \centering
  \includegraphics[width=\linewidth]{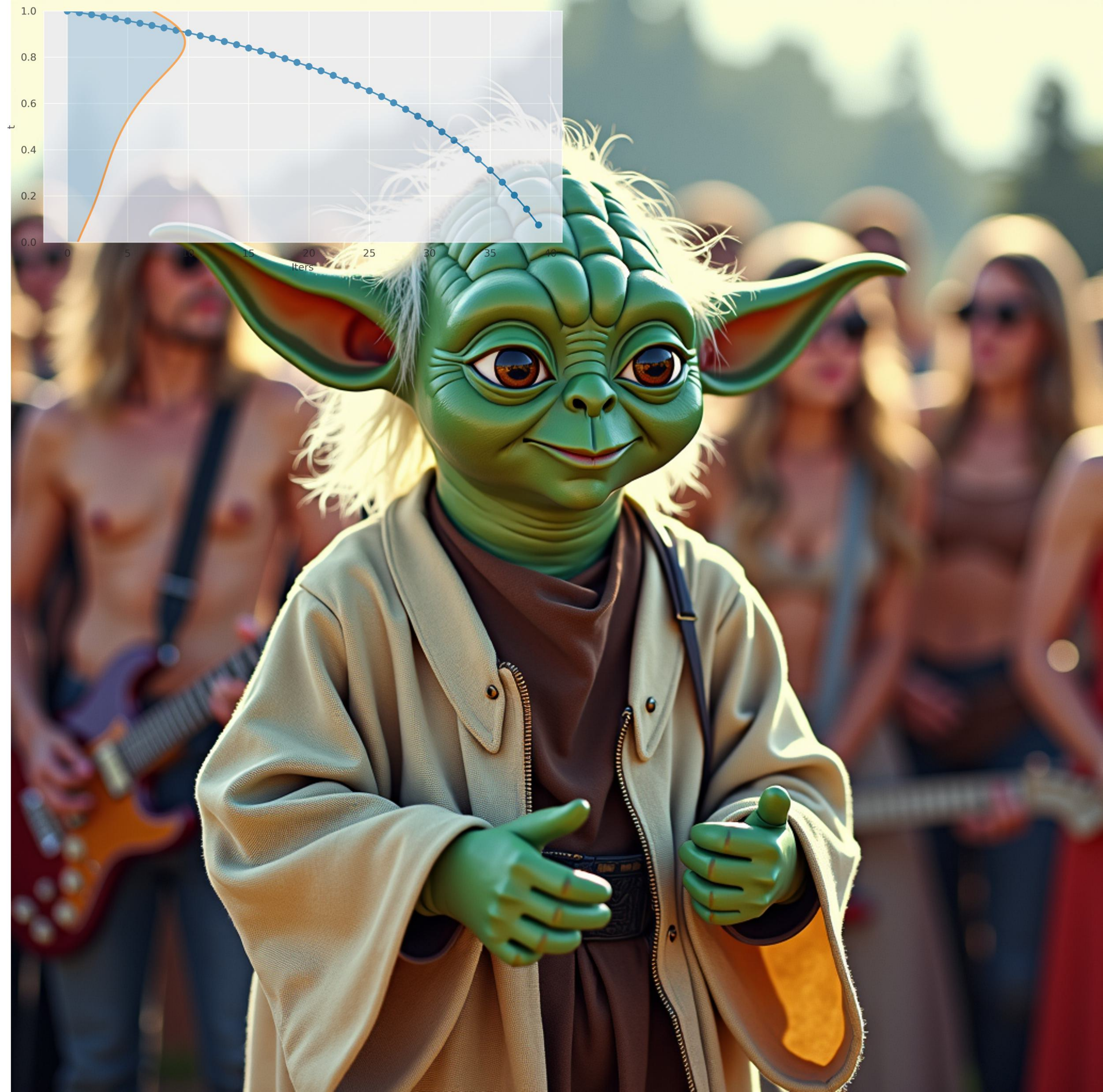}
  \label{fig:appx_gt2_r1c2}
\end{subfigure}\hfill
\begin{subfigure}[t]{0.315\linewidth}
  \centering
  \includegraphics[width=\linewidth]{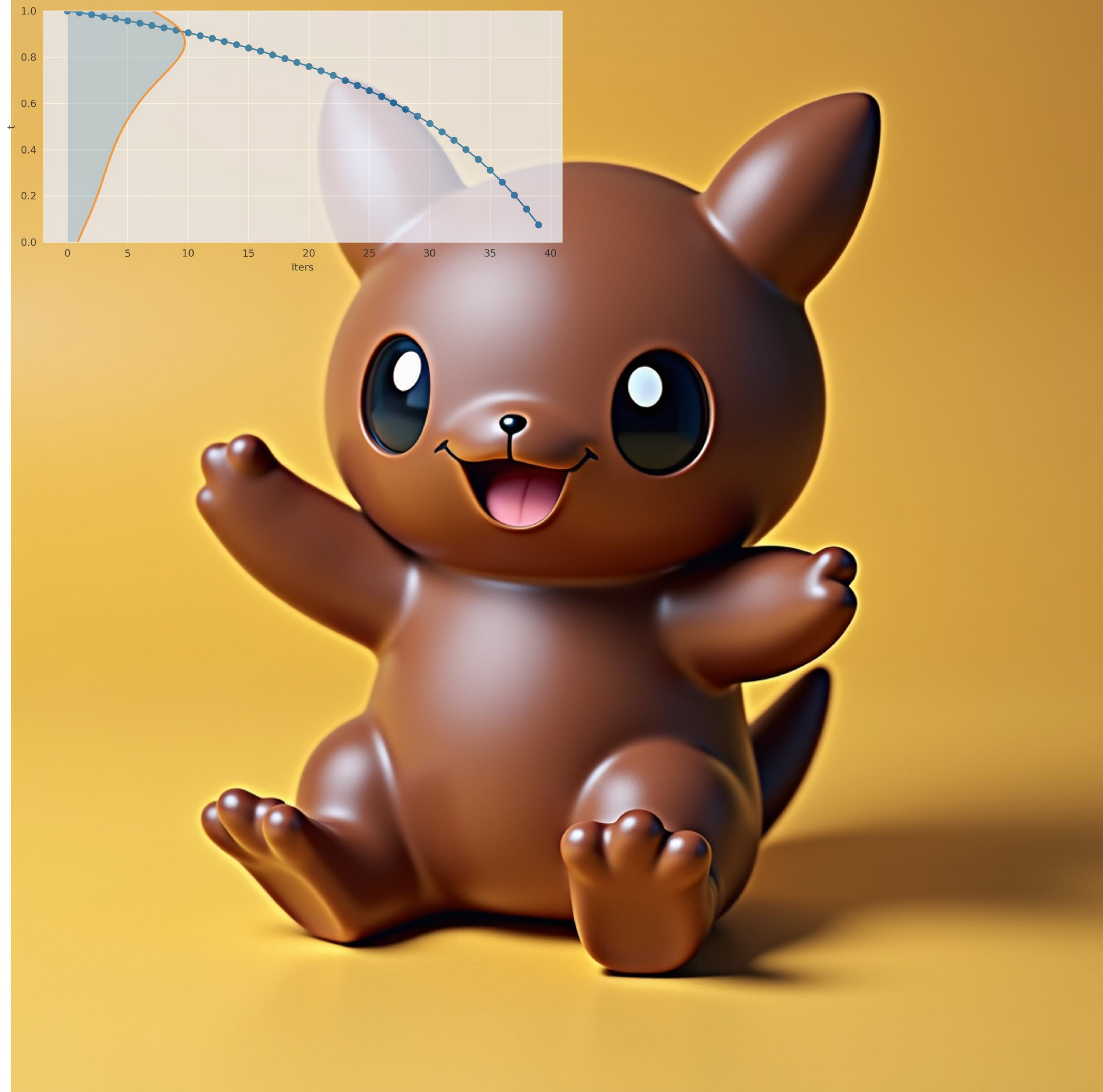}
  \label{fig:appx_gt2_r1c3}
\end{subfigure}

\vspace{-10pt} 

\begin{subfigure}[t]{0.315\linewidth}
  \centering
  \includegraphics[width=\linewidth]{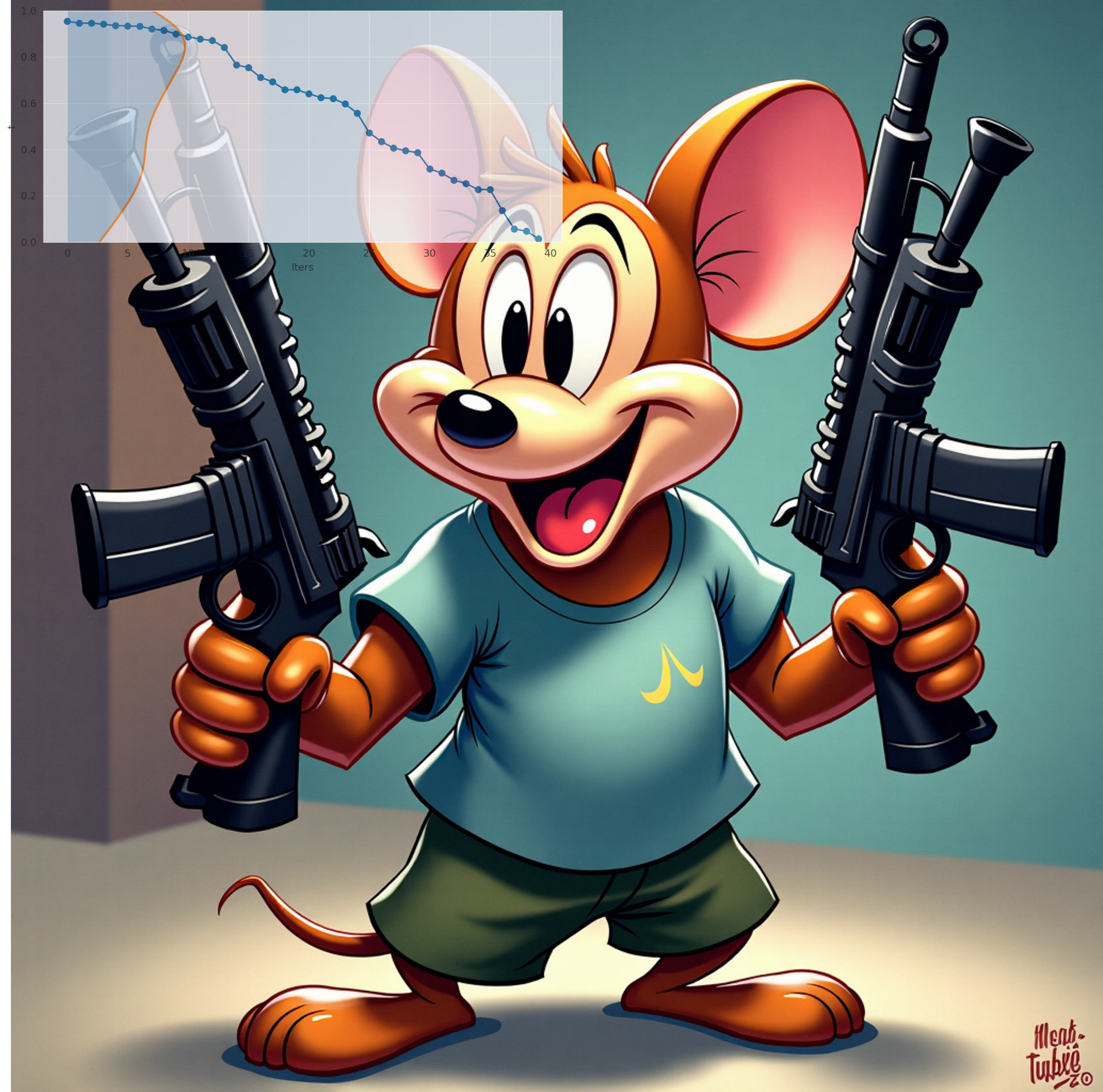}
  \label{fig:appx_gt2_r2c1}
\end{subfigure}\hfill
\begin{subfigure}[t]{0.315\linewidth}
  \centering
  \includegraphics[width=\linewidth]{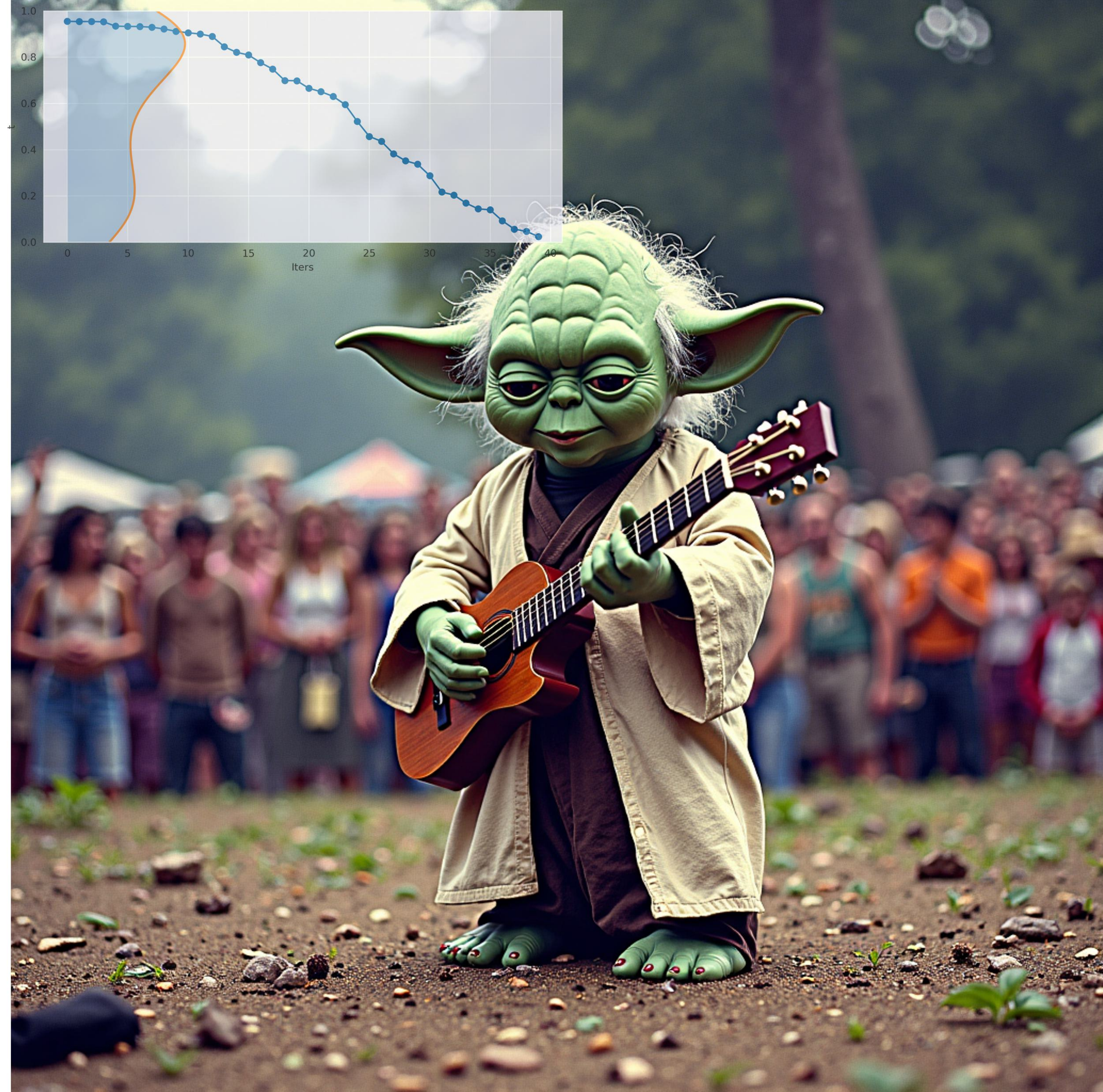}
  \label{fig:appx_gt2_r2c2}
\end{subfigure}\hfill
\begin{subfigure}[t]{0.315\linewidth}
  \centering
  \includegraphics[width=\linewidth]{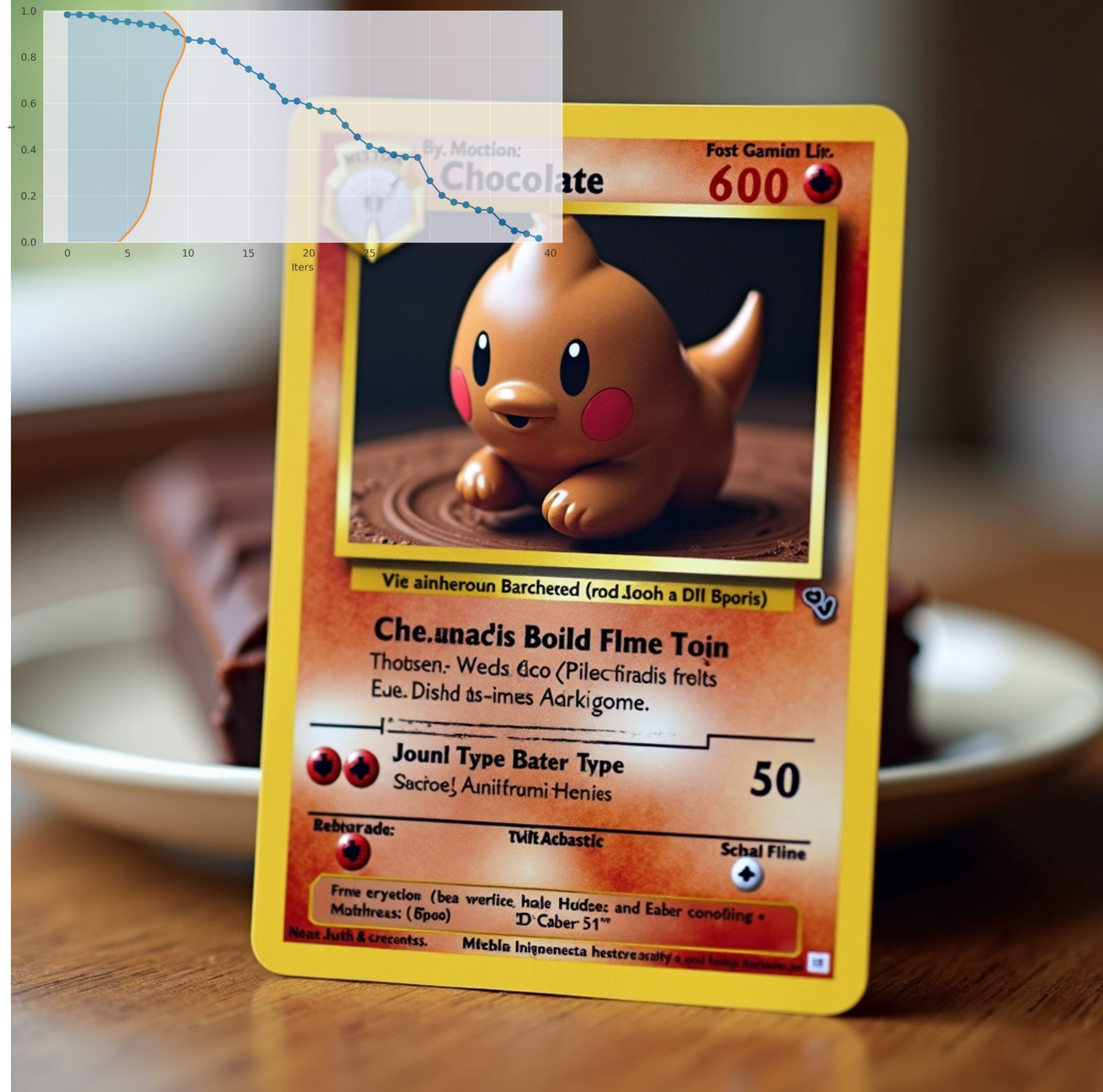}
  \label{fig:appx_gt2_r2c3}
\end{subfigure}
\vspace{-10pt}
\caption{\textbf{Rescheduling improves general T2I alignment.} Head-to-head comparisons between images generated with default schedules (upper) and our learned schedules (lower) from Flux-Dev with 40 steps.}
\label{fig:appx_gen_t2i_2}
\end{figure*}

\begin{figure*}[t]
\centering

\colhead{... that reads {\color{red}``Abandon All Hope''} in eerie, gothic lettering, set against a moonlit night...}\hfill
\colhead{...  featuring a glowing {\color{red}``AI Training Zone''} hologram floating in the center...}\hfill
\colhead{... prominently displaying {\color{red}``Staff Pick Book 15''} next to a stack of books...}

\begin{subfigure}[t]{0.315\linewidth}
  \centering  \includegraphics[width=\linewidth]{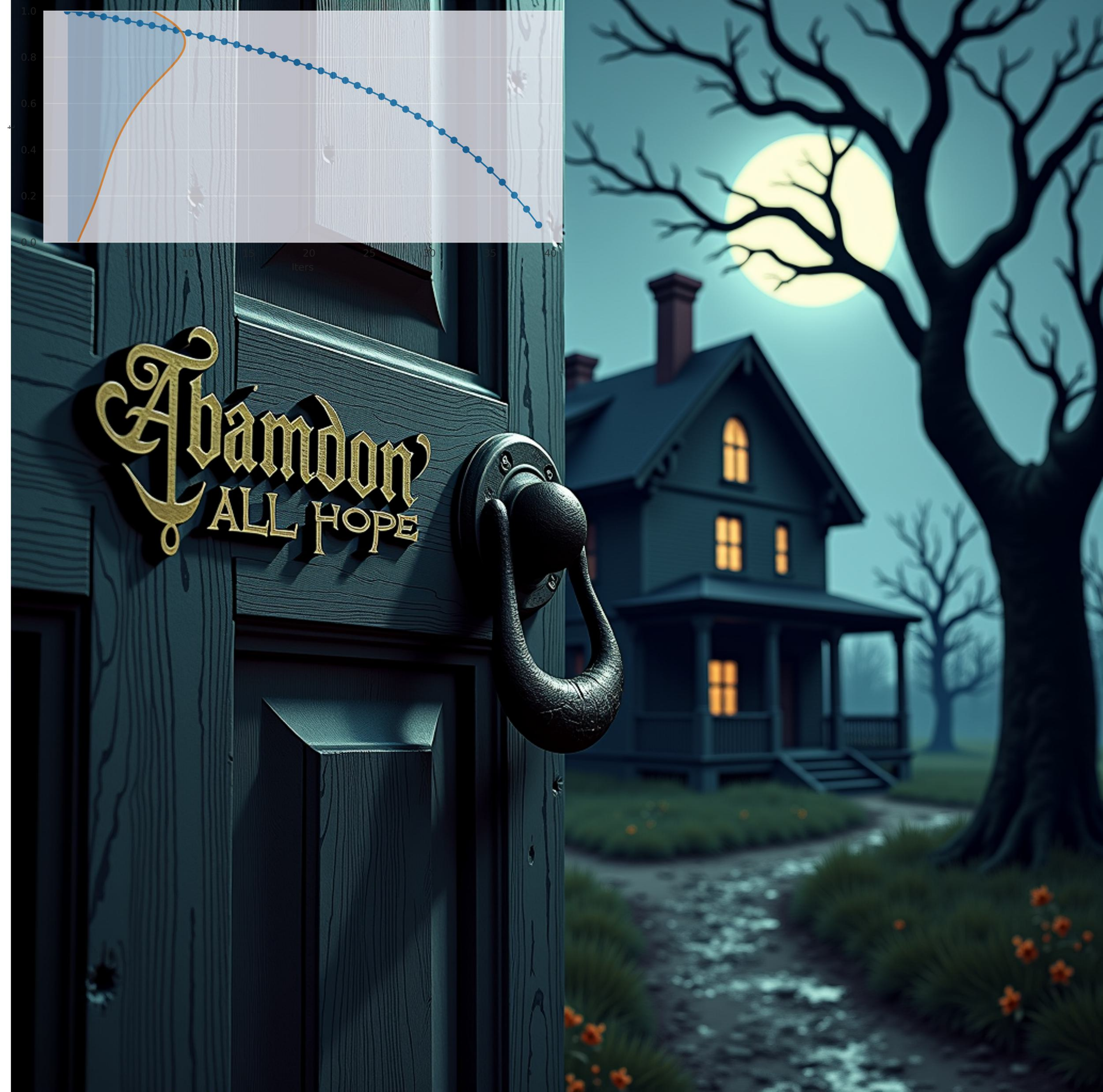}
  \label{fig:appx_tr_r1c1}
\end{subfigure}\hfill
\begin{subfigure}[t]{0.315\linewidth}
  \centering
  \includegraphics[width=\linewidth]{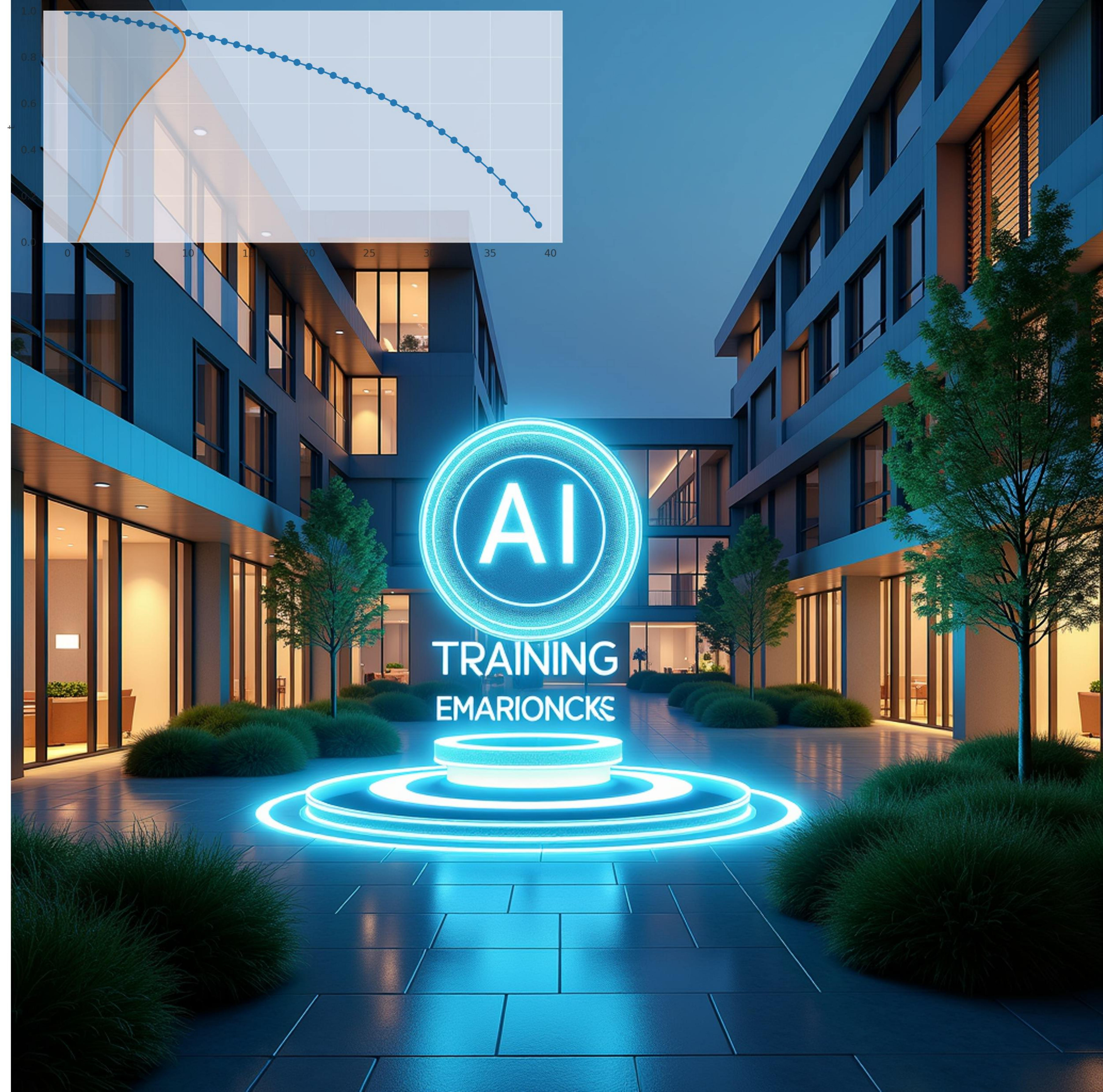}
  \label{fig:appx_tr_r1c2}
\end{subfigure}\hfill
\begin{subfigure}[t]{0.315\linewidth}
  \centering
  \includegraphics[width=\linewidth]{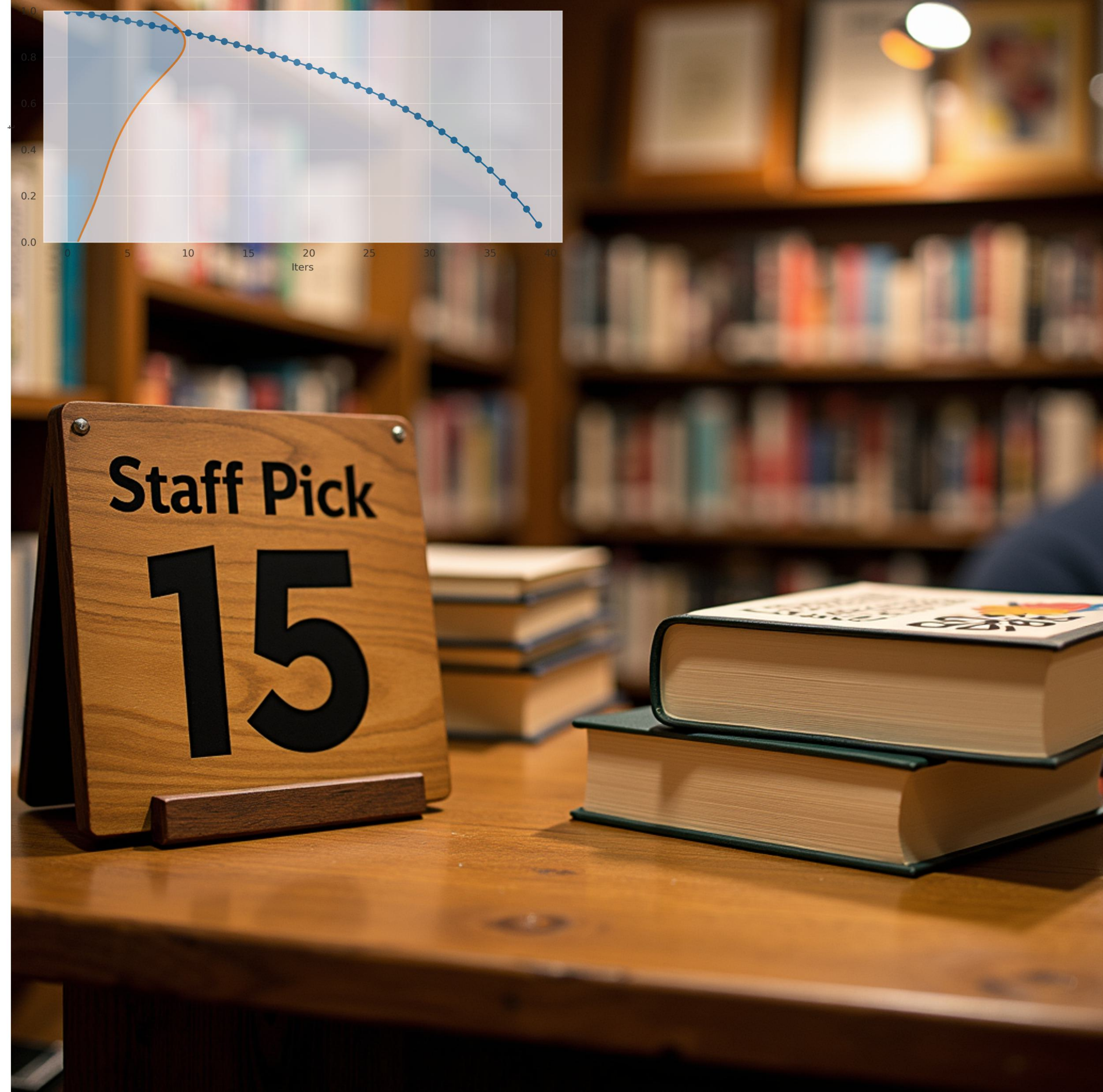}
  \label{fig:appx_tr_r1c3}
\end{subfigure}

\vspace{-10pt} 

\begin{subfigure}[t]{0.315\linewidth}
  \centering
  \includegraphics[width=\linewidth]{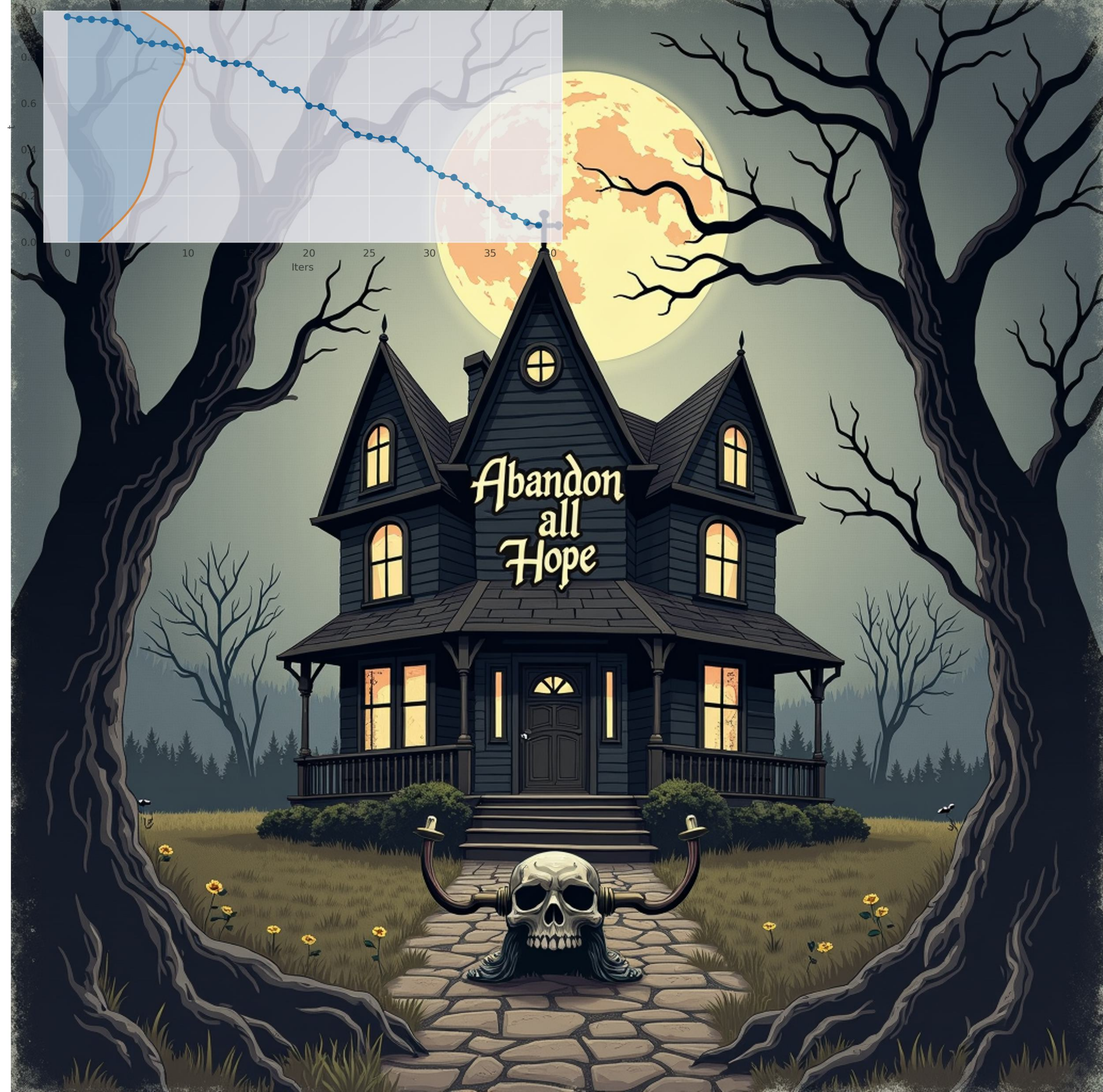}
  \label{fig:appx_tr_r2c1}
\end{subfigure}\hfill
\begin{subfigure}[t]{0.315\linewidth}
  \centering
  \includegraphics[width=\linewidth]{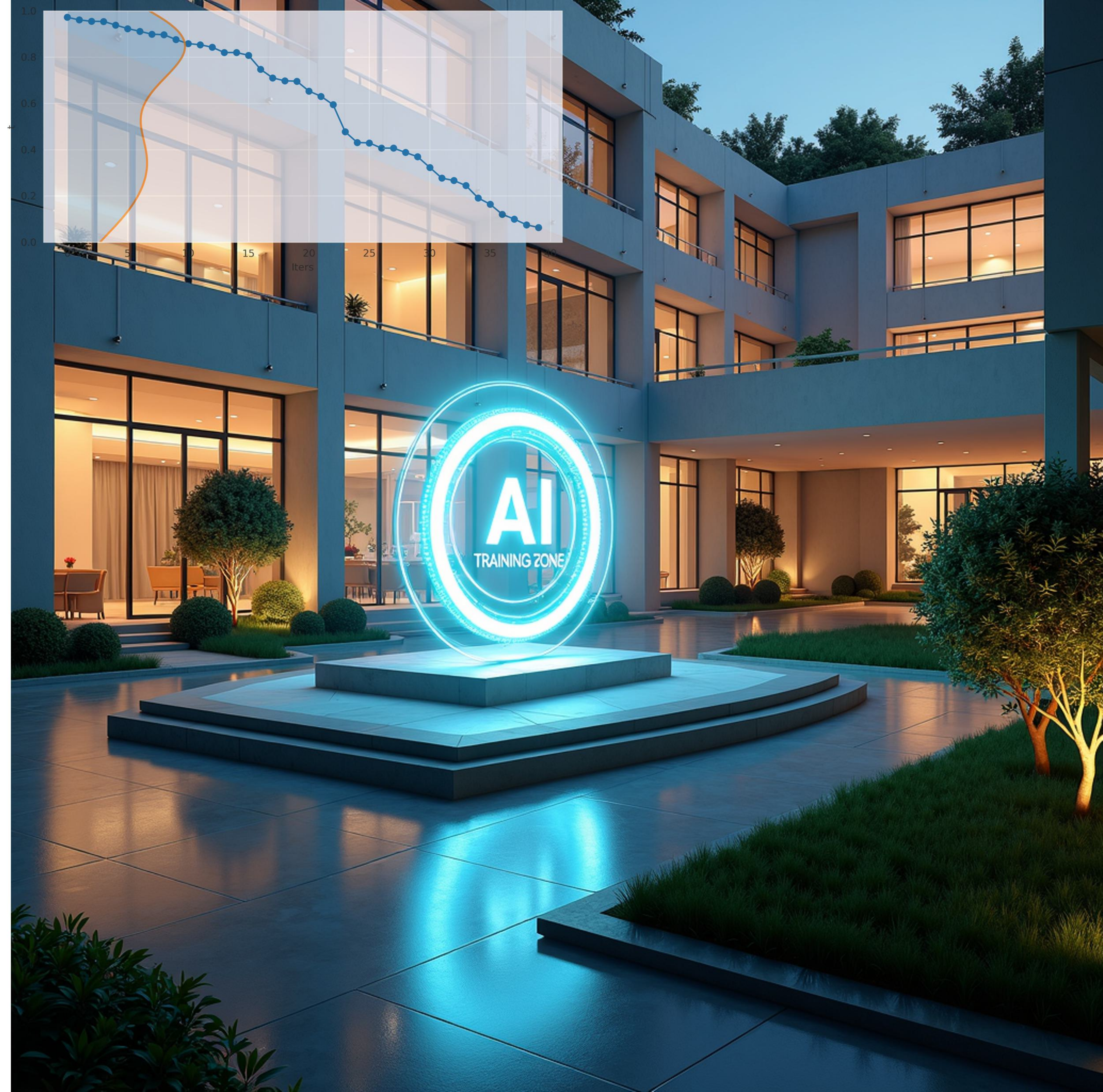}
  \label{fig:appx_tr_r2c2}
\end{subfigure}\hfill
\begin{subfigure}[t]{0.315\linewidth}
  \centering
  \includegraphics[width=\linewidth]{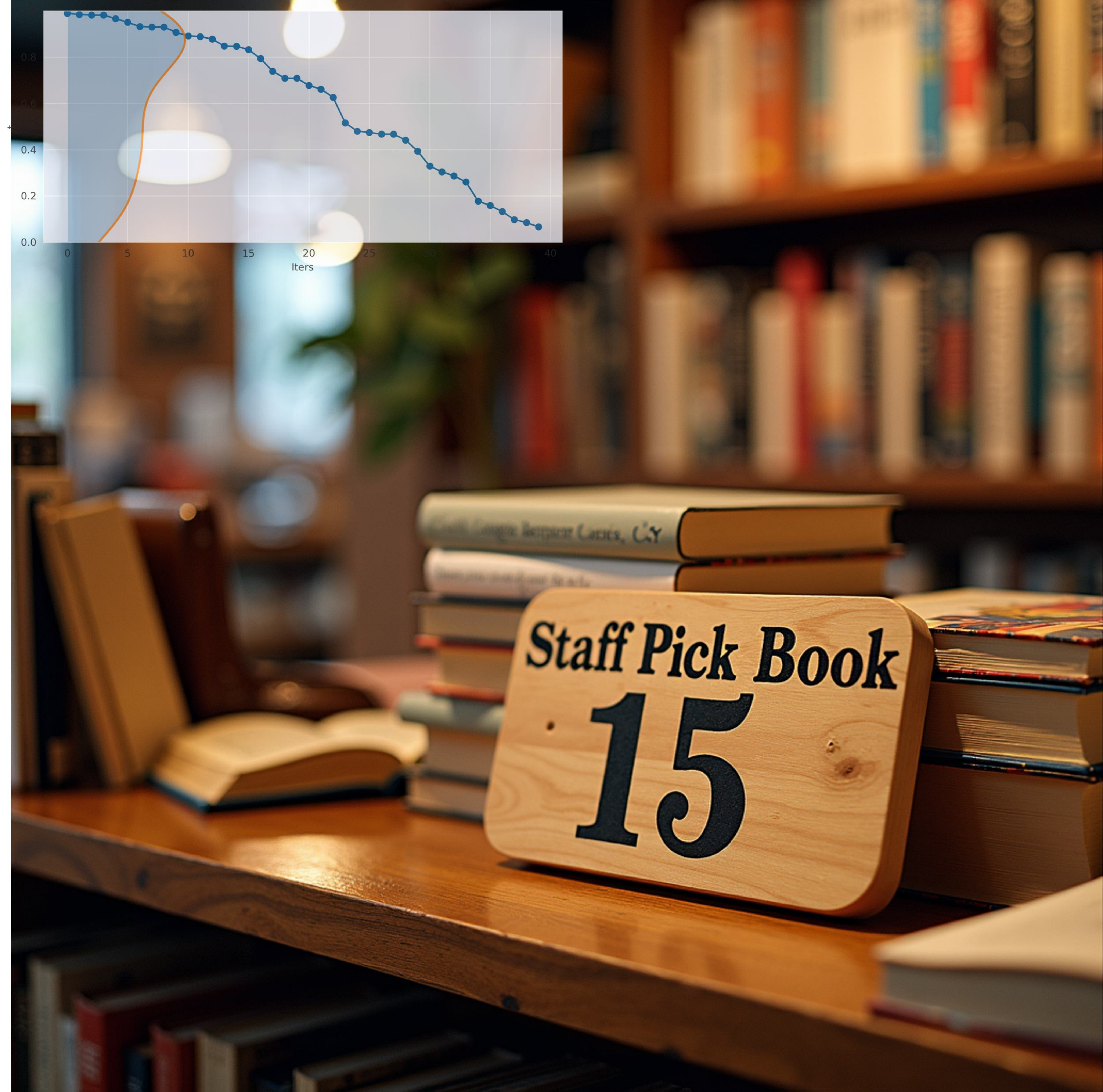}
  \label{fig:appx_tr_r2c3}
\end{subfigure}
\vspace{-10pt}
\caption{\textbf{Rescheduling improves text rendering}. We present comparisons between images generated with default schedules (upper) and our learned schedules (lower) from Flux-Dev.}
\label{fig:appx_text_render}
\vspace{-4pt}
\end{figure*}

\begin{figure*}[t]
\centering

\colhead{A book saying {\color{red}``How to Disappear Completely''}...}\hfill
\colhead{A hiking trail with a wooden signpost clearly displaying {\color{red}``Private Property No Entry''}, surrounded by dense, green foliage...}\hfill
\colhead{A vibrant skateboard deck featuring the bold graphic {\color{red}``SKATE OR DIE 4EVER''} in dynamic, graffiti-style lettering...}

\begin{subfigure}[t]{0.315\linewidth}
  \centering  \includegraphics[width=\linewidth]{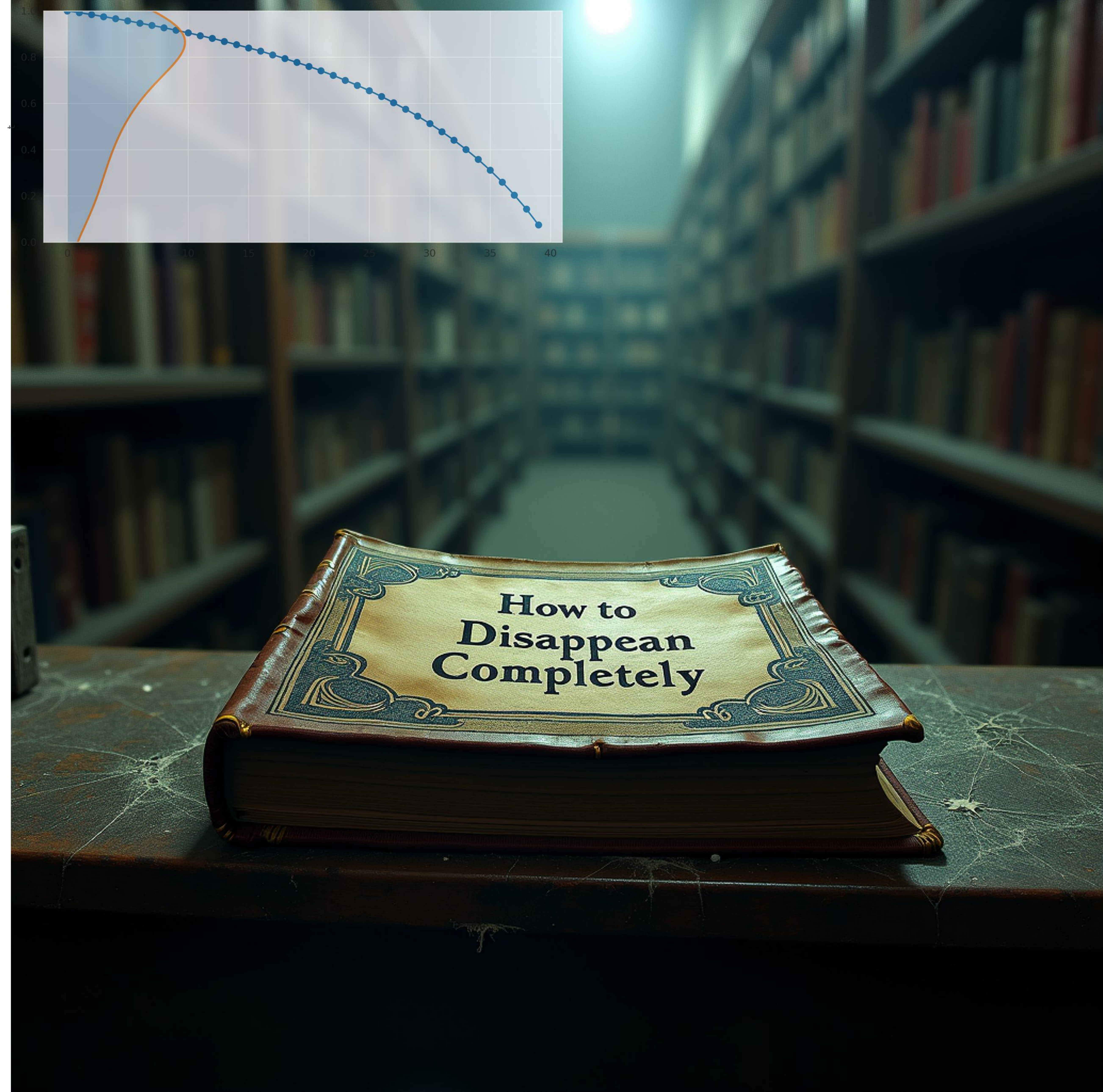}
  \label{fig:appx_tr2_r1c1}
\end{subfigure}\hfill
\begin{subfigure}[t]{0.315\linewidth}
  \centering
  \includegraphics[width=\linewidth]{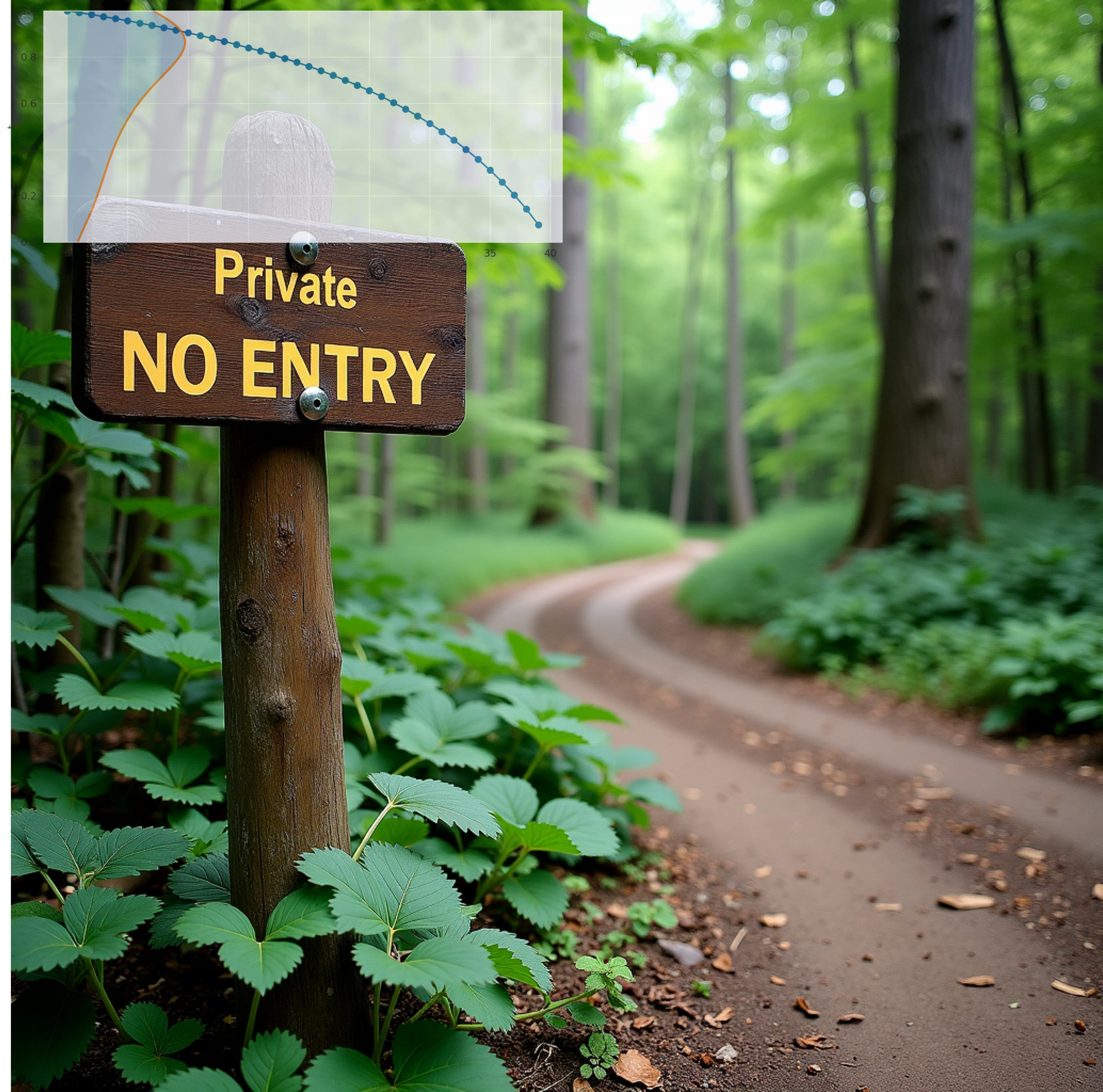}
  \label{fig:appx_tr2_r1c2}
\end{subfigure}\hfill
\begin{subfigure}[t]{0.315\linewidth}
  \centering
  \includegraphics[width=\linewidth]{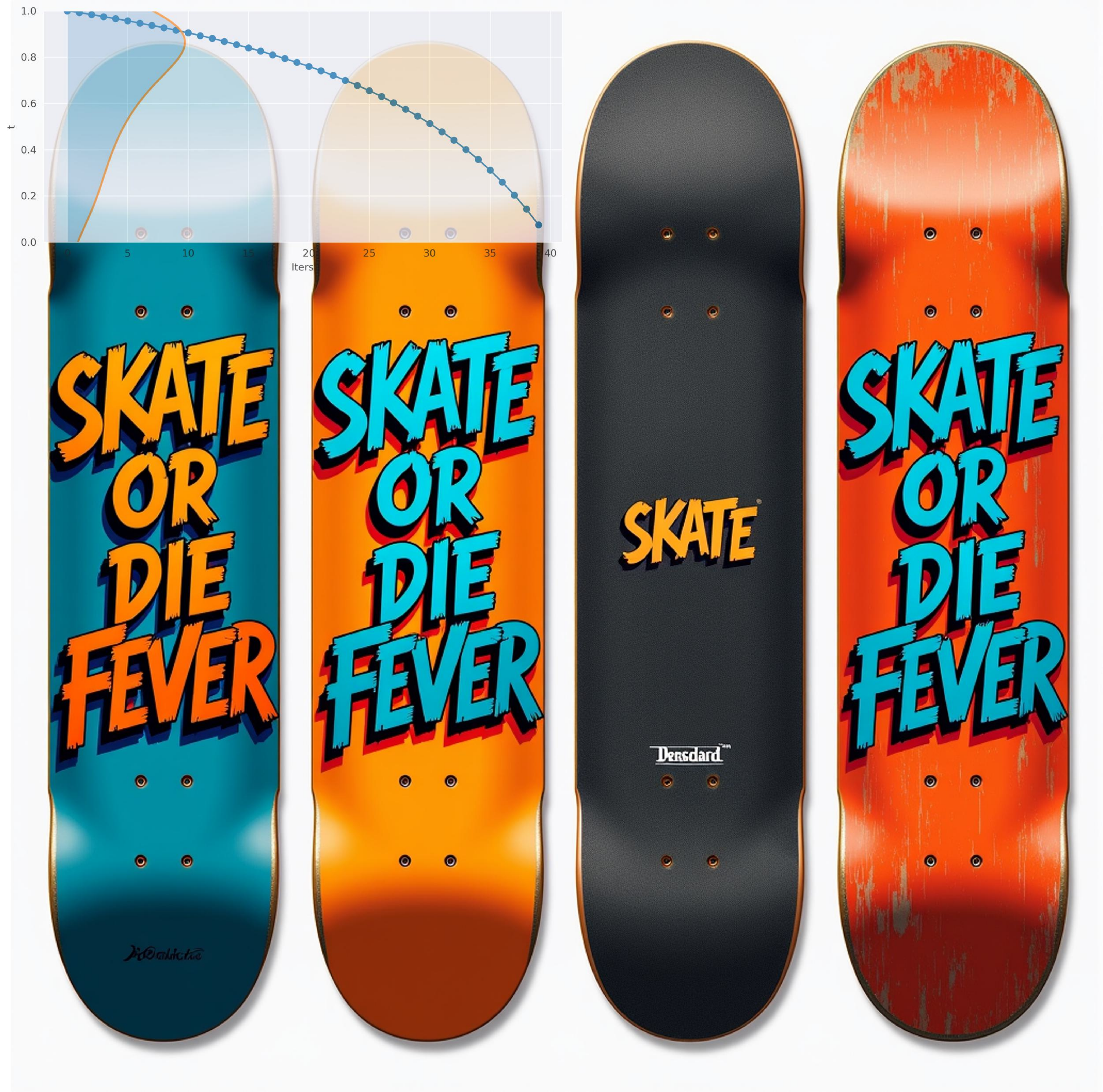}
  \label{fig:appx_tr2_r1c3}
\end{subfigure}

\vspace{-10pt} 

\begin{subfigure}[t]{0.315\linewidth}
  \centering
  \includegraphics[width=\linewidth]{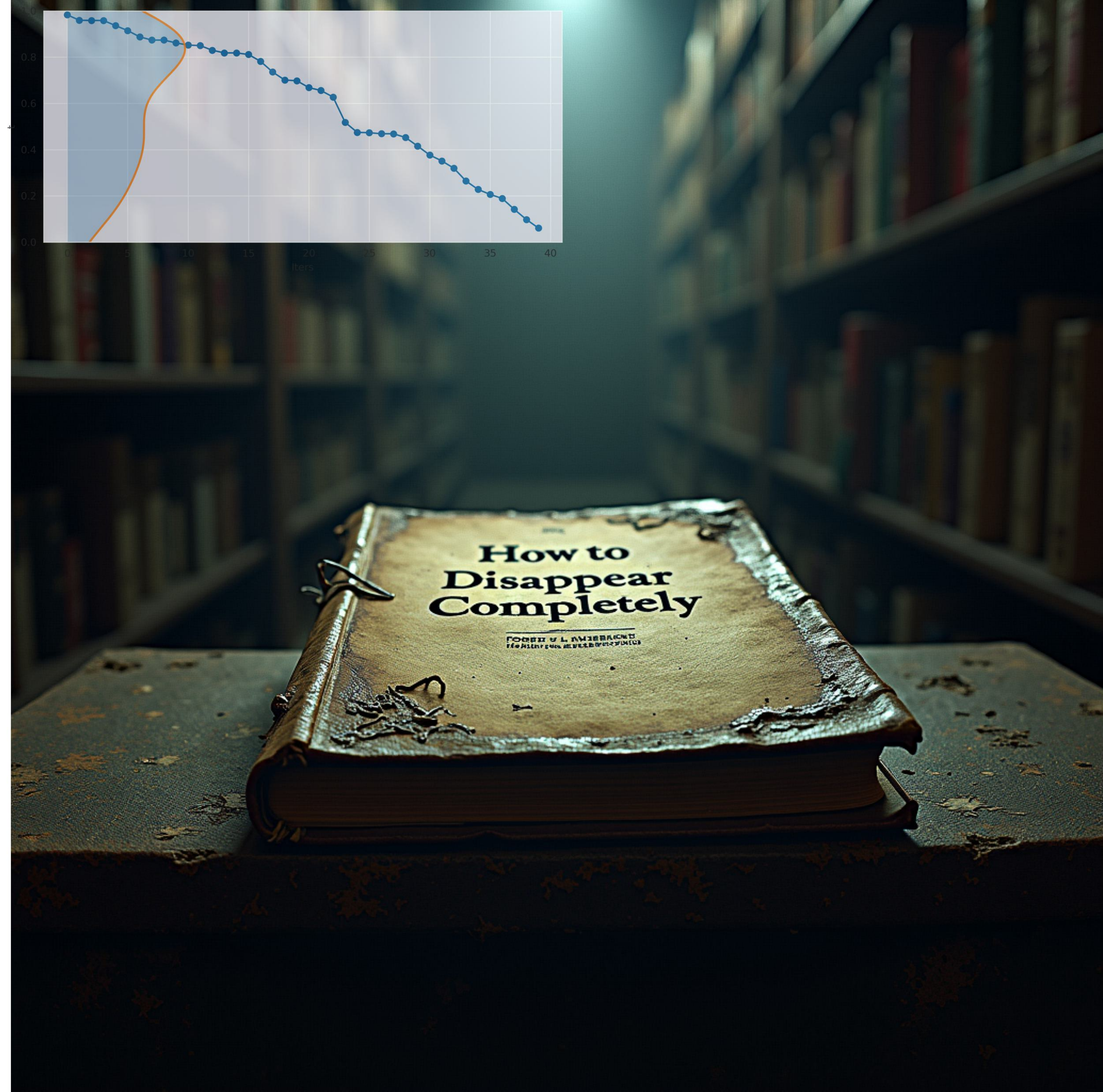}
  \label{fig:appx_tr2_r2c1}
\end{subfigure}\hfill
\begin{subfigure}[t]{0.315\linewidth}
  \centering
  \includegraphics[width=\linewidth]{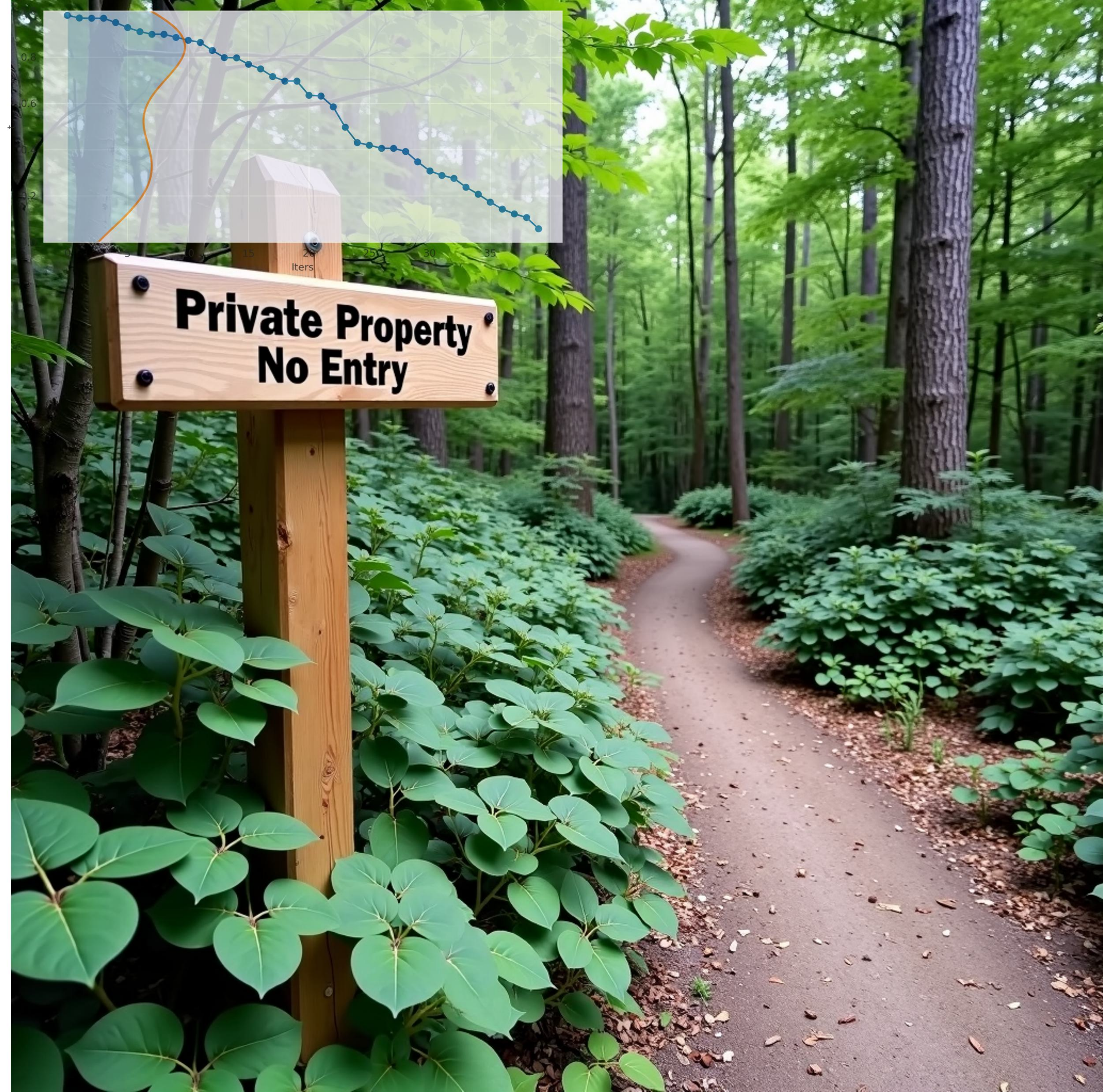}
  \label{fig:appx_tr2_r2c2}
\end{subfigure}\hfill
\begin{subfigure}[t]{0.315\linewidth}
  \centering
  \includegraphics[width=\linewidth]{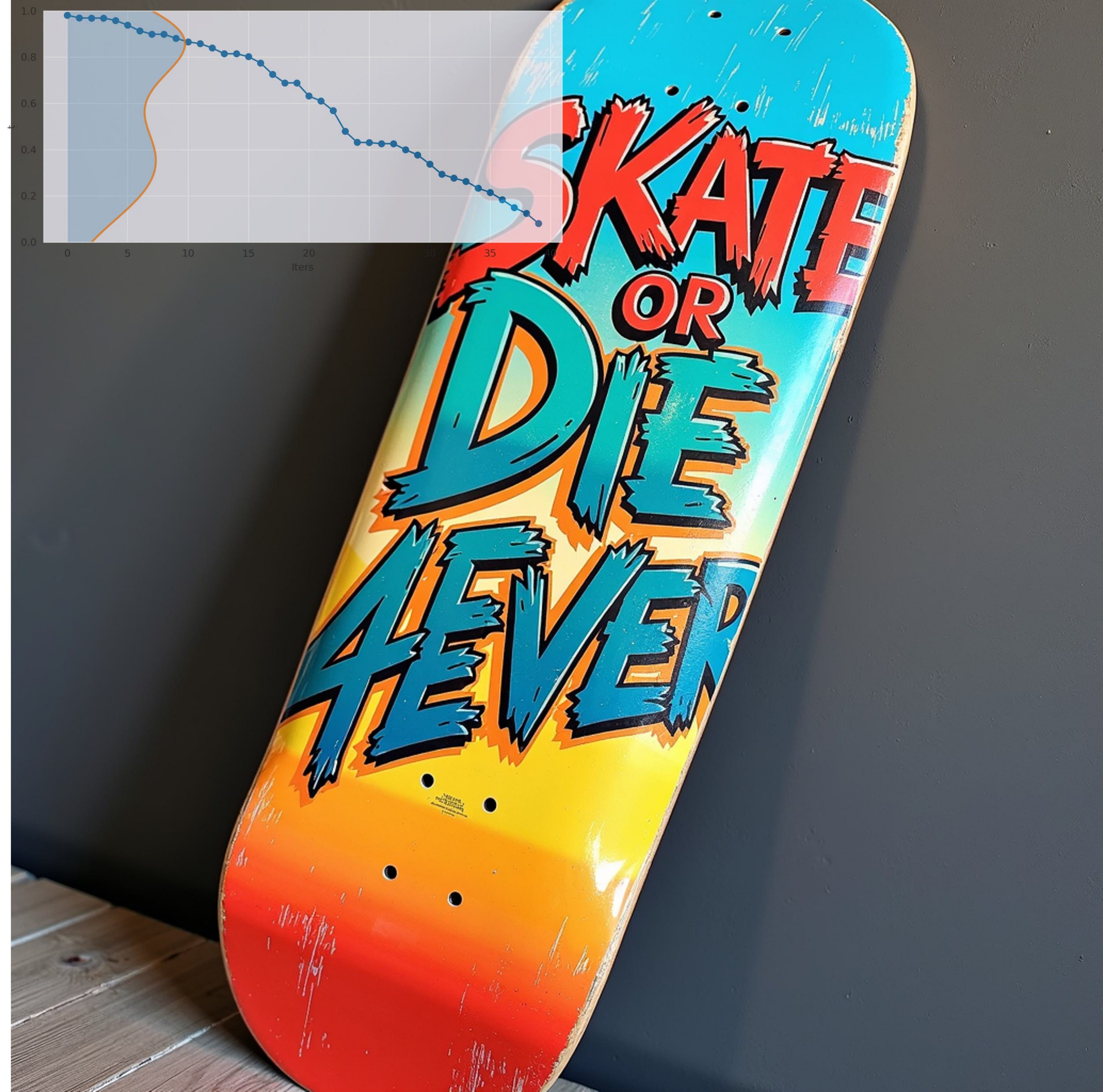}
  \label{fig:appx_tr2_r2c3}
\end{subfigure}
\vspace{-10pt}
\caption{\textbf{Rescheduling improves text rendering}. We present comparisons between images generated with default schedules (upper) and our learned schedules (lower) from Flux-Dev.}
\label{fig:appx_text_render_2}
\vspace{-4pt}
\end{figure*}

\begin{figure*}[t]
\centering

\colhead{a photo of {\color{red}a} carrot and a couch}\hfill
\colhead{a photo of {\color{red}four} clocks}\hfill
\colhead{a photo of a {\color{red}red} potted plant}

\begin{subfigure}[t]{0.315\linewidth}
  \centering  \includegraphics[width=\linewidth]{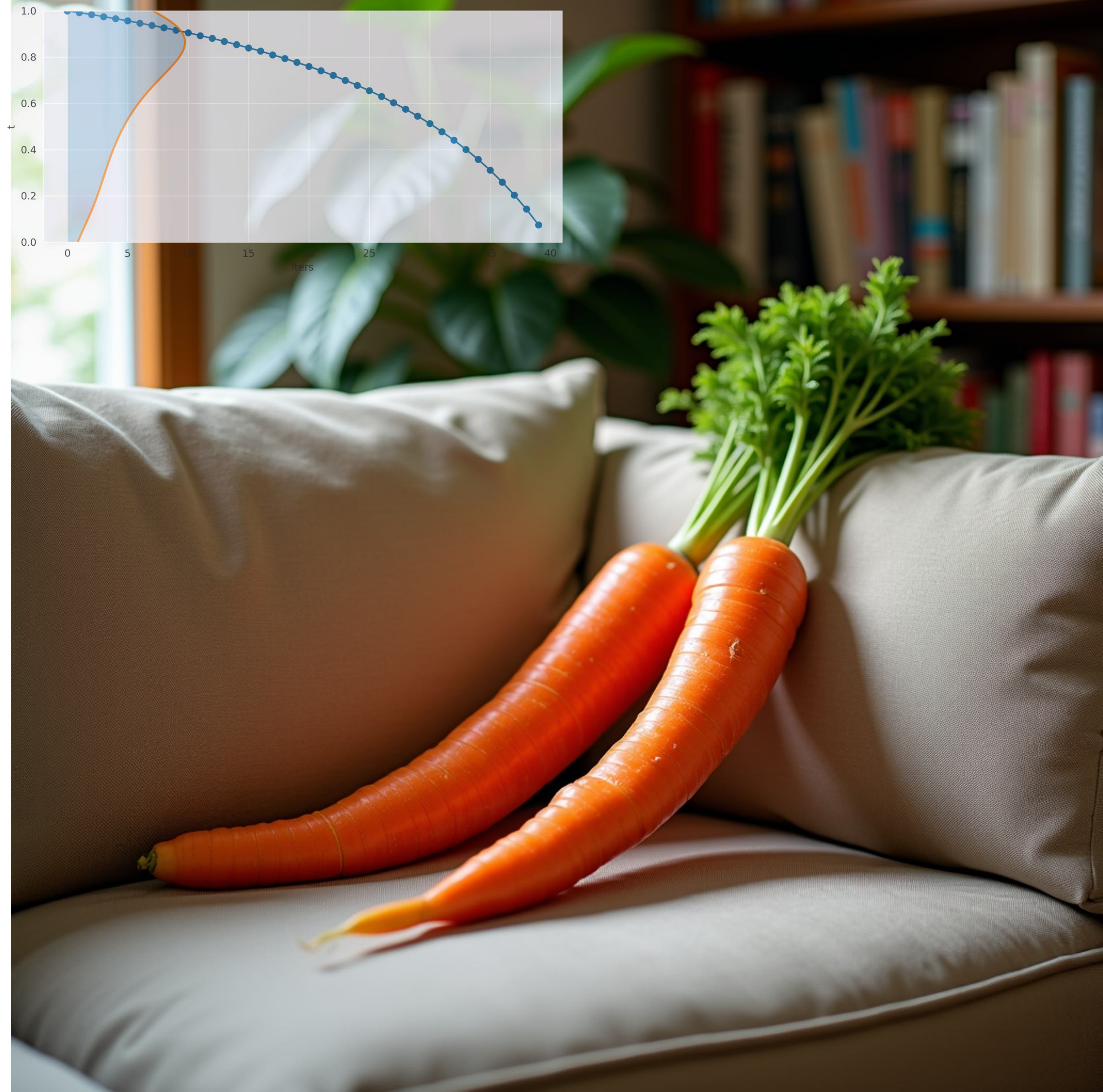}
  \label{fig:appx_ge_r1c1}
\end{subfigure}\hfill
\begin{subfigure}[t]{0.315\linewidth}
  \centering
  \includegraphics[width=\linewidth]{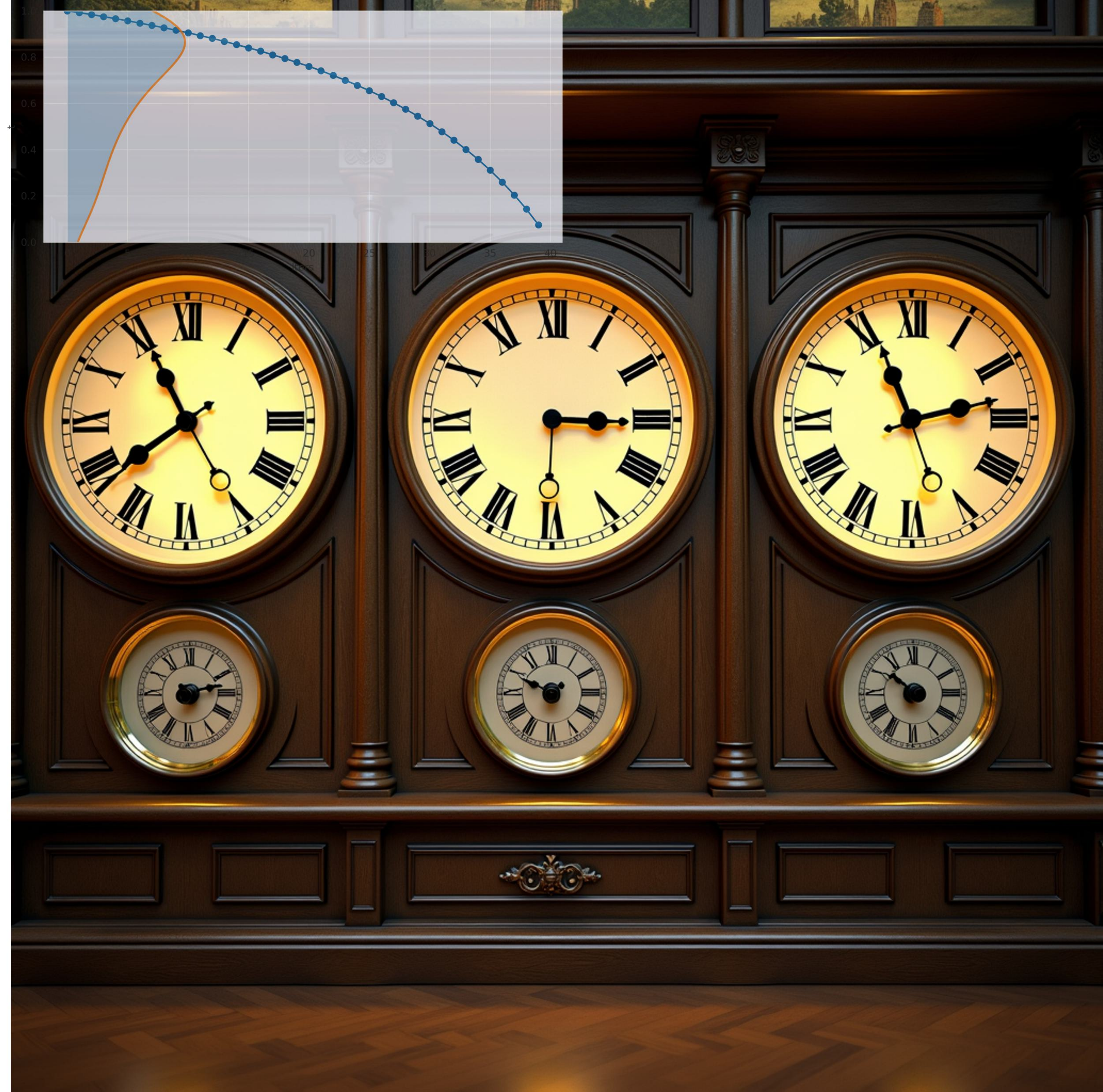}
  \label{fig:appx_ge_r1c2}
\end{subfigure}\hfill
\begin{subfigure}[t]{0.315\linewidth}
  \centering
  \includegraphics[width=\linewidth]{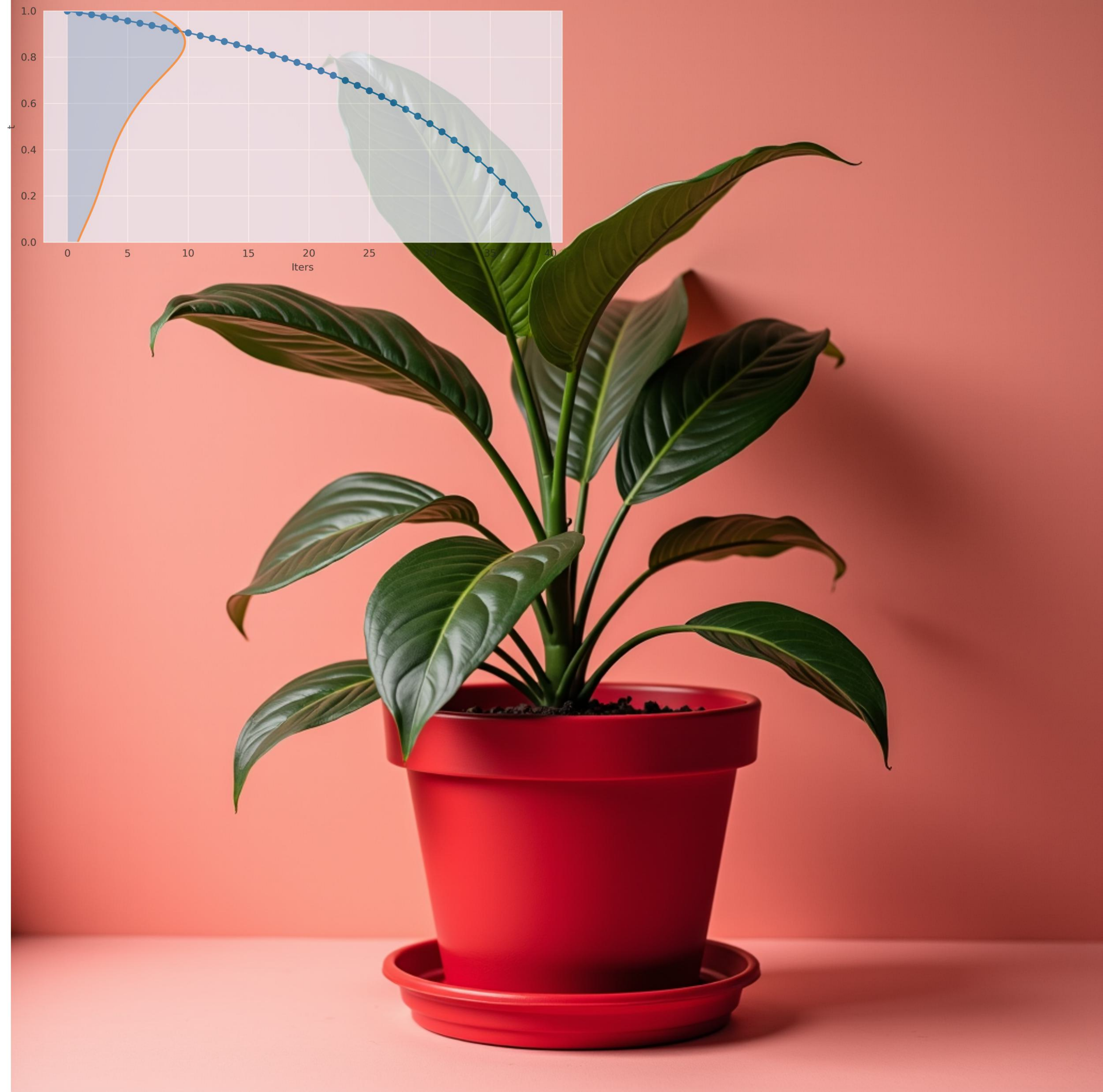}
  \label{fig:appx_ge_r1c3}
\end{subfigure}

\vspace{-10pt} 

\begin{subfigure}[t]{0.315\linewidth}
  \centering
  \includegraphics[width=\linewidth]{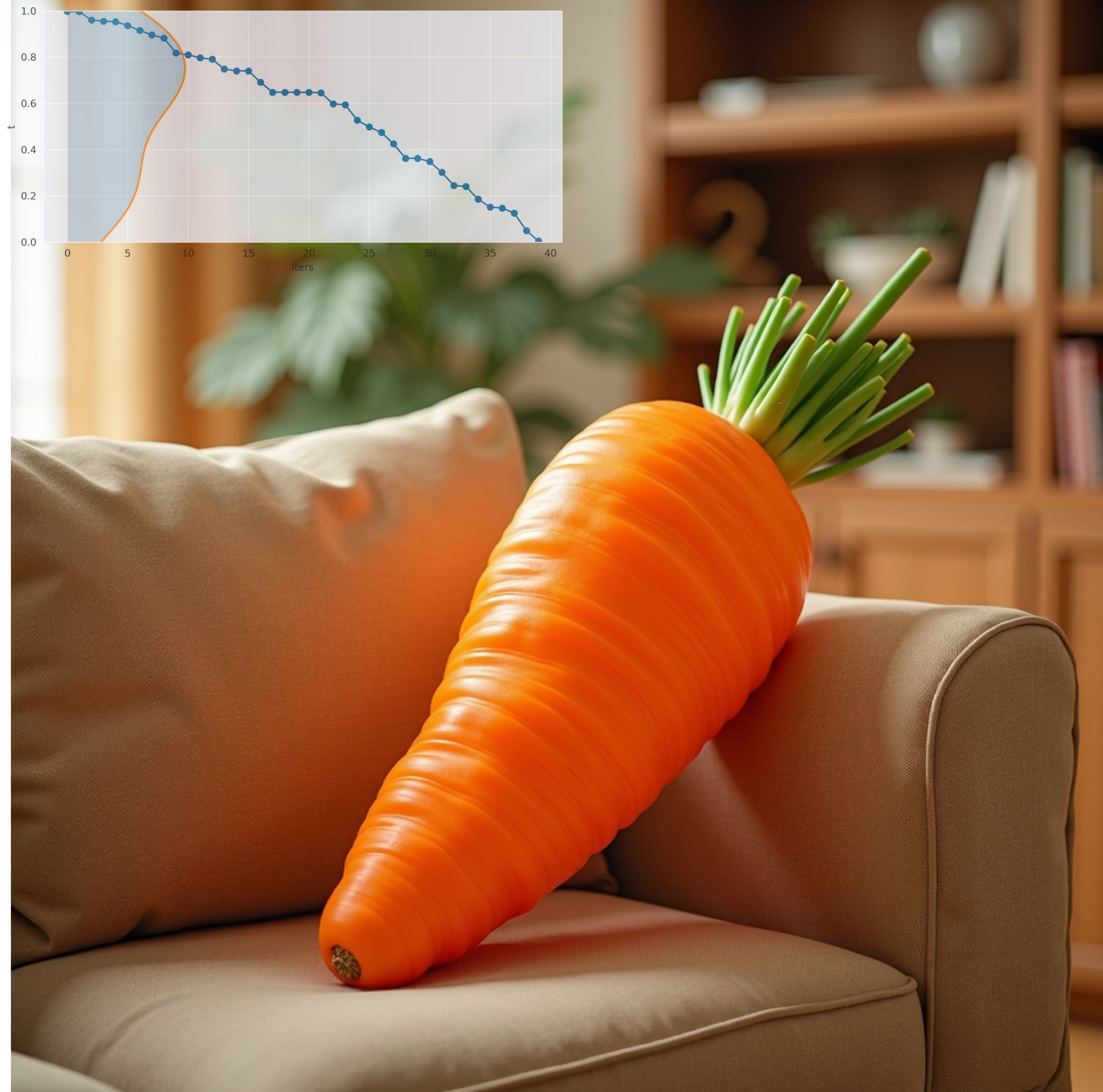}
  \label{fig:appx_ge_r2c1}
\end{subfigure}\hfill
\begin{subfigure}[t]{0.315\linewidth}
  \centering
  \includegraphics[width=\linewidth]{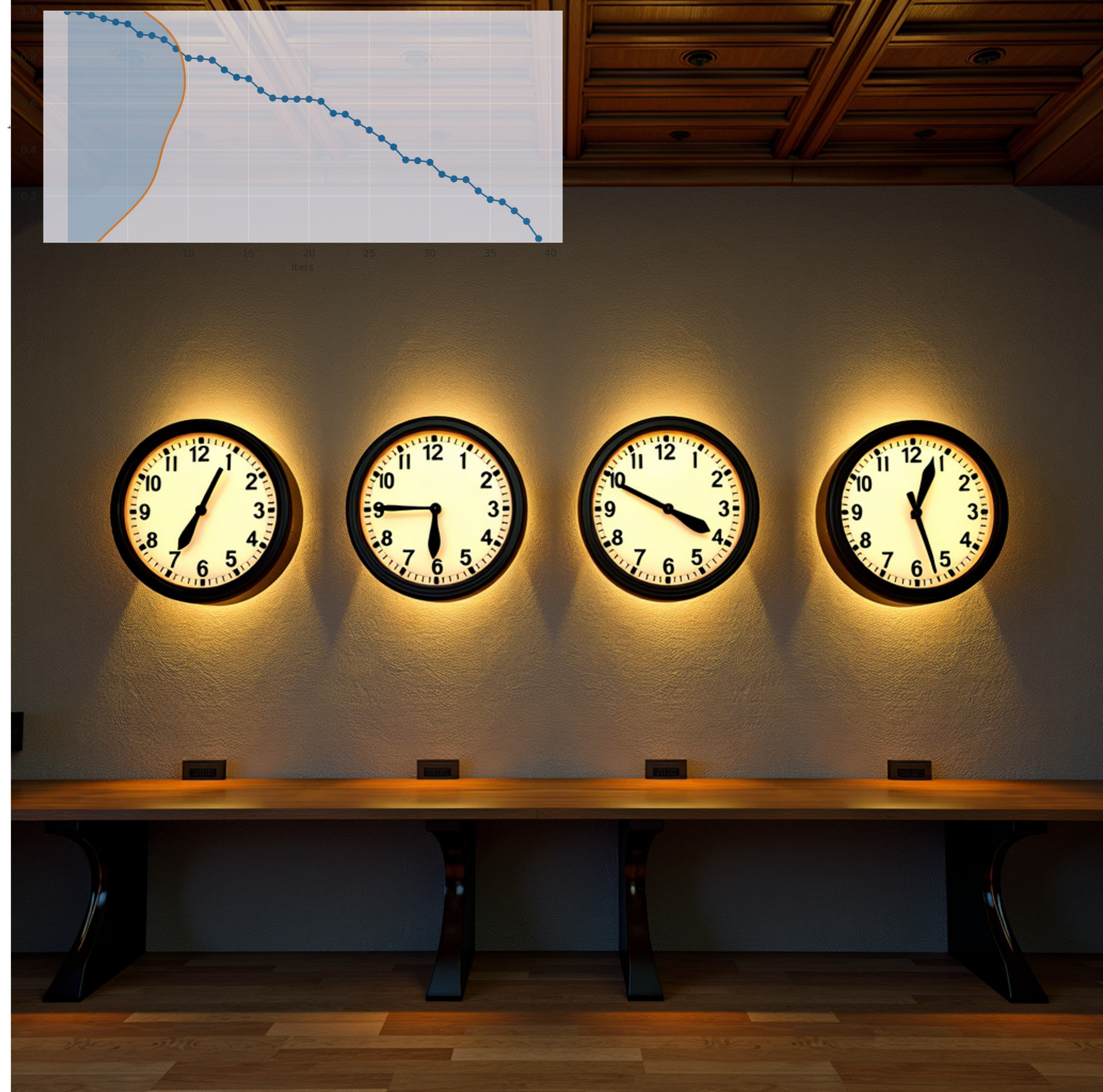}
  \label{fig:appx_ge_r2c2}
\end{subfigure}\hfill
\begin{subfigure}[t]{0.315\linewidth}
  \centering
  \includegraphics[width=\linewidth]{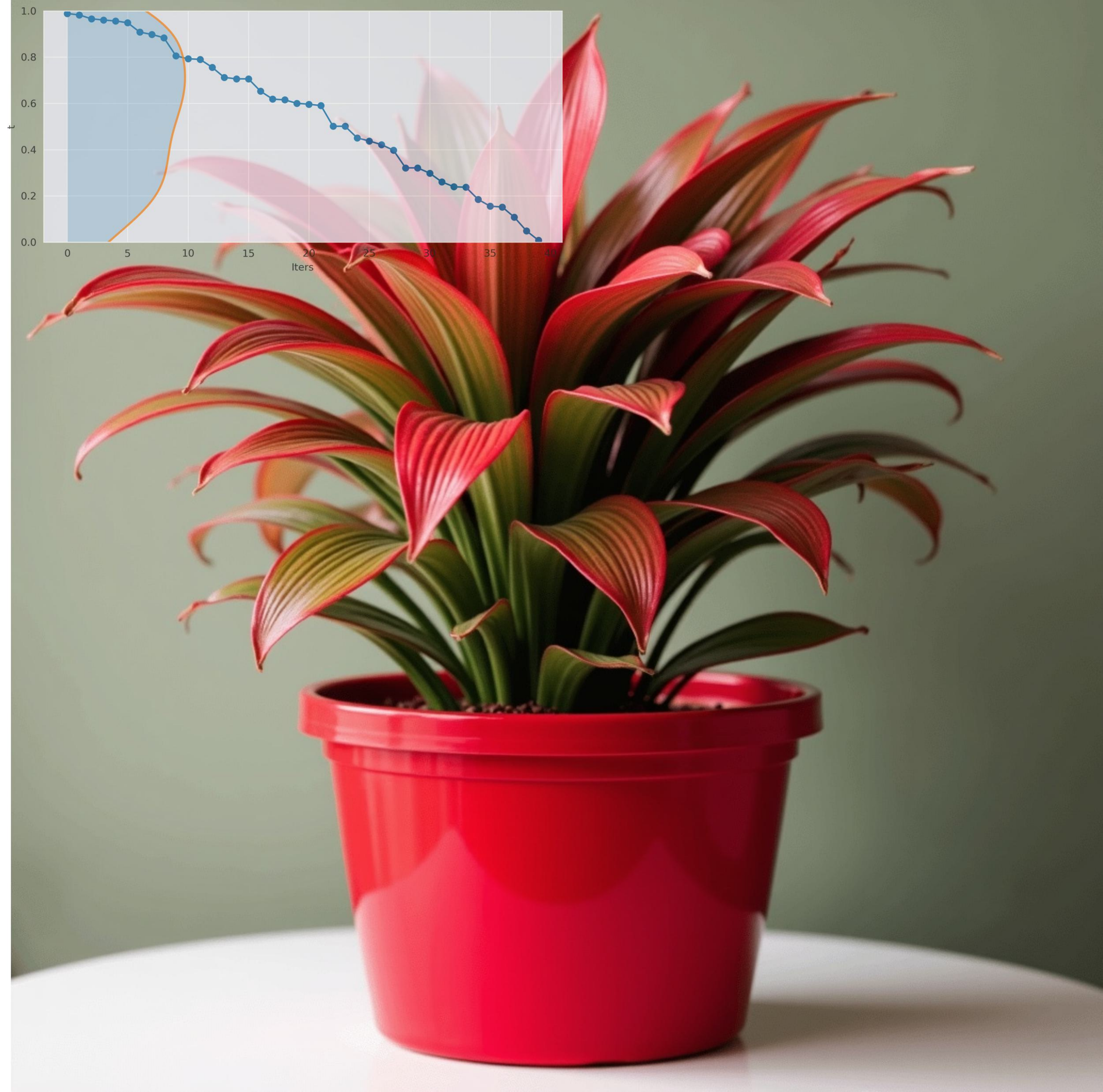}
  \label{fig:appx_ge_r2c3}
\end{subfigure}
\vspace{-10pt}
\caption{\textbf{Rescheduling improves fine grained alignment}. Comparisons between images generated with default schedules (upper) and our learned schedules (lower) from Flux-Dev.}
\label{fig:appx_geneval}
\end{figure*}

\begin{figure*}[t]
\centering

\colhead{a photo of two {\color{red}clocks}}\hfill
\colhead{a photo of a {\color{red} yellow computer keyboard} and a {\color{red} black sink}}\hfill
\colhead{a photo of a couch {\color{red}left of a toaster}}

\begin{subfigure}[t]{0.315\linewidth}
  \centering  \includegraphics[width=\linewidth]{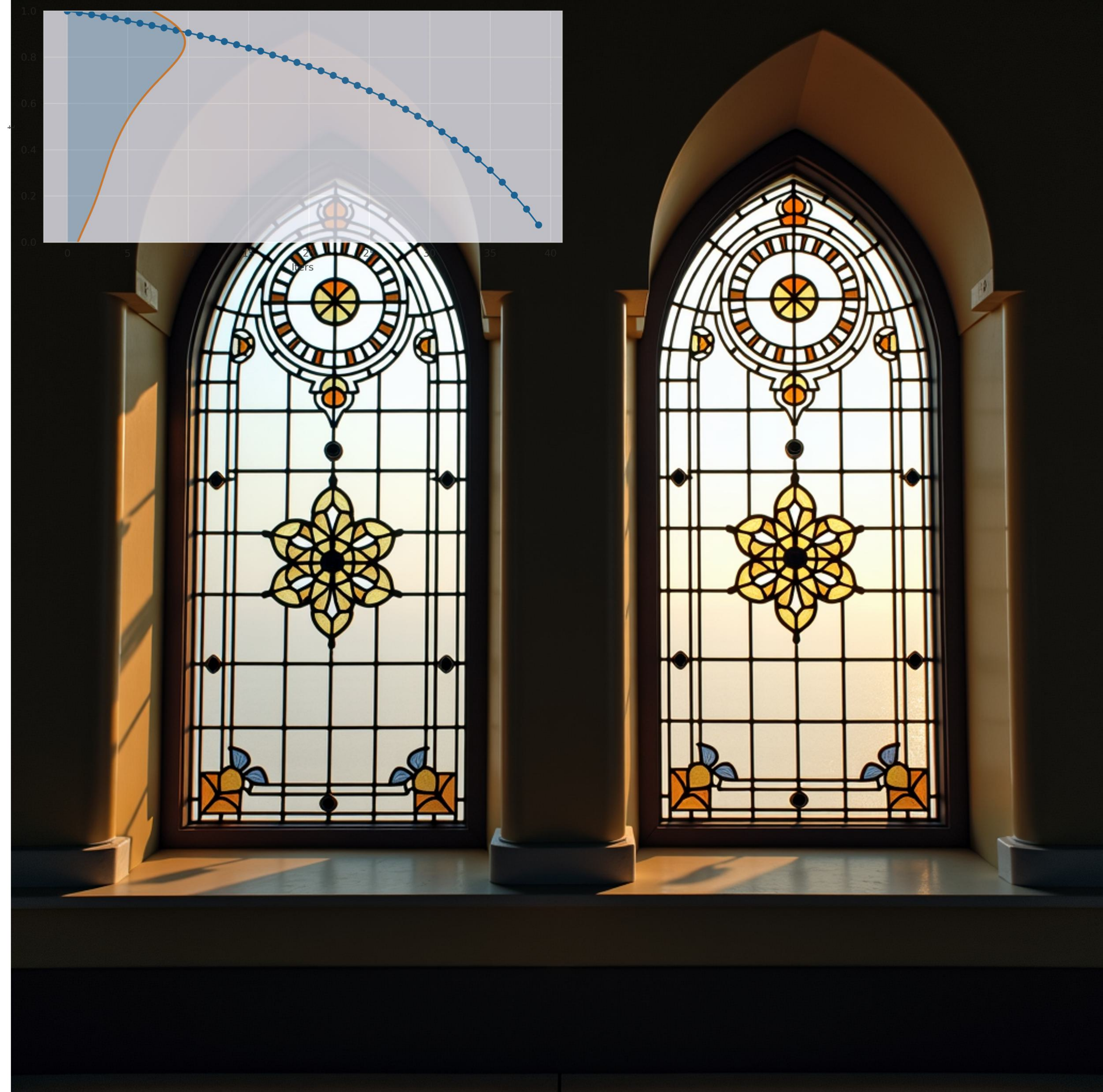}
  \label{fig:appx_ge2_r1c1}
\end{subfigure}\hfill
\begin{subfigure}[t]{0.315\linewidth}
  \centering
  \includegraphics[width=\linewidth]{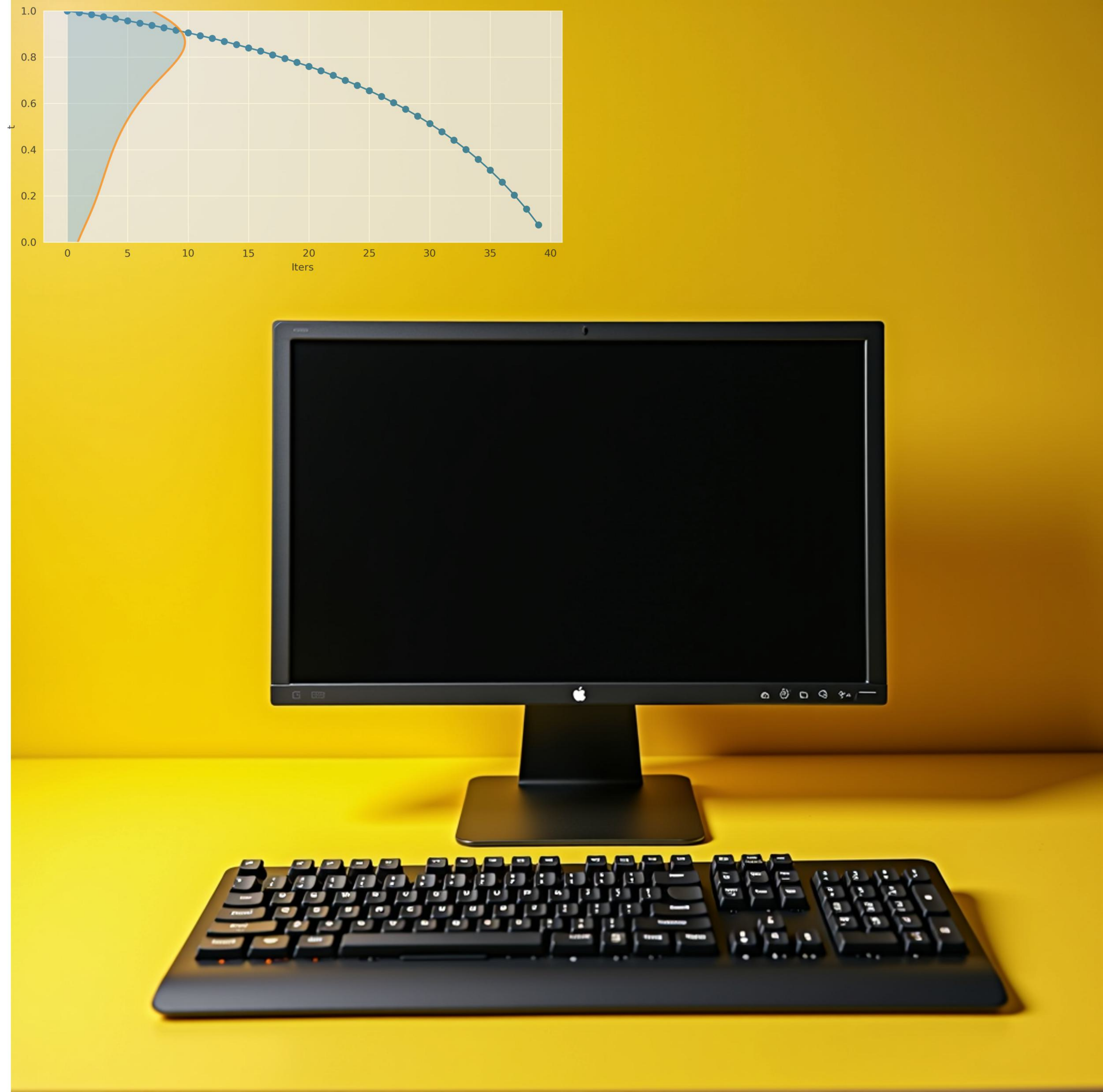}
  \label{fig:appx_ge2_r1c2}
\end{subfigure}\hfill
\begin{subfigure}[t]{0.315\linewidth}
  \centering
  \includegraphics[width=\linewidth]{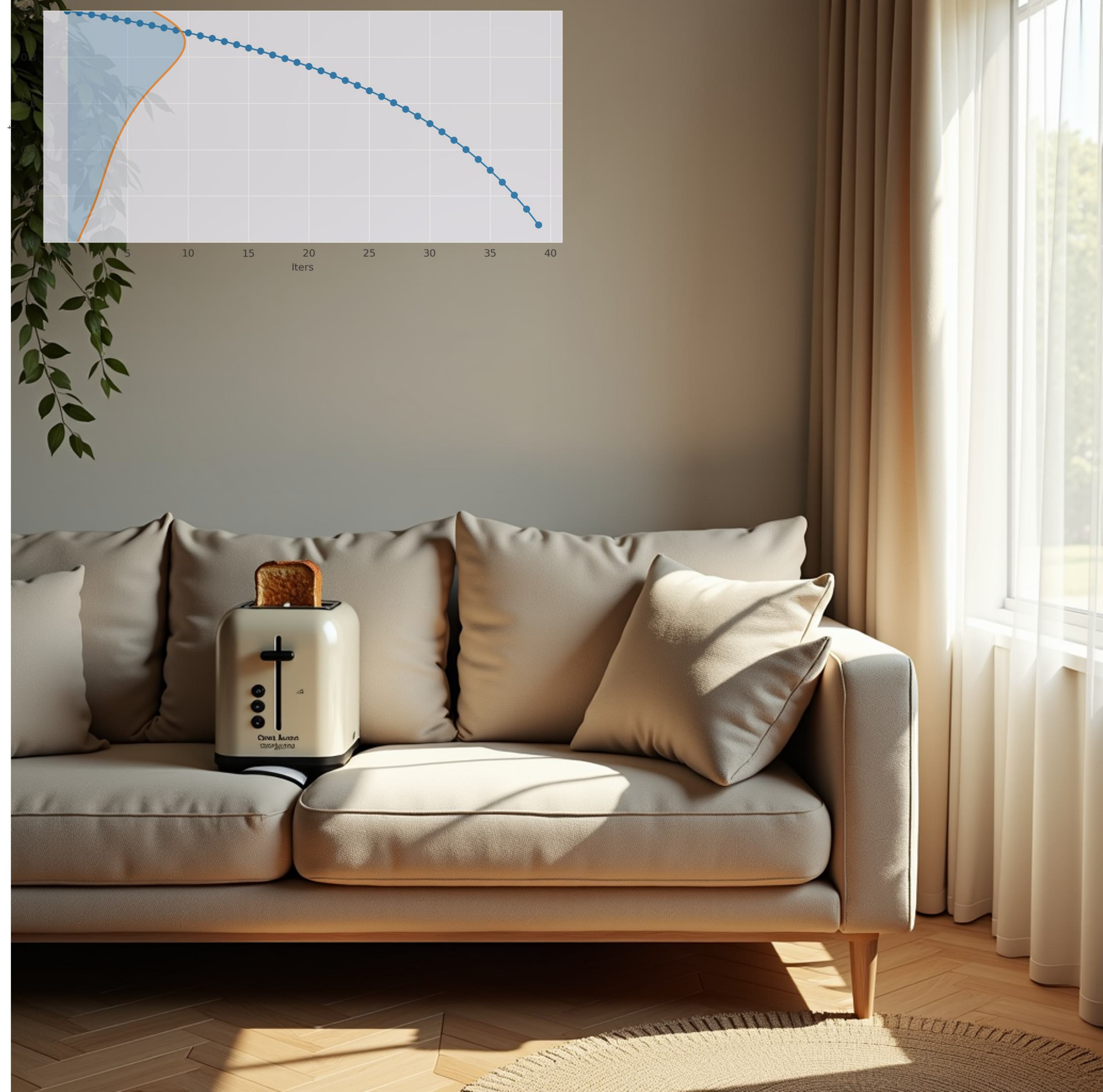}
  \label{fig:appx_ge2_r1c3}
\end{subfigure}

\vspace{-10pt} 

\begin{subfigure}[t]{0.315\linewidth}
  \centering
  \includegraphics[width=\linewidth]{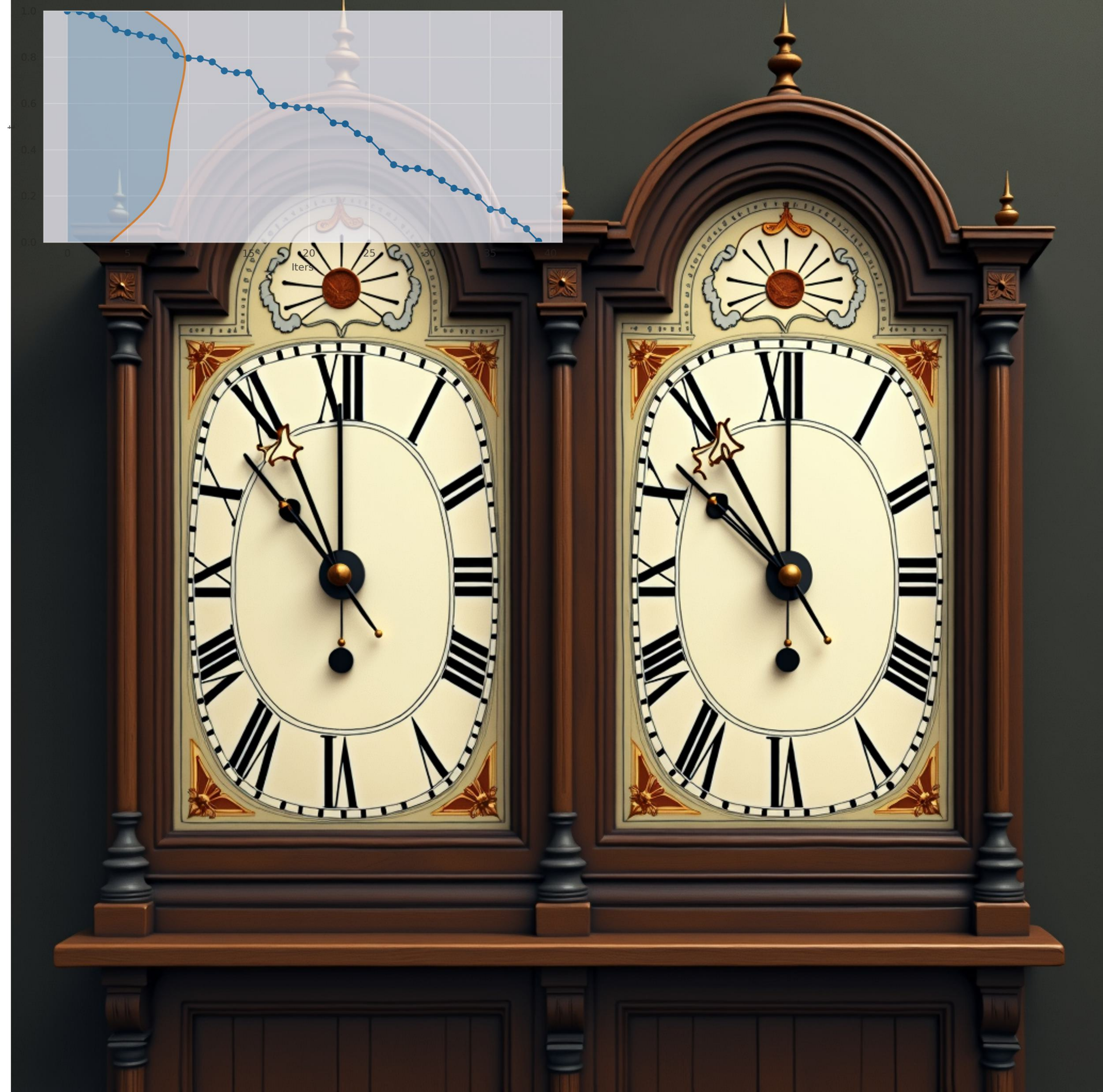}
  \label{fig:appx_ge2_r2c1}
\end{subfigure}\hfill
\begin{subfigure}[t]{0.315\linewidth}
  \centering
  \includegraphics[width=\linewidth]{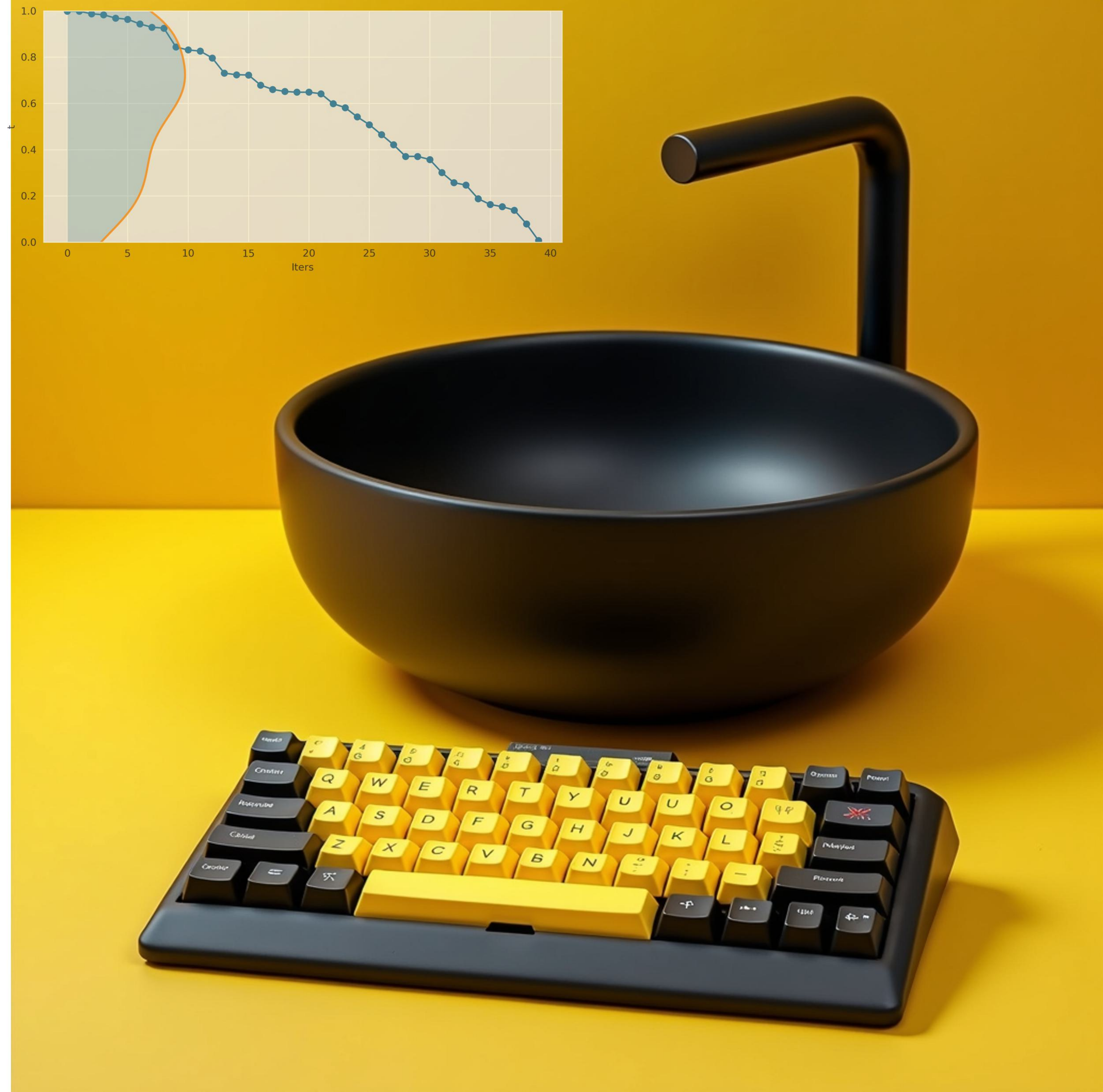}
  \label{fig:appx_ge2_r2c2}
\end{subfigure}\hfill
\begin{subfigure}[t]{0.315\linewidth}
  \centering
  \includegraphics[width=\linewidth]{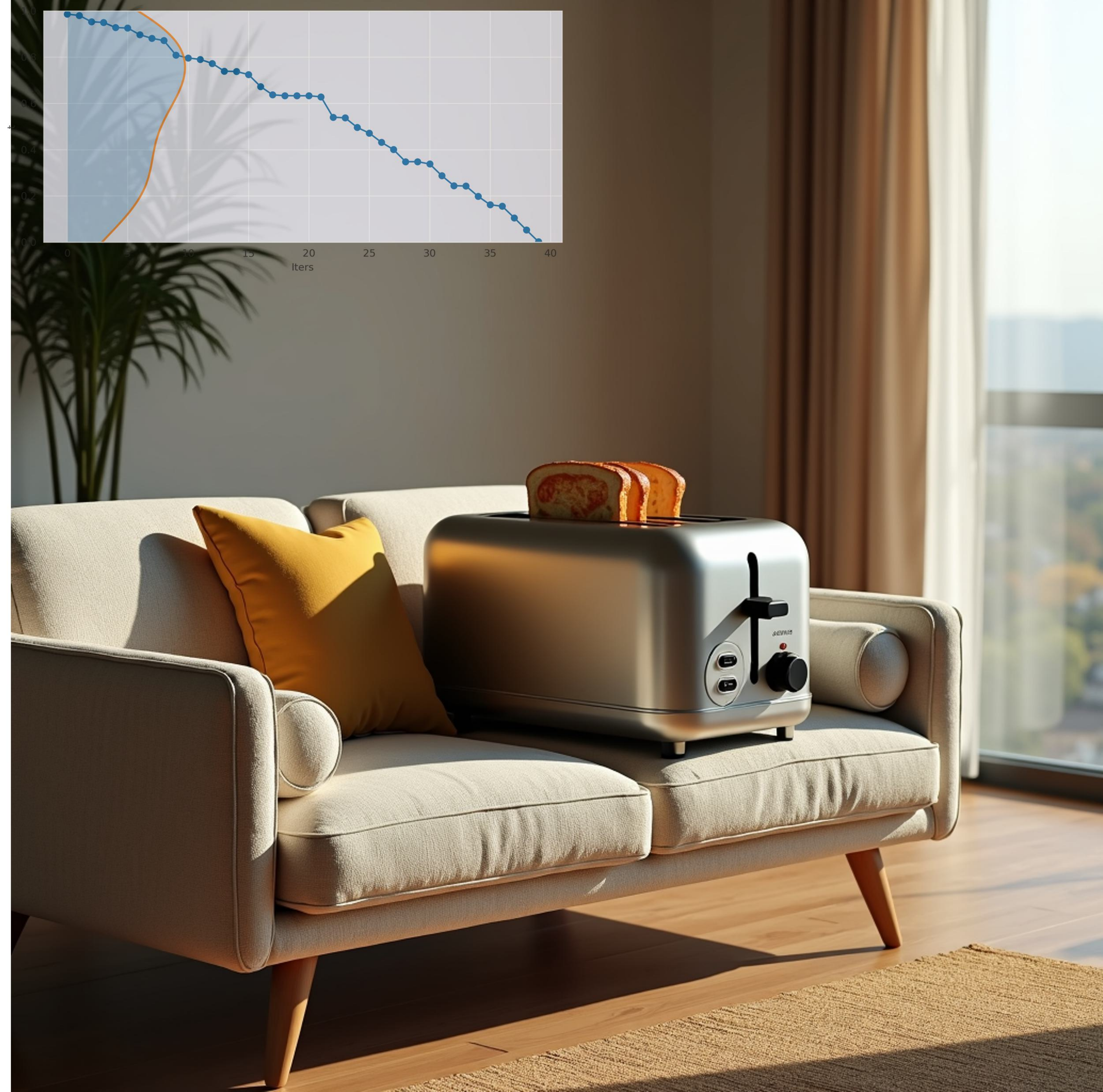}
  \label{fig:appx_ge2_r2c3}
\end{subfigure}
\vspace{-10pt}
\caption{\textbf{Rescheduling improves fine grained alignment}. Comparisons between images generated with default schedules (upper) and our learned schedules (lower) from Flux-Dev.}
\label{fig:appx_geneval_2}
\end{figure*}

We provide further T2I results for 5-step sampling (\cref{fig:appx_few_steps_1,fig:appx_few_steps_2}), general T2I generation on HPDv2 prompts (\cref{fig:appx_gen_t2i_1,fig:appx_gen_t2i_2}), text rendering (\cref{fig:appx_text_render,fig:appx_text_render_2}) and fine-grained object-focused generation (\cref{fig:appx_geneval,fig:appx_geneval_2}). We follow the format of \cref{fig:teaser} and plot the corresponding schedules for generation at the top left corner.

\end{document}